\documentclass[acmsmall,screen,authorversion,nonacm]{acmart}
\acmJournal{TOSEM}

\AtBeginDocument{%
  \providecommand\BibTeX{{%
    \normalfont B\kern-0.5em{\scshape i\kern-0.25em b}\kern-0.8em\TeX}}}

\setcopyright{acmcopyright}


\author{Xiaoning Feng}
\orcid{0009-0001-6344-2477}
\email{fengxiaoning1746@link.tyut.edu.cn}
\affiliation{%
    \institution{Taiyuan University of Technology}
    \city{Tai Yuan}
    \country{China}
}

\author{Xiaohong Han}
\authornote{Corresponding author}
\orcid{0000-0002-8779-6528}
\email{hanxiaohong@tyut.edu.cn}
\affiliation{%
    \institution{Taiyuan University of Technology}
    \city{Tai Yuan}
    \country{China}
}
\author{Simin Chen}
\authornote{Corresponding author}
\orcid{0000-0001-5035-3398}
\email{simin.chen@UTDallas.edu}
\affiliation{%
    \institution{The University of Texas at Dallas}
    \city{Dallas}
    \country{USA}
}

\author{Wei Yang}
\orcid{0000-0002-5338-7347}
\email{wei.yang@utdallas.edu}
\affiliation{%
    \institution{The University of Texas at Dallas}
    \city{Dallas}
    \country{USA}
}


\usepackage{amsmath,amsfonts,bm}
\usepackage{booktabs}
\usepackage{lscape}
\usepackage{CJKutf8}
\usepackage{bm}
\usepackage{lineno}
\usepackage{siunitx}
\usepackage{nicematrix}

\def\1{\bm{1}}

\DeclareMathAlphabet{\mathsfit}{\encodingdefault}{\sfdefault}{m}{sl}
\SetMathAlphabet{\mathsfit}{bold}{\encodingdefault}{\sfdefault}{bx}{n}

\usepackage{xspace}     
\usepackage{graphicx}
\usepackage{booktabs} 
\usepackage{multirow}
\usepackage{algorithm}
\usepackage{wrapfig}
\usepackage{diagbox}
\usepackage{algorithmic}
\usepackage{graphicx}
\usepackage{listings}
\usepackage{xcolor}
\usepackage{tcolorbox}
\usepackage{subfigure}
\usepackage{hyperref}
\usepackage{nicematrix}
\usepackage{booktabs}

\newcommand{\eg}{{\it e.g.,}\xspace}

\newcommand{\ie}{{\it i.e.,}\xspace}

\newcommand\figref[1]{Fig.~\ref{#1}}

\newcommand\tabref[1]{Table~\ref{#1}}

\newcommand\secref[1]{Sec.~\ref{#1}}
\newcommand\equref[1]{Eq.(\ref{#1})}

\usepackage{colortbl}

\def \tool{\texttt{LLMEffiChecker}\xspace}

\definecolor{codegreen}{rgb}{0,0.6,0}
\definecolor{codegray}{rgb}{0.5,0.5,0.5}
\definecolor{codepurple}{rgb}{0.58,0,0.82}
\definecolor{backcolour}{rgb}{0.95,0.95,0.92}

\lstdefinestyle{mystyle}{
  backgroundcolor=\color{backcolour},   commentstyle=\color{codegreen},
  keywordstyle=\color{magenta},
  numberstyle=\tiny\color{codegray},
  stringstyle=\color{codepurple},
  basicstyle=\ttfamily\footnotesize,
  breakatwhitespace=false,         
  breaklines=true,                 
  captionpos=b,                    
  keepspaces=true,                 
  numbers=left,                    
  numbersep=5pt,                  
  showspaces=false,                
  showstringspaces=false,
  showtabs=false,                  
  tabsize=2
}

\newcommand{\fakeparagraph}[1]{\vspace{1mm}\noindent\textbf{#1.}}
\usepackage{balance}

\begin{document}

\setcopyright{acmlicensed}
\acmJournal{TOSEM}
\acmYear{2024} 
\acmVolume{1} 
\acmNumber{1} 
\acmArticle{1} 
\acmMonth{1}
\acmDOI{10.1145/3664812}

\title{\tool: Understanding and Testing Efficiency Degradation of Large Language Models}

\begin{abstract}

Large Language Models (LLMs) have received much recent attention due to their human-level accuracy. While existing works mostly focus on either improving accuracy or testing accuracy robustness, the computation efficiency of LLMs, which is of paramount importance due to often vast generation demands and real-time requirements, has surprisingly received little attention. In this paper, we make the first attempt to understand and test potential computation efficiency robustness in state-of-the-art LLMs. 
By analyzing the working mechanism and implementation of 20,543 public-accessible LLMs, we observe a fundamental property in LLMs that could be manipulated in an adversarial manner to reduce computation efficiency significantly. 
Our interesting observation is that the output length determines the computation efficiency of LLMs instead of the input, where the output length depends on two factors: an often sufficiently large yet pessimistic pre-configured threshold controlling the max number of iterations and a runtime generated end of sentence (EOS) token. Our key motivation is to generate test inputs that could sufficiently delay the generation of EOS such that LLMs would have to go through enough iterations to satisfy the pre-configured threshold.
We present \tool, which can work under both white-box setting and black-box setting. In the white-box scenario,  \tool develops a gradient-guided technique that searches for a minimal and unnoticeable perturbation at character-level, token-level, and structure-level. In the black-box scenario, \tool employs a causal inference-based approach to find critical tokens and similarly applies three levels of imperceptible perturbation to them. Both the white-box and black-box settings effectively delay the appearance of EOS, compelling these inputs to reach the naturally-unreachable threshold.
To demonstrate the effectiveness of \tool, we conduct a systematic evaluation on nine public-available LLMs: Google T5, AllenAI WMT14, Helsinki-NLP translator, Facebook FairSeq, UNICAMP-DL translator, MarianMT, Google FLAN-T5, MBZUAI LaMini-GPT and Salesforce CodeGen. Experimental results show that \tool can increase on average LLMs' response latency and energy consumption by 325\% to 3244\% and 344\% to 3616\%, respectively, by perturbing just one character or token in the input sentence.
Our case study shows that inputs generated by \tool significantly affect the battery power in real-world mobile devices (\ie drain more than 30 times battery power than normal inputs).
\end{abstract}

\maketitle

\section{Introduction}

Large Language Model (LLM) is a promising approach that applies neural networks to resolve various text generation problems. LLMs have received significant recent attention from both academia~\cite{kang2023large,meyer2023chatgpt,chang2023survey,BelinkovB18} and industry~\cite{zhang2021cpm,zhang2022opt,workshop2022bloom,askell2021general,nakano2021webgpt, li2022competition, hoffmann2022training}, due to its advantages over traditional text generation methods (\eg N-gram language models \cite{siivola2005growing}).
For instance, due to being capable of capturing rather long dependencies in sentences, LLMs are seeing a wide adoption in commercial text generation including OpenAI’s GPT products(\eg ChatGPT)~\cite{radford2019language,brown2020language,chen2021evaluating,ouyang2022training} and Meta's LLaMA products~\cite{touvron2023llama, touvron2023llama2, roziere2023code}.

Much research has been done on enhancing the accuracy of LLMs~\cite{xiao2023smoothquant, lin2022language}. Recently, research~\citep{nmt_se1, nmt_se2, nmt_se3, nmt_se4} has been conducted to understand the accuracy robustness of existing LLMs by developing a series of adversarial test input generation frameworks that reduce the generation accuracy of existing LLMs. 
While accuracy robustness is clearly important, we observe that the computation efficiency of LLMs, particularly in terms of the latency and energy spent on generating an input with a specific length, is an equivalently critical property that has surprisingly received little attention. 
A common and unique characteristic of the LLMs domain is the need to process a huge amount of real-time requests (\eg OpenAI's ChatGPT has an average monthly visit volume of 15 billion and an average daily consultation volume of approximately 270 million times~\cite{george2023review, liu2023summary, ray2023chatgpt}). The vast demand for generation requests combined with the real-time requirements naturally makes the computation efficiency of any LLM be one of the most critical optimization goals. 
In this paper, we make the first attempt in understanding and testing potential vulnerabilities in terms of the computation efficiency of existing LLMs.

\noindent\textbf{Key observations revealing vulnerabilities on LLMs computation efficiency.} Our findings are motivated by several observations. 
Particularly, through analyzing the working mechanisms and detailed implementation of 20,543 public-accessible LLMs (\eg Google FLAN-T5 \cite{chung2022scaling}, BigScience T0 \cite{sanh2021multitask}), we observe a fundamental property of LLMs that could be manipulated in an adversarial manner to significantly reduce computation efficiency. 
Specifically, we observe that the computation efficiency of LLMs is highly sensitive to different inputs, even those exhibiting just minor differences.
For instance, slightly modifying an input could incur an order of magnitude more computation demand (\eg as shown in \figref{fig:motivation}, inserting a character ``b'' in token ``Genäckstück'' will increase the latency of HuggingFace's LLM from 0.876s to 20.382s, representing an over 20$\times$ latency increase). 
Such dramatic impact on computation efficiency may occur fundamentally because LLMs often need to invoke the underlying decoder with non-deterministic numbers of iterations to generate outputs~\citep{vaswani2017attention, liu2020multilingual}. 
Intuitively, the computation efficiency of LLMs is determined by the output length instead of the input, where the output length depends on two factors: an often sufficiently large yet pessimistic pre-configured threshold controlling the max number of iterations (e.g., as shown in \figref{fig:config_study}, a dominant number of our studied LLMs set this threshold to be over 300, which is significantly larger than the actual output length in most cases), and a runtime generated end of sentence (EOS) token. By observing such properties, our key motivation is that it may be possible to generate test inputs that could sufficiently delay the generation of EOS such that LLMs would have to go through max iterations to satisfy the pessimistic pre-configured threshold.

This implies an important yet unexplored vulnerability of LLMs: adversarially-designed inputs that may cause enormous, abnormal computation demand in existing LLMs, thus significantly wasting the computational resources and energy and may adversely impair user experience and even service availability.
Such adversarial inputs could result in devastating consequences for many real-world applications (also proved by our experiments). For example, abusing computational resources on commercial text generation service providers~(\eg \textit{HuggingFace}~\citep{wolf-etal-2020-transformers}) could negatively impact the quality of service (\eg enormously long response time or even denial of service). 
For application domains that are sensitive to latency or energy, such as mobile and IoT devices, abusing computational resources might consume battery in an unaffordable fast manner.

Motivated by these observations, we aim to systematically develop a framework that generates inputs to test the robustness w.r.t computation efficiency of LLMs. 
The generated test inputs may significantly increase the computational demand and thus hinder the computation efficiency regarding response latency, energy consumption, and availability. To make such testing practical, any generated LLMs test inputs shall not be attack-obvious. One objective is thus to make trivial or unnoticeable modifications on normal textual inputs to generate such test inputs.  
We present \tool that effectively achieves our objectives. 
\tool is developed based on the aforementioned observation. Specifically, LLMs iteratively compute the output token until either the system generates an end-of-sentence~(EOS) token or a pre-configured threshold controlling the max number of iterations has been met. 
For our studied 20,543 LLMs \footnote{https://huggingface.co/models?pipeline\_tag=text2text-generation\&sort=downloads}, the appearance of EOS is computed from the underlying DNNs output probability. \tool develops techniques that could perturb input sentences to change the underlying DNNs output probability and sufficiently delay the generation of EOS, thus forcing these inputs to reach the naturally-unreachable threshold. 
In the white-box setting, \tool further develops a gradient-guided technique that searches for a minimal perturbation (including both character-level, token-level, and structure-level ones) that can effectively delay the generation of EOS.
In the black-box setting, \tool utilizes a causal inference-based method to identify crucial tokens without relying on gradient information and correspondingly applies three levels of imperceptible perturbation to effectively degrade the efficiency of LLMs.
Applying the above minimal perturbation on the seed input would result in significantly longer output, costing LLMs more computational resources and thus reducing computation efficiency.

\noindent\textbf{Implementation and evaluation.} 
We have conducted extensive experiments to evaluate the effectiveness of \tool. 
Particularly, we applied \tool on nine real-world public-available and widely used (\eg with more than 2,714,275 downloads in Nov 2023) LLMs~(\ie Google T5~\citep{raffel2019exploring, T5}, AllenAI WMT14 \cite{AllenAI}, Helsinki-NLP~\cite{H-NLP}, Facebook Fairseq \cite{ng2019facebook}, UNICAMP-DL Translator \cite{lopes2020lite}, MarianMT \cite{MarianMT}, Google FLAN-T5 \cite{chung2022scaling}, MBZUAI LaMini-GPT \cite{wu2023lamini} and Salesforce CodeGen \cite{nijkamp2022codegen}).
The selected LLMs are trained with different corpus and feature diverse DNN architectures as well as various configurations.
We compare \tool against four state-of-the-art methods that focus on testing LLMs' accuracy and correctness.
Evaluation results show that \tool is highly effective in generating test inputs to degrade the computation efficiency of the LLMs under test. 
Specifically, \tool generates test inputs that could increase the LLMs' CPU latency, CPU energy consumption, GPU latency, and GPU energy consumption by 322\% to 3154\%, 366\% to 3053\%,   327\% to 1969\%, and 322\% to 1966\%, respectively, through only perturbing one character or token in any seed input sentences.
Our case study shows that inputs generated by \tool significantly affect the battery power in real-world mobile devices (\ie drain more than 30 times battery power than normal inputs).

\noindent\textbf{Contribution.}  Our contributions are summarized as follows:
\begin{itemize}

\item Characterization: We are the first to study and characterize the computation efficiency vulnerability in state-of-the-art LLMs, which may critically impair latency and energy performance, as well as user experience and service availability. Such vulnerability is revealed by conducting extensive empirical studies on 20,543 public-available LLMs, which have been downloaded more than 3,260,064 times in Nov/2023. The results show that the revealed vulnerability could widely exist due to a fundamental property of LLMs.

\item Approach: We design and implement \tool, the first framework for testing LLMs' computation efficiency. Specifically, given a seed input, \tool applies gradient-guided  and causal inference-based methods to mutate the seed input to generate test inputs in white-box and black-box settings respectively. Test inputs generated by \tool only perturb one to three tokens in any seed inputs.

\item Evaluation: We evaluate \tool on nine real-world public-available LLMs (\ie  Google T5, AllenAI WMT14, Helsinki-NLP, Facebook FairSeq, U-DL
Translator, MarianMT, FLAN-T5, LaMini-GPT and CodeGen) against four correctness-based testing methods. In addition, we propose a series of metrics (\equref{eq:metric}) to quantify the effectiveness of the triggered computation efficiency degradation.
Evaluation results suggest existing correctness-based testing methods cannot generate test inputs that impact computation efficiency. In contrast, \tool generates test inputs that increase LLMs' latency and energy consumption by 291\% to 12536\% and 207\% to 11172\%, respectively.

\item Mitigation: We propose a lightweight method to mitigate possible computation efficiency degradation: running a detector at runtime for input validation. We evaluate the performance of our proposed mitigation method in terms of accuracy and additional overheads. Results confirm the efficacy and efficiency of our proposed mitigation method.

\end{itemize}

This article represents a substantial expansion of our prior research featured in ESEC/FSE 2022 \cite{chen2022nmtsloth}. This extension encompasses several key advancements: (1) Diversification of Testing Scope: We have broadened our focus from efficiency testing specific to neural machine translation (NMT) models to encompass a broader range, specifically targeting General Large Language Models (LLMs). The scope of our study is now more inclusive, as detailed in the \secref{sec:preliminary}. (2) Introduction of a Black-Box Approach: In addition to the original white-box methodology, we have introduced a novel black-box approach, as explained in \secref{sec:black_box}. This innovative methodology is designed to operate effectively under realistic scenarios, offering a more robust evaluation of the model's performance. (3) Expanded Subject Evaluation: Going beyond the confines of NMT models, our research evaluates our proposed framework across a wider array of subjects. This includes a comprehensive assessment of the framework's applicability to LLMs for diverse applications, such as sentence completion and code generation.

\section{Background}
\label{sec:background}

\subsection{Working Mechanism Of Large Language Models}
\label{sec:nmt}


\begin{figure}[htbp]
\centering
\subfigure[The Encoder-Decoder architecture]{
\begin{minipage}[t]{0.48\textwidth}
\centering
\includegraphics[height=3cm]{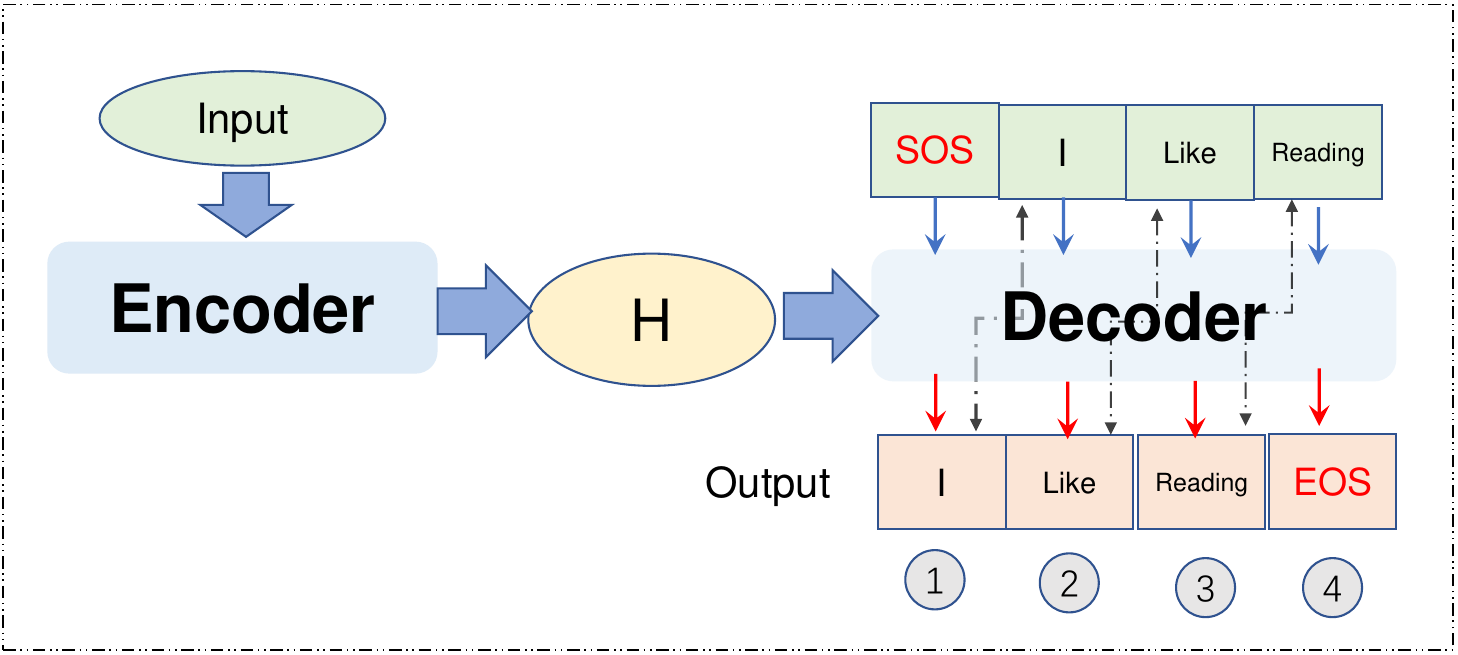}
\end{minipage}
}
\subfigure[The Decoder-Only architecture]{
\begin{minipage}[t]{0.48\textwidth}
\centering
\includegraphics[height=3cm]{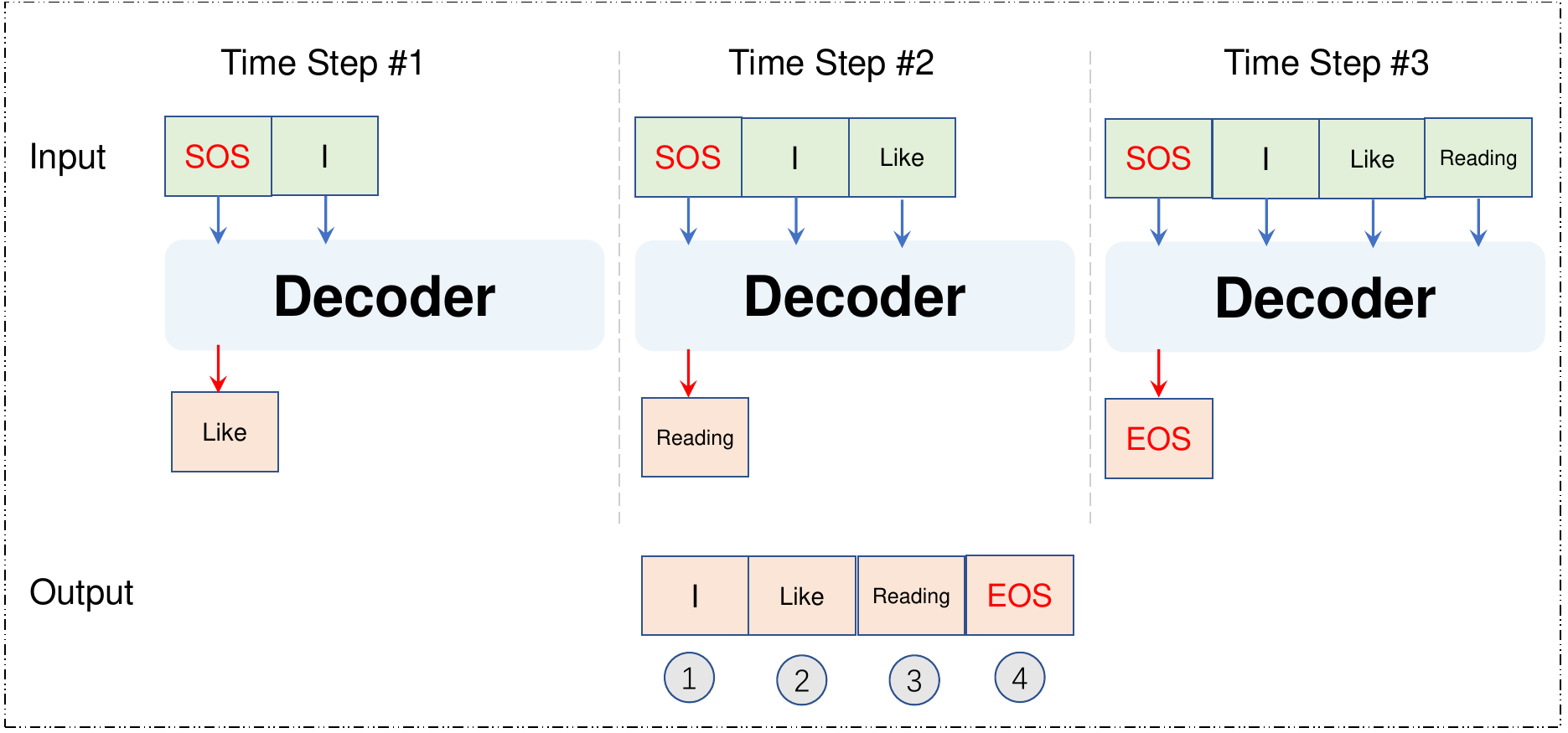}
\end{minipage}
}
\caption{Working mechanism of LLMs}
\label{fig:nmt}
\end{figure}

Much recent research has been done towards developing more accurate and efficient large language models (LLMs) \cite{xiao2023smoothquant, liu2020multilingual, vaswani2017attention, sutskever2014sequence, google_blog1, google_blog2, google_blog3}.
The language model computes the conditional probability $P(Y | X)$, where $X = [x_1, x_2, \cdots, x_m]$ is the input token sequence and $Y = [y_1, y_2, \cdots, y_n]$ is the output token sequence.
Modern LLMs apply the neural networks to approximate such conditional probability $P(Y | X)$.
As shown in \figref{fig:nmt}, 
The structure of LLMs can be broadly categorized into two types: the Encoder-Decoder architecture (e.g., Google T5 series) and the Decoder-Only architecture (e.g., OpenAI GPT series). The encoder $f_{en}(\cdot)$ encodes the source input $X$ into hidden representation $H$, then $H$ is fed into the decoder for decoding. 
Notably, the attention layers in the encoder possess the capacity to analyze all words within the initial sentence, whereas the attention layers of the decoder $f_{de}(\cdot)$ can only access the words positioned before a given word in the input. Consequently, these two architectures are often chosen for different tasks. The Encoder-Decoder architecture is well-suited for tasks involving sequence-to-sequence mappings, (\eg  translation and summarization). 
While the Decoder-Only architecture is more fitting for autoregressive generation tasks, characterized by the sequential generation of output sequences (e.g., text continuation and dialogue systems), it excels in predicting the next piece of text based on the sequence that has already been generated (or a given initial text).
An implementation example of LLMs' decoding process is shown in Listing \ref{lst:logic} \footnote{The code snippet is downloaded from \texttt{PyTorch} LLM tutorial}.
From the code snippet, we observe that the decoding process starts with a special token~(SOS) and iteratively accesses $H$ for an auto-regressive generation of each token $y_i$ until the end of sequence token (EOS) or the maximum iteration (\eg \texttt{max\_length}) is reached (whichever condition is reached earlier).
To improve LLMs' accuracy, a common practice is to apply the beam search algorithm to search multiple top tokens at each iteration and select the best one after the whole decoding process.

\begin{lstlisting}[language=Python,label=lst:logic, caption=Source Code Example of LLMs Implementation]
'''
Encoding process
'''
decoded_words = ['<SOS>']
for di in range(max_length):
    decoder_output, decoder_hidden = decoder( decoder_input, decoder_hidden, encoder_outputs)
    topv, topi = decoder_output.data.topk(1)
    if topi.item() == EOS_token:
        decoded_words.append('<EOS>')
        break
    else:
        decoded_words.append(index2word[topi.item()])
        decoder_input = topi.squeeze().detach()
return decoded_words
\end{lstlisting}

\subsection{Robustness Testing for NLP Systems}
 
Although modern NLP systems demonstrate human-level performance in terms of accuracy, NLP systems are still far from robust due to the complexity and intractability of the underlying neural networks. 
To improve the robustness of NLP systems, a series of testing methods have been proposed, which focus on accuracy testing.
The core idea of existing work is to perturb seed input sentences with different perturbations and detect output inconsistency between perturbed and seed outputs.
At high-level, the perturbations in existing work can be categorized into three types.
\textit{(i) character-level:} This type of perturbations~\citep{BelinkovB18, li2018textbugger, character_nlp1, character_nlp2, character_nlp3} represents the natural typos and noises in textual inputs. For example, character swap (\eg noise $\rightarrow$ nosie),  order random (\eg noise $\rightarrow$ nisoe), character
insertions (\eg noise $\rightarrow$ noisde), and keyboard typo (\eg noise $\rightarrow$ noide);
\textit{(ii) token-level:} This type of perturbations \cite{nmt_se2, token_nlp1, li2018textbugger, token_nlp2, cheng2020seq2sick, token_nlp3} replaces a few tokens in the seed sentences with other tokens. However, token replacement sometimes would completely change the semantic of the input text; thus, this type of perturbation usually appears in adversary scenarios;
\textit{(iii) structure-level:}  Different from the above two perturbations, this type of perturbations \cite{sentence_nlp1, nmt_se1, nmt_se3, nmt_se4} seeks to generate legal sentences that do not contain lexical or syntactic errors. For example, \citep{nmt_se1} proposes a structure invariant testing method to perturb seed inputs with \texttt{Bert} \cite{jin2020bert}, and the perturbed sentences will exhibit similar sentence structure with the seed sentences.

\section{Motivation \& Preliminary Study}
\label{sec:preliminary}

In this section, we first give a motivating example in detail to show efficiency degradation issues in real-world large language models (LLMs). We then present a comprehensive empirical study based on 20,543 state-of-the-art LLMs, which reveals an important vulnerability in existing LLMs that may suffer from significant efficiency degradation.

\subsection{Motivating Example}
\label{sec:motivation}



\begin{figure}[htbp]
\centering
\subfigure[Model in Translation]{
\begin{minipage}[t]{0.46\textwidth}
\centering
\includegraphics[height=3.3cm]{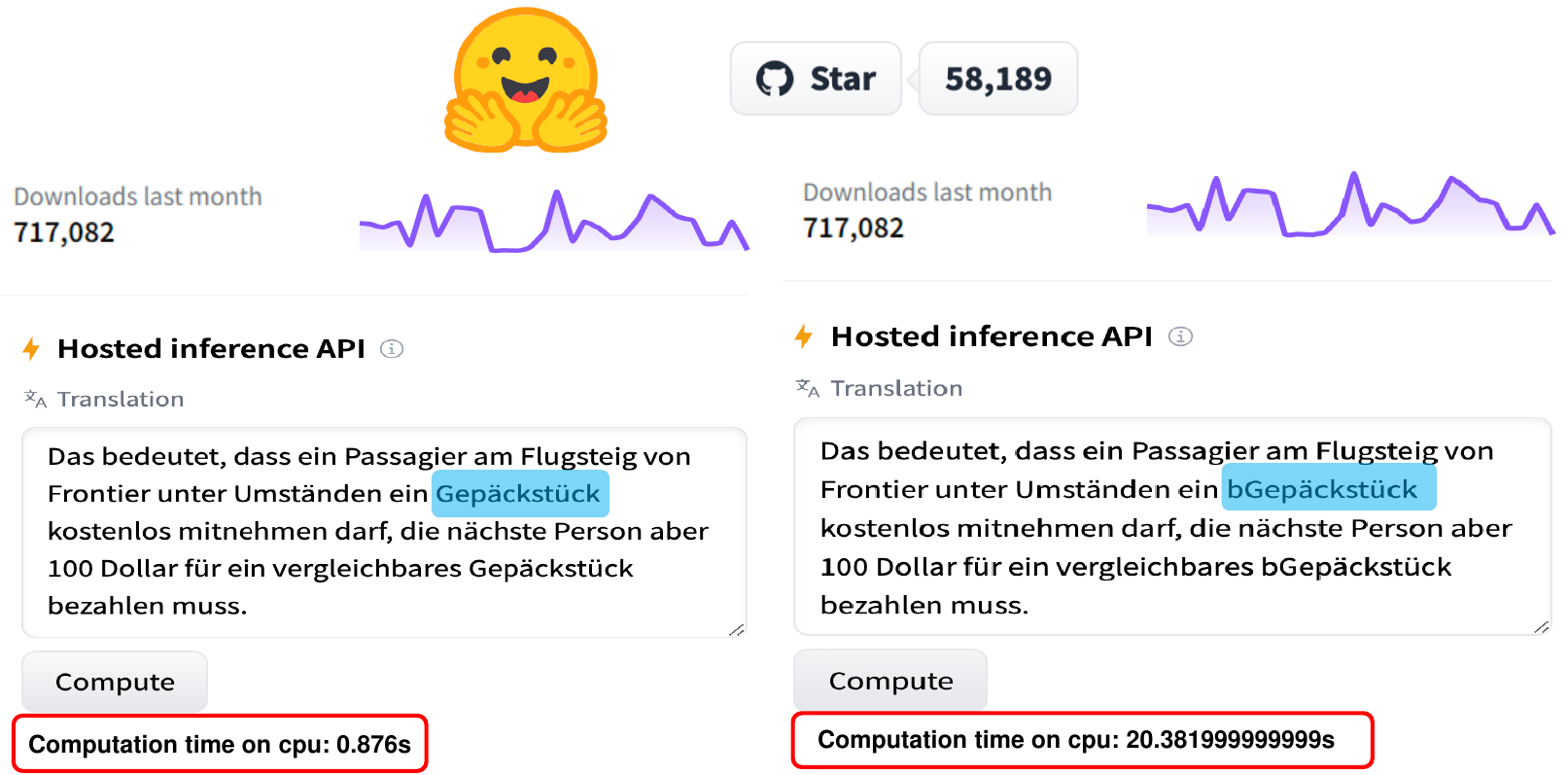}
\end{minipage}
}
\subfigure[Model in Code Generation]{
\begin{minipage}[t]{0.50\textwidth}
\centering
\includegraphics[height=3.5cm]{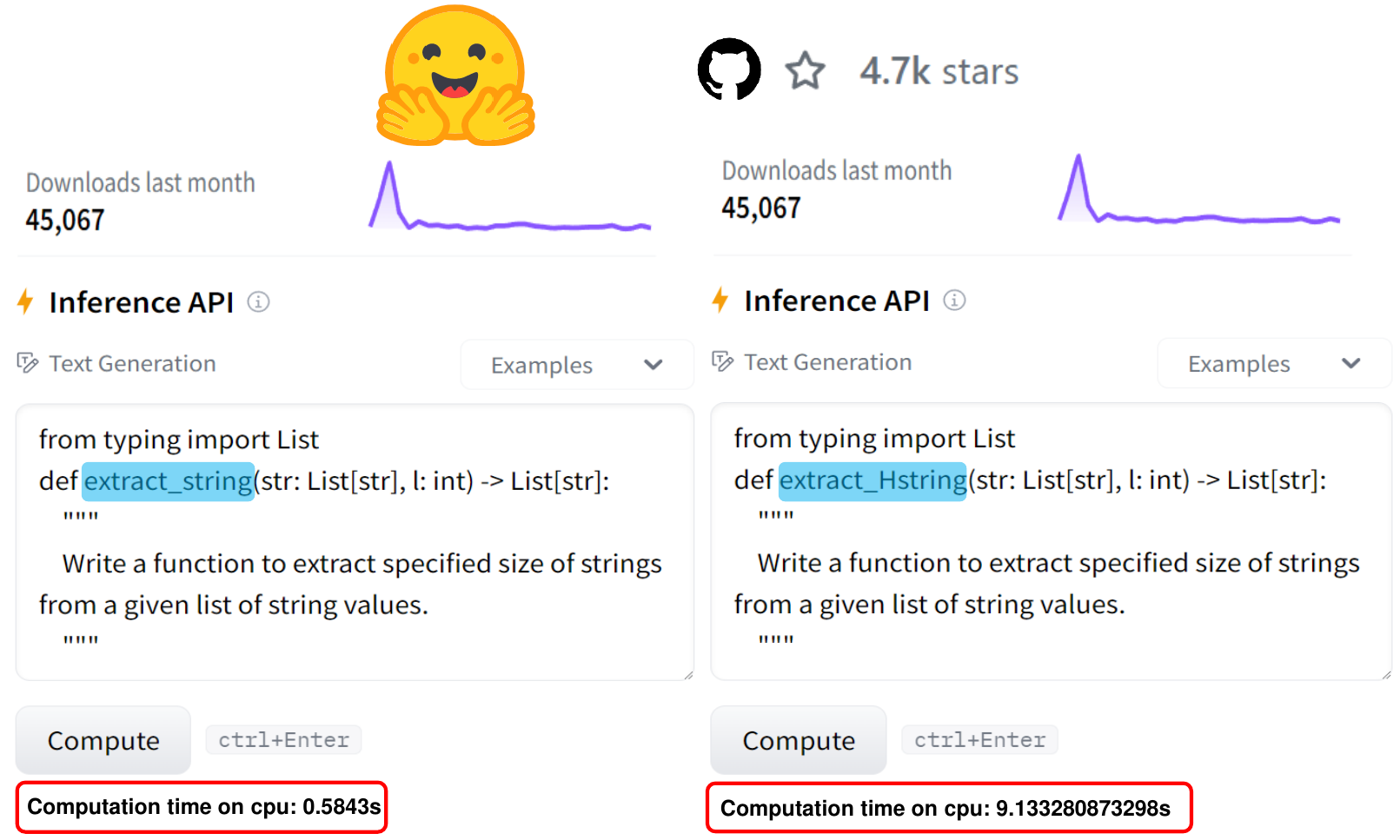}
\end{minipage}
}
\caption{Examples illustrating LLMs' efficiency degradation by 
    inserting one character (using HuggingFace API)}
\label{fig:motivation}
\end{figure}

\figref{fig:motivation} illustrates the efficiency degradation issue that HuggingFace LLMs APIs may experience due to  unnoticeable perturbations.
Sub-figure (a) depicts Helsinki's model~\footnote{https://huggingface.co/Helsinki-NLP/opus-mt-de-en} specialized in translating from German to English, while sub-figure (b) showcases Salesforce's CodeGen model~\footnote{https://huggingface.co/Salesforce/codegen-350M-mono} tailored for code synthesis tasks.
The selected LLMs APIs are rather popular among developers, with 717,082 and 45,067 downloads merely in Feb 2024.
\figref{fig:motivation} shows the computation time of LLMs in different scenarios using two input sentences, where a normal (abnormal) input is used in the left (right)  part of the sub-figure. Note that the abnormal input differs from the normal input by only one character ``b'' or ``H'' (highlighted in blue). Nonetheless, due to such a one-character difference in the input, the computation time increases from 0.876s to 20.382s (a 2226.7\% increase) and 0.5843s to 9.133s (a 1474.1\% increase). This real-world example reveals that state-of-the-art LLMs may have critical yet unrevealed vulnerabilities that negatively impact computation efficiency.

 As we discussed in \secref{sec:nmt}, the working mechanism of LLMs is to iteratively call the decoder $f_{de}(\cdot)$ to generate output tokens until either the particular token EOS is reached or the pre-configured threshold is met. 
Thus, LLMs with more decoder calls (\ie denoted as $||f_{de}(\cdot)||$) will consume more computational resources and incur longer computational times.
An intuitive approach to mitigate the efficiency degradation issue in \figref{fig:motivation} is to set a small threshold to limit $||f_{de}(\cdot)||$.
However, this solution is impractical due to inherently significant differences of $||f_{de}(\cdot)||$ in the text generation corpus. 
According to our empirical study of  20,543 LLMs (detailed in \ref{sec:practice}), the majority of them set \texttt{max\_length} over 300.
To further understand why this intuitive approach does not work, we conduct a comprehensive empirical study using 20,543 state-of-the-art LLMs. Specifically, we focus on answering the following two research questions.

\begin{itemize}
    \item \textbf{RQ 1.1}: \textit{What is the current engineering configurations in real-world LLMs that control $||f_{de}(\cdot)||$} (\secref{sec:practice})
    \item \textbf{RQ 1.2}: \textit{Why small threshold is impractical to mitigate efficiency degradation?} (\secref{sec:variance})
\end{itemize}

\subsection{Current Engineering Configurations}
\label{sec:practice}

\vspace{-0.3cm}
\begin{table}[htbp]
  \centering  

  \caption{Top 10 popular LLMs on HuggingFace website (the order is based on the number of downloads)}
  \resizebox{0.58\textwidth}{!}{
      \begin{tabular}{c|lcc}
    \toprule
    \textbf{Rank} & \multicolumn{1}{c}{\textbf{Model Name }} & \textbf{max\_length} & \textbf{\# of Downloads} \\
    \midrule
    \textbf{1} & gpt2  & 50    & 23,723,037 \\
    \textbf{2} & tiiuae/falcon-7b-instruct & 2048  & 8,068,318 \\
    \textbf{3} & distilgpt2 & 50    & 4,812,521 \\
    \textbf{4} & Kyle1668/boss-toxicity-t5-large & 300    & 4,400,913 \\
    \textbf{5} & facebook/mbart-large-50 & 200    & 4,080,895 \\
    \textbf{6} & stabilityai/StableBeluga-7B & 4096  & 3,480,702 \\
    \textbf{7} & Kyle1668/boss-sentiment-t5-large & 200    & 3,402,617 \\
    \textbf{8} & t5-small & 300  & 2,714,275 \\
    \textbf{9} & t5-base & 300  & 2,132,545 \\
    \textbf{10} & google/flan-t5-base & 300  & 1,307,572 \\
    \bottomrule
    \end{tabular}%
  \label{tab:llm}%
  }
\end{table}%
\vspace{-0.3cm}

\subsubsection{Study Methodology}
We investigate the configurations of existing mainstream LLMs.
Specifically, we study 20,543 LLMs (\eg Google Flan-T5, BigScience BLOOMZ) from HuggingFace online LLMs service \footnote{\href{https://huggingface.co/}{https://huggingface.co/}}.
HuggingFace is a commercial platform that provides third-party real-time NLP service, which covers almost all LLMs architectures.
LLMs on the HuggingFace platform are open-source and widely used by public, as shown in \tabref{tab:llm} (\eg the most popular LLMs in HuggingFace have been downloaded for more than 23,723,037 times in Nov 2023).
HuggingFace provides high-level abstraction API for LLMs service. List \ref{lst:api} shows code snippets of using HuggingFace API to load Google T5  service.
All language model classes are inherited from a common parent class, \texttt{GenerationMixin}, which contains all functions supporting text generation. 
We parse the source code of the \texttt{GenerationMixin.generate} function and observe that the generation flow could be divided into nine parts. 
Among all nine parts, we find that the eighth part determines the critical stopping criteria.
The source code of the eighth part is shown in List \ref{lst:stop}.
From the source code, we observe that two variables, \texttt{max\_length} and \texttt{max\_time}, determine the stopping criteria.
\texttt{max\_length} is a variable from the LLMs' configuration file that determines the maximum length of the sequence to be generated, equivalent to the maximum number of decoder calls mentioned earlier. Similarly, \texttt{max\_time} is a variable that determines the maximum computation time. According to HuggingFace programming specifications, only one of these two fields needs to be set.
Finally, we select all LLMs in the Text2Text Generation column from HuggingFace's API services~\footnote{\href{https://huggingface.co/models?pipeline_tag=text2text-generation&sort=downloads}{https://huggingface.co/models?pipeline\_tag=text2text-generation\&sort=downloads}} and parse each LLM's configuration file to check how \texttt{max\_length} and \texttt{max\_time} have been set.

\begin{lstlisting}[language=Python,label=lst:api, caption=HuggingFace libraries high-level text generation API]
# HuggingFace high-level API for text generation
model = AutoModelWithLMHead.from_pretrained("t5-base")
s = "CS is the study of computational systems" 
input_tk = tokenizer(s, return_tensors="pt").input_ids
res_tk = model.generate(input_tk)

\end{lstlisting}

\begin{lstlisting}[language=Python,label=lst:stop, caption= Stopping criteria in text generation]
# 8. prepare stopping criteria
stopping_criteria = self._get_stopping_criteria(
                max_length=max_length,
                max_time=max_time, 
                stopping_criteria=stopping_criteria)
\end{lstlisting}

\begin{figure}[h]
	\centering
	\begin{minipage}[t]{0.48\textwidth}

		\includegraphics[width=1.0\textwidth]{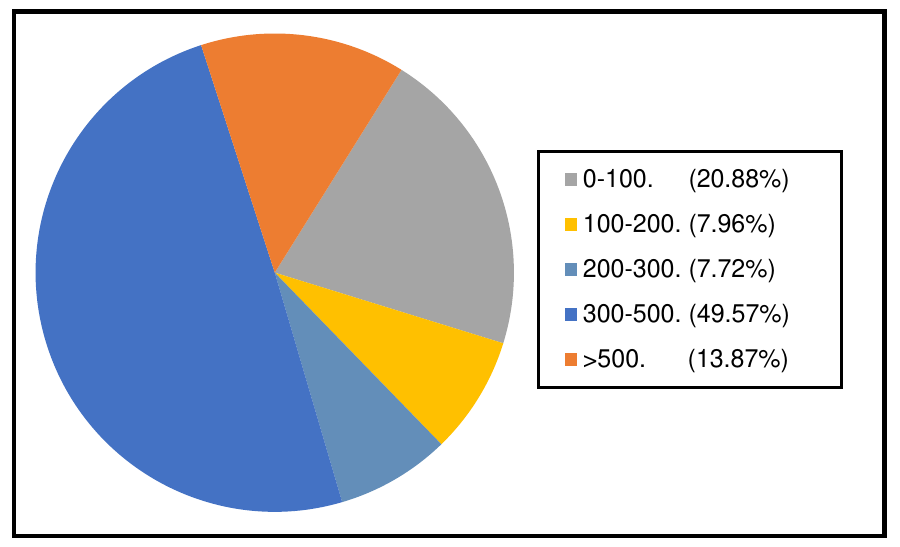}
		\caption{The distribution of \texttt{max\_length} values}
		\label{fig:config_study}
	\end{minipage}
\end{figure}

\subsubsection{Study Results}

Among all $20,826$ LLMs, we successfully collect $20,543$ configuration files, where $14,266$ of them include the \texttt{max\_length} field and none of them includes the \texttt{max\_time} field. This is mainly because the \texttt{max\_time} field is hardware-dependent. 
The statistical results of the \texttt{max\_length} values are shown in \figref{fig:config_study}.
We have the following two observations. First, there is a significant variance in the \texttt{max\_length} value (ranging from 8 to 16896); second, the majority of LLMs (63.44\%) configure the \texttt{max\_length} to values surpassing 300, \ie the maximum decoder invocation exceeds this threshold.
Furthermore, if there are no specifications for \texttt{max\_length} in the model configuration, it potentially indicates a bug, as this omission could lead to unpredictable behavior and may not align with the user's expectations for the generated text. We present the following two evidences. Firstly, when utilizing HuggingFace's transformers library to load a model (\eg List \ref{lst:api}), if \texttt{max\_length} is not specified in the model configuration file, the default value is set to 20. It is strongly advised in the official documentation to set an appropriate value manually\footnote{\url{https://huggingface.co/docs/transformers/v4.38.1/en/llm_tutorial}}. The default small value is a conservative choice to facilitate a quick start for users, as longer outputs necessitate increased computational resources (time and memory) for generation, processing, and storage. However, this default value is insufficient to convey adequate information, necessitating users to define a reasonable \texttt{max\_length} manually. Detailed arguments on this matter will be provided in the subsequent \secref{sec:variance}. Secondly, decoder-only LLMs also return the input prompt as part of the output. Consequently, if the input length exceeds 20 tokens, the model will not produce any output and trigger a UserWarning: "Input length exceeds the default max\_length (=20)." This may result in unexpected behavior. 
Note that real-world LLMs prefer to set such a large threshold just to prevent unresponsiveness (\eg dead-loop). However, in most cases with normal inputs, such a threshold will not yield any real impact as the EOS token often appears much earlier (\eg in code generation applications, setting the \texttt{max\_length} of LLMs to 512 is a widely adopted practice \cite{liu2024your,zheng2023codegeex,cassano2023multipl}).

\subsection{Feasibility Analysis of an Intuitive Solution}
\label{sec:variance}

\subsubsection{Study Methodology}

An intuitive solution to mitigate the efficiency degradation is to limit $||f_{de}(\cdot)||$ (\ie the \texttt{max\_length} field).
In this section, we conduct a statistical analysis to prove that such an intuitive solution is infeasible. 
We analyze the distribution of \texttt{max\_length} of the target sentence (ground truth) in the training corpus. 
We select the MultiUN dataset~\cite{multiun} as the subject in our empirical study because of the following criteria: \textit{(i)} the datasets are open-source and public-available; \textit{(ii)} the datasets are widely studied in recent works (with more than 1,000 citations until Nov 2023); \textit{(iii)} the datasets are diverse in covering various areas (\eg different languages, concepts, etc). 
MultiUN dataset is a collection of translated documents from the United Nations. It includes seven languages with 489,334 files and a total number of 81.41M sentence fragments.
We parse the source/target sentence pairs in the MultiUN dataset and measure the length of all target sentences.

\subsubsection{Study Results}

\begin{table}[htbp]
  \centering
  \caption{Statistical results of efficiency differences in LLMs (1\%, 10\%, 50\%, 90\%, 100\% represent quantile)}
  \resizebox{0.86\textwidth}{!}{
    \begin{tabular}{cc|r|cccc|ccccc}
    \toprule
    \multicolumn{2}{c|}{Language} & \multirow{2}[2]{*}{\# of pairs} & \multicolumn{4}{c|}{Quantile of Target Length } & \multicolumn{5}{c}{Quantile of Length Ratio} \\
    Src   & Tgt   &       & 10\%    & 50\%    & 90\%    & 100\% (max)   & 1\% (min)   & 10\%    & 50\%    & 90\%    & 100\% (max) \\
    \midrule
    fr    & en    & 13,172,019 & 4.00  & 24.00  & 52.00  & 97.00  & 0.50  & 0.87  & 1.10  & 1.47  & 3.00  \\
    zh    & en    & 9,564,315 & 11.00  & 41.00  & 87.00  & 179.00  & 0.90  & 1.38  & 1.83  & 3.00  & 8.26  \\
    zh    & es    & 9,847,770 & 10.00  & 40.00  & 87.00  & 176.00  & 0.75  & 1.19  & 1.57  & 2.68  & 8.50  \\
    zh    & fr    & 9,690,914 & 11.00  & 41.00  & 88.00  & 178.00  & 0.74  & 1.21  & 1.63  & 2.85  & 8.29  \\
    zh    & ru    & 9,557,007 & 10.00  & 42.00  & 90.00  & 180.00  & 0.62  & 1.60  & 2.25  & 5.00  & 13.75  \\
    \bottomrule
    \end{tabular}%
}
  \label{tab:statistic}%
\end{table}%

The statistic results of the output length are shown in \tabref{tab:statistic} (full results could be found in an anonymous website~\footnote{https://github.com/Cap-Ning/LLMEffiChecker}). Column ``Quantile of Target Length'' shows the target sentence length under different quantiles, and Column ``Quantile of Length Ratio'' shows the ratio of sentence length between the source and target.
From the results, we observe that the lengths of target sentences (ground truth) are in sparse distributions. Particularly, the ratio of sentence length between the source and target exhibits rather large variance. For instance, the length of target sentence varies from 4 to 97 and the ratio is from 0.62 to 13.75 for language \texttt{fr} and \texttt{en}. As a result, setting a small \texttt{max\_length} field will lead to low-precision generation results. For instance, in the last line of \tabref{tab:statistic}, \ie generating \texttt{zh} to \texttt{ru}, if setting \texttt{max\_length} to 42, at least 50\% of data will not be generated completely. Thus, we can conclude that the intuitive solution, \ie setting a small \texttt{max\_length} field, is impractical to avoid efficiency degradation issues.
On the contrary, setting a sufficiently large \texttt{max\_length} can address the limitation of incomplete text generation while not incurring efficiency issues for any ordinary inputs due to the EOS mechanism.

\section{Problem Formulation}

\begin{figure*}[ht]
    \centering

    \includegraphics[width=0.9\textwidth]{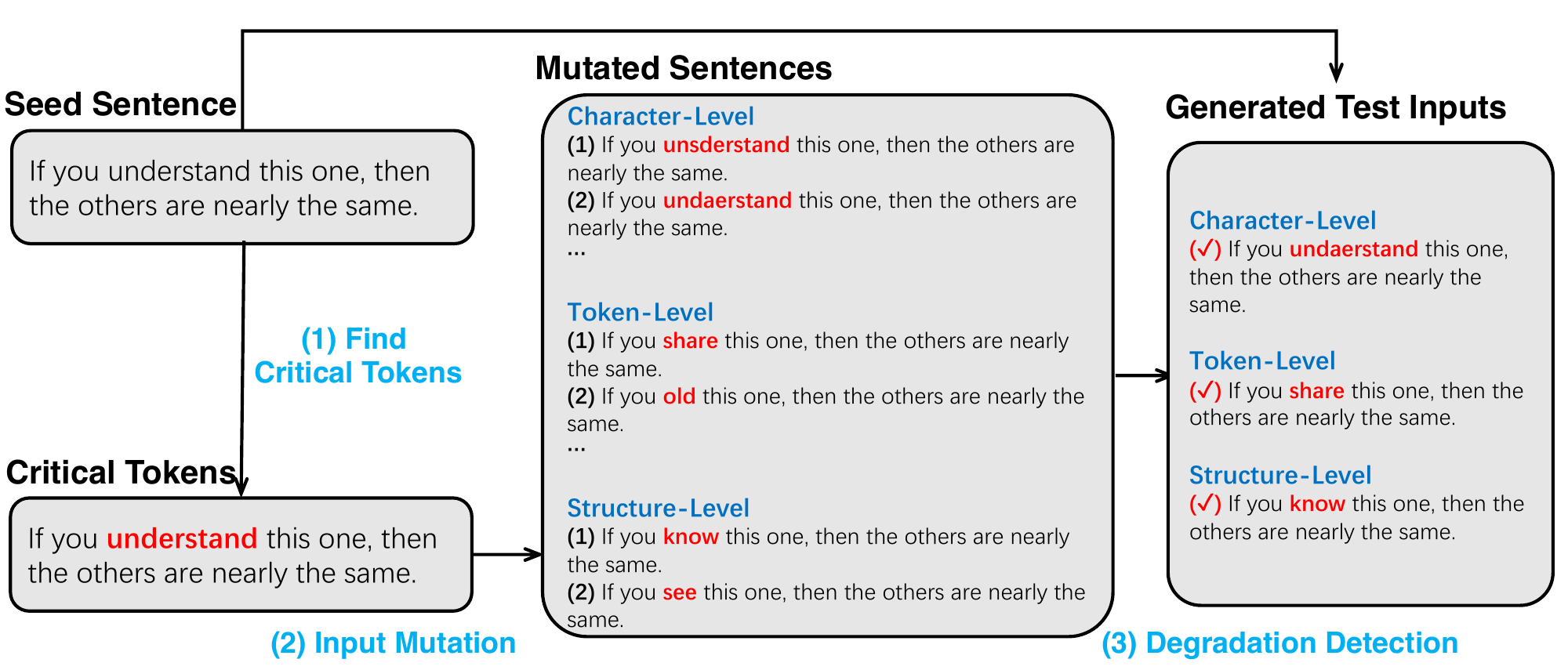}

    \caption{Design overview of \tool}
    \label{fig:overview}
\end{figure*}

Our goal is to generate test inputs that can degrade the computation efficiency of LLMs.
Our proposed method seeks to perturb a seed sentence to craft test inputs.
The perturbed test inputs will incur significantly longer computation time, thus impairing user experience and even causing service unavailability.
Note that we allow general and unnoticeable perturbation patterns, including adding limited number of characters (\eg 1-3 characters) at arbitrary positions and replacing arbitrary tokens using semantic-equivalent alternatives.
%
As we discussed in \secref{sec:background}, LLMs' computation efficiency depends on the output length, where a lengthier output implies less computation efficiency.
Thus, our goal can be achieved through increasing LLMs' output length through generating effective test inputs.
We thus formulate our problem of generating test inputs for computation efficiency testing as the following optimization:
\begin{equation}
\label{eq:definition}
    \Delta = \text{argmax}_{\delta} \quad ||f_{de}(x + \delta)|| \quad\quad s.t. \; ||\delta|| \le \epsilon,
\end{equation}
\noindent where $x$ is the seed input, $f_{de}(\cdot)$ is the decoder of the LLMs under test, $\epsilon$ is the maximum allowed perturbation, and $||f_{de}(\cdot)||$ measures the number of times of LLMs's decoders being called.
Our proposed \tool tries to search a perturbation $\Delta$ that maximizes the decoders' calling times~(decreasing target LLMs efficiency) within a minimum allowable perturbation threshold (which ensures unnoticeable perturbations).

\section{Methodology}
\label{sec:whitebox}

We now present \tool, designed for both white-box and black-box scenarios. It provides three specific implementations: character-level perturbation, token-level perturbation, and structure-level perturbation. 

\subsection{Design Overview}

\tool demonstrates practicality by functioning seamlessly in both white-box and black-box settings. In either scenario, \tool employs an iterative process where it systematically perturbs a single token within a seed sentence using various types of perturbations.
A design overview of the procedural steps for each iteration is presented in \figref{fig:overview}. This illustration encapsulates three pivotal steps applicable to both white-box and black-box settings:

\begin{enumerate}
    \item  \textit{Finding critical tokens.} For each seed sentence, we feed it to LLMs under test. 
    In the white-box setting, \tool applies a gradient-based approach to identify critical tokens with the highest impact on the computation efficiency of LLMs. Conversely, in the black-box setting, \tool employs a casual inference-based instead of a gradient-based approach to pinpoint critical tokens that significantly influence LLMs' computational efficiency.

    \item  \textit{Mutating seed input sentences.} After identifying the critical tokens in the seed sentences, we mutate the seed sentences with three types of perturbations and generate three lists of similar sentences.

    \item \textit{Detecting efficiency degradation.} We feed the mutated sentences and the seed sentence into LLMs and detect any efficiency degradation.
    
\end{enumerate}

\subsection{White-Box Detail Design}
\label{sec:white_box}
\noindent\textbf{Finding Critical Tokens:} 
Given a seed sentence $x = [tk_{1}, \cdots, tk_m]$, the first step is to identify tokens that are critical to LLMs' efficiency.
As we discussed earlier, LLMs' computation efficiency depends on the corresponding output length given any input, which is determined by the pre-configured threshold and the EOS token. In \secref{sec:preliminary}, we showed that the pre-configured threshold is set as a fixed value in the configuration files of LLMs. Thus, to generate effective testing inputs, our objective is to decrease the probability that the EOS token would appear given a specific input to reduce LLMs' computation efficiency. 

Formally, let LLM's output probability be a sequence of vectors, \ie  $ [p_1, p_2,\cdots, p_n]$, and  the probability of EOS token appearance be $[p_1^{eos}, p_2^{eos}, \cdots, p_n^{eos}]$. 
We seek to find the importance of each token $tk_i$ in $x$ to this probability sequence.
We also observe that the output token sequence will affect  EOS's probability \cite{gao2024inducing}.
Specifically, LLMs generate tokens in the generated sequences based on the generated probability distribution. When the generated sequence is semantically complete or matches a common grammatical structure that typically ends, the model may predict a higher probability for the EOS token. To encourage deviations from the original generated token sequence and focus more on other possible candidate tokens, we incorporate $p_i^{o_i}$ into $f(x)$ to enhance the output uncertainty on each generated token, promoting longer, more complex sequences.
Thus, we define the importance score of token $tk_i$ as $g_i$, shown in \equref{eq:obj}.
\begin{equation}
\label{eq:obj}
        o_i = \text{argmax} (p_i) \qquad  f(x) = \frac{1}{n}\sum_{i}^{n} (p_i^{eos} + p_i^{o_i})  \qquad
       g_i = \sum_{j} \frac{\partial f(x)}{ \partial tk_i^j},
\end{equation}
\noindent where $[o_1, o_2, \cdots, o_n]$ is the current output token, $f(x)$ is the probability we seek to minimize, 
it can delay the generation of the EOS token and introduce more uncertainty for each generated token in the prediction process to break the existing output dependency, thereby maximizing the generation of longer sentences to the fullest extent. $tk_i^j$ is the $j^{th}$ dimension of $tk$'s embeddings, and $g_i$ is the derivative of $f(x)$ to $i^{th}$ token's embedding.
The score $g_i$ assesses the importance of the token $tk_i^j$ for the output length. It is calculated by summing the gradients, which quantify the sensitivity of $f(x)$ to variations in each dimension of the token's embedding.

\noindent\textbf{Input Mutation:} After identifying important tokens, the next step is to mutate the important token with unnoticeable perturbations.
In this step, we get a set of perturbation candidate $L$ after we perturb the most important tokens in the original input.
We consider three kinds of perturbations, \ie character-level perturbation, token-level perturbation and structure-level perturbation. \tabref{tab:per} shows some examples of character-level, token-level and structure-level perturbations with different perturbation sizes $\epsilon$ (the perturbation is highlighted with the color red).

\begin{table}[tbp!]
  \centering
  \caption{Examples of character-level, token-level, and structure-level perturbation under different size }
  \label{tab:per}
  \resizebox{0.65\textwidth}{!}{
    \begin{tabular}{r|c| l}
    \toprule
    \multicolumn{1}{l|}{Original} & $\epsilon$ & Do you know who Rie Miyazawa is? \\
    \midrule

         & 1 & Do you know who Rie \textcolor{red}{Miya-zawa} is? \\
    \multicolumn{1}{l|}{Character-Level} & 2 & Do you know \textcolor{red}{whoo} Rie \textcolor{red}{Miya-zawa} is? \\
    \midrule

          & 1 & Do \textcolor{red}{Hello} know who Rie Miyazawa is? \\
    \multicolumn{1}{l|}{Token-Level} & 2 & Do \textcolor{red}{Hello} know who \textcolor{red}{Hill} Miyazawa is? \\
    \midrule

         & 1 & Do you \textcolor{red}{remember} who Rie Miyazawa is? \\
    \multicolumn{1}{l|}{Structure-Level} & 2 & Do you \textcolor{red}{remember} \textcolor{red}{what} Rie Miyazawa is? \\

    \bottomrule
    \end{tabular}%
    }
  \label{tab:addlabel}%
\end{table}%

For character-level perturbation, we consider character insertion perturbation.
Specifically, we insert one character $c$ into token $tk$ to get another token $\delta$. The character-insert perturbation is common in the real world when typing quickly and can be unnoticeable without careful examination.
Because character insertion is likely to result in out-of-vocabulary~(OOV), it is thus challenging to compute the token replacement increment at token-level. 
Instead, we enumerate possible $\delta$ after character insertion to get a candidate set $L$. Specifically, we consider all letters and digits as the possible character $c$ because humans can type these characters through the keyboard, and we consider all positions as the potential insertion position.
Clearly, for token $tk$ which contains $l$ characters, there are  $(l + 1) \times ||C||$ perturbation candidates, where $||C||$ denotes the size of all possible characters.
For token-level perturbation, we consider replacing the original token $tk$ with another token $\delta$.
To compute the target token $\delta$, we define token replace increment $\mathcal{I}_{src,tgt}$ to measure the efficiency degradation of replacing token $src$ to $tgt$.
As shown in \equref{eq:inc}, $E(\cdot)$ is the function to obtain the corresponding token's embedding, $E(tgt) - E(src)$ represents the vector increment in the embedding space, capturing the semantic and syntactic variation and measuring the impact of the replacement on the sentence's meaning and structure. It explores a wider range of potential outputs, further breaking the original output dependency, leading to more diverse and complex sequences, making it difficult for LLMs to converge to a coherent output.
 Recall to \equref{eq:obj}, $\frac{\partial f(x)}{ \partial tk_i^j}$ indicates the sensitivity of output length to each embedding dimension. Therefore, $\mathcal{I}_{src, tgt}$ denotes the total benefits of replacing token $src$ with $tgt$. We search the target token $\delta$ in the vocabulary to maximize the token replace increment with the source token $tk$.
\begin{equation}
\label{eq:inc}
       \mathcal{I}_{src, tgt} = \sum_{j} (E(tgt) - E(src)) \times \frac{\partial f(x)}{ \partial tk_i^j} 
    \qquad  \delta = \text{argmax}_{tgt} \; \mathcal{I}_{tk, tgt}; 
\end{equation}

\begin{figure}[htbp]
    \centering
    \includegraphics[width=0.65\textwidth]{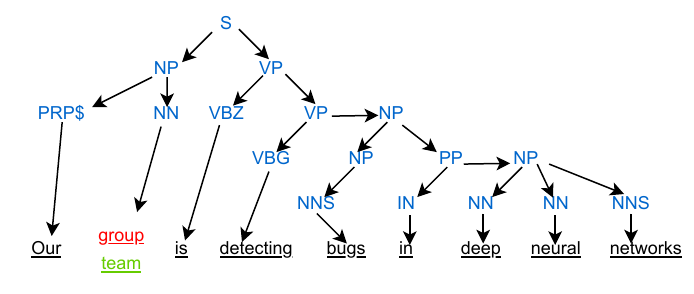}
    \caption{Constituency tree of sentence}
    \label{fig:constituency}
\end{figure}

For structure-level perturbation, we follow existing work \cite{nmt_se1, nmt_se2} to parse the seed input sentence as a constituency tree and replace the critical token with another token based on \texttt{Bert} \cite{brendel2019accurate}. Unlike token-level perturbation, the structure-level perturbation ensures the constituency structure of the perturbed sentence is the same as the seed one.
\figref{fig:constituency} shows an example of the structure-level perturbation. To enhance clarity, our explanation utilizes the left section of the tree as an illustrative example. At the apex, the "S" symbolizes the sentence in its entirety. Descending from the top, the sentence splits into a noun phrase (NP) and a verb phrase (VP), representing the basic Subject-Verb-Object (SVO) pattern inherent to elementary English structure. The NP itself breaks down further into a possessive pronoun “PRP\$” (our) and a common noun “NN” (group), indicating “our group” as the subject of the sentence.  Within this seed sentence, “group” has been identified as the critical token. After feeding the parsed information from the sentence constituent tree into the BERT model, the token "team" is produced as a structural perturbation. This method of critical token replacement retains the original sentence structure, affirming the integrity of the constituency tree post-perturbation.

\noindent\textbf{Efficiency Degradation Detection:}
After collecting candidate perturbations $L$, we select an optimal perturbation from the collected candidate sets.
Since our objective is searching this perturbation candidate set that will produce a longer output length, we straightforwardly test all perturbations in this set and select the optimal perturbation that produces the maximum output length.

\subsection{Black-Box Detail Design}
\label{sec:black_box}
\noindent\textbf{Finding Critical Tokens:} 
Note that selecting critical tokens is relatively straightforward in a white-box scenario since it can be easily accomplished by inspecting the gradients of LLMs, while most other tokens are irrelevant. However, in the more common black-box setup, model gradients are unavailable. In black-box settings, employing random mutation to generate test inputs often proves ineffective due to the vastness of the search space. To overcome this challenge, we propose a novel approach grounded in the concepts of delta debugging \cite{wang2021probabilistic} and causal inference \cite{yao2021survey} to identify the critical tokens with the utmost impact on the computational efficiency of LLMs. Additionally, our approach is based on the fundamental conclusion discussed in \secref{sec:background}, which states that the computational efficiency of LLMs depends on the resulting output length for any given input. Longer outputs necessitate more frequent invocations of the decoder during input processing, thereby demanding a higher volume of floating-point operations (FLOPs).
Specifically, we first decompose the input by removing each token from the original input sentence, breaking it down into multiple subsets. By comparing the output length of each subset with the original output length, we pinpoint the sentence with the most substantial difference in output length from the seed sentence. Subsequently, we identify the missing token in this sentence, which constitutes the critical tokens we are seeking. Through this strategic division of the search process, our approach adeptly identifies the critical tokens in black-box scenarios.

Formally, given a seed sentence $S_{\text{orig}} = [tk_{1}, tk_{2}, \cdots, tk_m]$, we generate debugging subsets $S_i$ by removing the token $tk_i$ from $S_{\text{orig}}$. Subsequently, we feed each $S_i$ and $S_{\text{orig}}$ into the target LLM to obtain the corresponding output lengths $O_i$ and $O$. Our objective is to identify the index $j$ that maximizes $\gamma_j$. Once this index $j$ is determined, the critical token is $tk_j$ in $S_{\text{orig}}$ (refer to ~\equref{eq:b_obj}).
\begin{equation}
        \gamma_i = |O_i - O| \qquad  j = argmax_i\gamma_i  
    \label{eq:b_obj}
\end{equation}

Specifically, we conceptualize LLMs as a sequence of mappings that transition from an input domain to an output domain, with each distinct input eliciting a unique output \cite{jin2020bert}. By employing causal inference methods, we modify the inputs and monitor the resultant variations in the outputs. This process enables us to infer the correlation between diverse inputs and their corresponding output lengths, which serve as indicators of the LLMs' computational efficiency. Through this analytical approach, we aim to pinpoint the critical tokens that are instrumental in this dynamic.

{\noindent\textbf{Input Mutation: }The character-level perturbations and structure-level perturbations described in \secref{sec:white_box}} are well-suited for black-box settings. Consequently, we focus specifically on modifying token-level perturbations in this section. Our intuition is that even in the black-box scenario, obtaining the model's vocabulary is relatively straightforward. This is because models performing the same task in the same language typically share similar vocabularies, and the tokens within it are visible in the model input. Consequently, upon identification of the critical tokens, we proceed to randomly select tokens from the vocabulary to effect replacements.

\noindent\textbf{Efficiency Degradation Detection:}
Upon compiling a set of candidate perturbations, denoted as $L$, we proceed to select the optimal perturbation from this collection. Since our aim is to identify a perturbation candidate that leads to a longer output length, we systematically assess all perturbations within this set and choose the one that yields the maximum output length.





\section{Evaluation}
\label{sec:evaluation}

We evaluate \tool and answer the following research questions.

\begin{itemize}
    \item \textit{\textbf{RQ 2.1 (Severity)}}: How severe will \tool degrade LLMs efficiency?
    \item \textit{\textbf{RQ 2.2 (Effectiveness)}}: How effective is \tool in generating test samples that degrade LLMs efficiency?
    \item \textit{\textbf{RQ 2.3 (Sensitivity)}}: Can \tool generate useful test samples that decrease LLMs efficiency under different LLMs' configurations?
    \item \textit{\textbf{RQ 2.4 (Overheads)}}: What is the overhead of \tool in generating test samples?
    \item \textit{\textbf{RQ 2.5 (Ablation Study)}}: How much does each component in \tool contribute to the overall performance?

\end{itemize}

\subsection{Experimental Setup}

\noindent\textbf{Models and Datasets.} As shown in \tabref{tab:model}, we consider the following nine public LLMs as our evaluation models: Google's T5~\citep{raffel2019exploring}, AllenAI's WMT14 Transformer~\citep{ng2019facebook}, and Helsinki-NLP's H-NLP Translator~\citep{helsinki}, Facebook's Fairseq Transformer \cite{ng2019facebook}, UNICAMP-DL's U-DL Translator \cite{lopes2020lite}, Fine-tuned MarianMT \cite{MarianMT}, Google's FLAN-T5 \cite{chung2022scaling}, Mohamed Bin Zayed University's LaMini-GPT \cite{wu2023lamini} and Salesforce's CodeGen \cite{nijkamp2022codegen}. The first six models are employed for translation tasks, and the subsequent two models are capable of handling various downstream Natural Language Processing tasks. In this paper, our focus is on sentence completion as the subject of investigation. The last model is specialized in code generation.
Individually, T5 is released by Google, which is first pre-trained with multiple language problems, and then fine-tuned on the English-German translation task.
We apply English sentences from dataset \texttt{ZH19} as seed inputs to generate test samples.
AllenAI's WMT14 is one of LLMs from the company AllenAI, which is trained on the WMT19 shared news translation task based on the transformer architecture. We select the WMT14 en-de model as our evaluation model, which is designed for the English-German translation task.
H-NLP is a seq2seq model, where the source language is English and the target language is Chinese. 
For each experimental subject, we randomly select 1,000 inputs from the test dataset as the seed inputs.

To further validate the efficiency loopholes in LLMs for translation, we have additionally chosen three publicly available and high-performing translation LLMs. Fairseq is one of the language
models that Facebook FAIR submitted to the WMT19 shared news translation task, and it’s based on the FFN transformer architecture. We select Fairseq’s en-de model as our
victim model, which is designed for the English-German translation task. U-DL, developed by Natural Language and Deep Learning Process Laboratory of Universidade Estadual de Campinas, is a model built on the T5 architecture and fine-tuned for tasks involving English and Portuguese translation. Marian is a Neural Machine Translation framework, which is mainly developed by the
Microsoft Translator team, and it is released under MIT License. MarianMT Framework's flexibility and efficiency have made it exceptionally popular in the translation field. We choose English-Chinese translator as our evaluation model. 
To ensure experiment consistency, we randomly selected 1,000 English sentences from the ZH19 dataset as seed inputs.

In addition, we selected three open source LLMs for other application scenarios:  Flan-T5 (Encoder-Decoder) instruction-finetune on a collection of data sources using a diverse set of instruction templates. Its performance and ability to generalize to unseen tasks are notably superior to those of the baseline T5 model. LaMini-GPT (Decoder-Only), released by Mohamed Bin Zayed University of Artificial Intelligence, is built on the GPT-2 framework, fine-tuned and distilled with a large-scale instruction data set derived from ChatGPT, all while being more compact and efficient carried out within the structure.
We employ the dataset HellaSwag \cite{zellers2019hellaswag} to assess the sentence completion tasks for the two aforementioned large language models. Likewise, we randomly select 1000 data samples from this dataset as initial seed inputs.
CodeGen, a creation of company Salesforce, is part of the CodeGen family, specializing in autoregressive language models for code generation. Our evaluation of this model involves the utilization of the mbpp dataset \cite{austin2021program}, which comprises 427 Python programming challenges and is a widely recognized benchmark for code generation tasks. It's important to note that this dataset falls into the category of "zero-shot" datasets, as it lacks any input/output demonstrations within its prompts. 
To improve the efficiency of our experiments, we have implemented a modification in the prompt format. In particular, we processed each problem by incorporating a function header and converting the language instructions into function docstrings. Note that this same modification is also used in existing works \cite{cassano2023multipl}.

We select subjects (\ie model, dataset) following policies below.
\begin{itemize}
    \item  \textbf{Availability and Accessibility:} The selected subjects are publicly available, ensuring our research can be widely replicated and expanded upon.

    \item  \textbf{Adoption and Prevalence: }The chosen subjects are widely used across various fields. For example, the H-NLP model had 263,348 downloads on Huggingface in February 2024 ~\citep{helsinki}, Flan-T5 has been cited over 1,300 times \cite{chung2022scaling}, and the MBPP dataset represents the mainstream benchmarks for evaluating code generation models and has gained widespread utilization in prior research \cite{paperswithcode, wang2021codet5, fried2022incoder}.
    
    \item  \textbf{Diversity and Representativeness:} Our selection of datasets and models emphasizes diversity and representativeness across various dimensions. Specifically, for LLMs used in translation tasks, our chosen models have different model architectures, training corpora, translation languages, and training processes. Such a strategic selection underpins the universality and reliability of our results.
    For sentence completion applications, we have chosen flagship models representing the two principal architectures in contemporary text generation: Google's Flan-T5, embodying the encoder-decoder framework, and MBZUAI's LaMini\_GPT, a decoder-only model. Notably, LaMini\_GPT has undergone extensive fine-tuning with high-quality ChatGPT instructions, achieving performance metrics that eclipse those of OpenAI's GPT-2 \cite{wu2023lamini}. In the realm of code generation, our selection includes models from the CodeGen family. Upon its release, CodeGen was recognized as a leading state-of-the-art model in code generation, showcasing remarkable capabilities in automating programming tasks and epitomizing the cutting edge of the field \cite{paperswithcode}.
\end{itemize}

\begin{table}[tbp]
  \centering
  \caption{The LLMs under test in our experiments}
  \resizebox{0.96\textwidth}{!}{
    \begin{tabular}{c|ccccc}
    \toprule
    Model  & Task Category & \texttt{Model\_size} & Vocab Size &  \texttt{Max\_length} &URL \\
        \midrule
        H-NLP & En-Zh Translation & 298 MB & 65,001 & 512   & \href{https://huggingface.co/Helsinki-NLP/opus-mt-en-de}{\url{https://huggingface.co/Helsinki-NLP/opus-mt-en-de}} \\
        AllenAi & En-De Translation & 235 MB & 42,024 & 200   & \href{https://huggingface.co/allenai/wmt16-en-de-dist-12-1}{\url{https://huggingface.co/allenai/wmt16-en-de-dist-12-1}} \\
        T5    & En-Zh Translation & 242 MB & 32,100 & 200   & \href{https://huggingface.co/t5-small}{\url{https://huggingface.co/t5-small}} \\
        U-DL  & En-Pt Translation & 892 MB & 32,128 & 200   & \href{https://huggingface.co/unicamp-dl/translation-en-pt-t5}{\url{https://huggingface.co/unicamp-dl/translation-en-pt-t5}} \\
        FairSeq & En-De Translation & 1.08 GB & 42,024 & 200   & \href{https://huggingface.co/facebook/wmt19-en-de}{\url{https://huggingface.co/facebook/wmt19-en-de}} \\
        MarianMT & En-Zh Translation & 310 MB & 65,001 & 512   & \href{https://huggingface.co/DDDSSS/translation\_en-zh}{\url{https://huggingface.co/DDDSSS/translation\_en-zh}} \\
        Flan-T5 & Sentence Completion & 308 MB & 32,128 & 300   & \href{https://huggingface.co/google/flan-t5-small}{\url{https://huggingface.co/google/flan-t5-small}} \\
        LaMini-GPT & Sentence Completion & 510 MB & 50,258 & 200   & \href{https://huggingface.co/MBZUAI/LaMini-GPT-124M}{\url{https://huggingface.co/MBZUAI/LaMini-GPT-124M}} \\
        CodeGen & Code Generation & 797 MB & 51,200 & 200  & \href{https://huggingface.co/Salesforce/codegen-350M-mono}{\url{https://huggingface.co/Salesforce/codegen-350M-mono}} \\
        \bottomrule
    \end{tabular}%
    }
  \label{tab:model}%
   \vspace{-2mm}
\end{table}%

\noindent\textbf{Comparison Baselines.} A branch of existing works have been proposed for testing LLMs~\cite{nmt_se1,nmt_se2, nmt_se3, nmt_se4, cheng2020seq2sick, BelinkovB18}. However, all of them focus on testing LLMs' correctness.
To the best of our knowledge, we are the first to study LLMs' efficiency degradation issue.
To show that existing correctness testing methods can not generate test inputs that trigger efficiency degradation for LLMs. We compare \tool against four state-of-the-art correctness testing methods, which are designed to generate testing inputs that produce incorrect results.
Specifically, we choose \texttt{SIT}~\cite{nmt_se1}, \texttt{TransRepair}~\cite{nmt_se2}, \texttt{Seq2Sick}~\citep{cheng2020seq2sick}, and \texttt{SynError}~\citep{BelinkovB18} as our comparison baselines. 
\texttt{SIT} proposes a structure-invariant testing method, which is a metamorphic testing approach for validating language models. Given a seed sentence, \texttt{SIT} first generates a list of similar sentences by modifying tokens in the seed sentence. After that, \texttt{SIT} compares the structure of the original outputs and the generated outputs to detect generation errors.
\texttt{TransRepair} is similar to \texttt{SIT}, with the difference that the unperturbed parts of the sentences preserve their adequacy and fluency modulo the mutated tokens. Thus, any perturbed input sentence violating this assumption will be treated as incorrect.
\texttt{Seq2Sick} replaces the tokens in seed inputs to produce adversarial generation outputs that are entirely different from the original outputs.
\texttt{SynError} is a character-level testing method, which minimizes the LLMs's accuracy~(BLEU score) by introducing synthetic noise. Specifically, \texttt{SynError} introduces four character-level perturbations: swap, fully random, and keyboard typos to perturb seed inputs to decrease the BLEU score.

\noindent\textbf{Experimental Procedure.} 
We run \tool in both white-box and black-box settings to test the above-mentioned nine LLMs.
Given a seed input, \tool perturbs the seed input with different types of perturbations.
\tool has one hyper-parameter ($\epsilon$) that is configurable.
In our experiments, we follow existing works \cite{li2018textbugger} and set perturbation size (\ie $\epsilon$) from 1 to 3, representing different degrees of perturbation. 
For RQ1 (severity), we measure the percentage of the average and maximum increased computational resource, in terms of iteration loops, latency, and energy consumption ~(\equref{eq:metric}), due to the generated test inputs compared to the seed inputs. 
For RQ2 (effectiveness), we measure the degradation success ratio~(\equref{eq:validity}), which quantifies the percentage of the test inputs out of all generated by \tool that can degrade the efficiency to a degree that is larger than a pre-defined threshold. 
A higher ratio would imply better efficacy in generating useful test inputs.
For RQ3 (sensitivity), we run \tool on LLMs with different configurations to study whether the efficacy of \tool is sensitive to configurations.
For RQ4 (overheads), we measure the average overheads of running \tool to generate test inputs.
For RQ5 (ablation study), we conduct an ablation study to validate the contribution of each component in \tool.
It is worth noting that, due to the unique nature of code generation tasks, for the evaluation of the CodeGen model, we have made modifications to the stopping criteria. Specifically, we have expanded the list of default EOS  tokens (\ie "\texttt{$<$|endoftext|$>$}", 
"\verb|\ndef|", "\verb|\n#|", "\verb|\nif|" and "\verb|\nclass|"). This method finds widespread application in code generation works\cite{cassano2023multipl, wang2021codet5, fried2022incoder, liu2024your} and proves to be effective in enhancing the efficiency of the model.

\noindent\textbf{Implementation.}
We implement \tool with the \texttt{PyTorch} library, using a server with Intel Xeon E5-26 CPU and eight Nvidia A4500 GPUs. 
For the baseline methods, we implement \texttt{SIT} and \texttt{TransRepair} using the authors' open sourced code~\cite{nmt_se1, RobustNLP}.
We re-implement \texttt{Seq2sick} and \texttt{SynError} according to the corresponding papers as the original implementations are not open sourced.
For LLMs used in our evaluation, we download the pre-trained models using the HuggingFace APIs, and we configure LLMs using both default configurations and varied configurations to answer RQ3.

\subsection{RQ 2.1: Severity}
\label{sec:severity}

\noindent\textbf{Metrics.} 
Our evaluation considers both hardware-independent metrics (\ie number of iteration loops) and hardware-dependent metrics (\ie latency and energy consumption), which quantitatively represent LLMs' efficiency. The number of iteration loops is a widely used hardware-independent metric for measuring software computational efficiency~\cite{weyuker2000experience}. 
In this experiment, the focus is on calculating the number of decoder calls presented in \secref{sec:nmt}, which corresponds to the number of output tokens. Higher decoder calls indicate that LLMs cast more floating-point operations (FLOPs) to handle the input text, which leads
to less efficiency \cite{chen2022nicgslowdown}.
Response latency (\ie the output generation time) and energy consumption are two widely-used hardware-dependent metrics for measuring systems efficiency. Larger latency and energy consumption clearly indicate less efficiency. 
\begin{equation}
\label{eq:metric}
\begin{split}
    & \quad \text{I-Loops} = \frac{\text{Loops}(x') - \text{Loops}(x)}{\text{Loops}(x)}  \times 100\% \\
     & \text{I-Latency} = \frac{\text{Latency}(x') - \text{Latency}(x)}{\text{Latency}(x)}  \times 100\% \\
    & \text{I-Energy} = \frac{\text{Energy}(x') - \text{Energy}(x)}{\text{Energy}(x)}  \times 100\%
\end{split}
\end{equation}
We use I-Loops, I-Latency, and I-Energy to denote the number of iteration loops, response latency, and energy consumption respectively.
The formal definitions of I-Loops, I-Latency, and I-Energy are shown in~\equref{eq:metric}, where $x$ denotes the seed input and $x'$ represents the perturbed input under \tool, Loops$(\cdot)$,  Latency$(\cdot)$ and Energy$(\cdot)$ denote the functions which calculate the average number of iteration loops, latency, and energy consumption, respectively. Larger values of I-Loops, I-Latency, 
and I-Energy indicate a more severe efficiency degradation caused by the test inputs generated under \tool.
In our evaluation, we measure the hardware-dependent efficiency metrics (\ie latency and energy consumption) on two popular hardware platforms: Intel Xeon E5-2660v3 CPU and Nvidia A4500 GPU. For precise measurement of energy consumption on both CPU and GPU, we employ advanced monitoring libraries. Intel's Running Average Power Limit (RAPL) interface is used for the CPU, offering an effective method to observe and manage the power usage of its various components. For the GPU, we utilize Nvidia’s Python Library for NVIDIA Management Library (PyNVML), which serves as a Python wrapper for NVML, enabling accurate tracking and analysis of energy consumption. 
This rigorous methodology allows us to capture comprehensive data on the energy efficiency of these platforms across different operational scenarios, providing critical insights into their performance dynamics and sustainability footprint. Furthermore, to mitigate potential biases introduced by hardware dependencies in the evaluated metrics, we enhanced the reliability and reproducibility of our measurements by averaging the experimental results over three runs.

\begin{table*}[tp]
  \centering
  \caption{The Average Effectiveness Results of \tool in Degrading LLMs Performance}
    \resizebox{\linewidth}{!}{
        \begin{tabular}{llccccccccccccccc}
  
    \toprule[1.2pt]
    \multirow{2}[2]{*}{\textbf{Subject}} & \multirow{2}[2]{*}{\textbf{Methods}} & \multicolumn{3}{c}{\textbf{I-Loops}} & \multicolumn{3}{c}{\textbf{I-Latency(CPU)}} & \multicolumn{3}{c}{\textbf{I-Energy(CPU)}} & \multicolumn{3}{c}{\textbf{I-Latency(GPU)}} & \multicolumn{3}{c}{\textbf{I-Energy(GPU)}} \\
    \cmidrule{3-17}
          &       & \multicolumn{1}{c}
          {$\bm{\epsilon = 1 }$} & \multicolumn{1}{c}
          {$\bm{\epsilon = 2 }$} & \multicolumn{1}{c}
          {$\bm{\epsilon = 3 }$} & \multicolumn{1}{c}
          {$\bm{\epsilon = 1 }$} & \multicolumn{1}{c}
          {$\bm{\epsilon = 2 }$} & \multicolumn{1}{c}
          {$\bm{\epsilon = 3 }$} & \multicolumn{1}{c}
          {$\bm{\epsilon = 1 }$} & \multicolumn{1}{c}
          {$\bm{\epsilon = 2 }$} & \multicolumn{1}{c}
          {$\bm{\epsilon = 3 }$} & \multicolumn{1}{c}
          {$\bm{\epsilon = 1 }$} & \multicolumn{1}{c}
          {$\bm{\epsilon = 2 }$} & \multicolumn{1}{c}
          {$\bm{\epsilon = 3 }$} & \multicolumn{1}{c}
          {$\bm{\epsilon = 1 }$} & \multicolumn{1}{c}
          {$\bm{\epsilon = 2 }$} & \multicolumn{1}{c}
          {$\bm{\epsilon = 3 }$} \\
\midrule
\multirow{10}[3]{*}{\textbf{H-NLP}} & \textbf{\textbf{Seq2Sick}} & 4.31  & 5.84  & 12.28 & 4.83  & 8.85  & 19.55 & 4.84  & 8.85  & 21.47 & 3.73  & 5.90  & 13.24 & 3.77  & 5.96  & 13.33 \\
      & \textbf{SynError} & 19.09 & 19.59 & 19.59 & 19.35 & 19.82 & 19.82 & 19.63 & 20.10 & 20.10 & 14.14 & 14.52 & 14.52 & 14.27 & 14.65 & 14.65 \\
      & \textbf{SIT}   & 11.83 & 5.99  & 5.35  & -1.68 & -8.53 & -11.21 & 8.17  & 6.32  & 7.41  & 9.84  & 5.50  & 5.75  & 9.90  & 5.58  & 5.83 \\
      & \textbf{TransRepair} & 0.17  & 0.17  & 0.17  & 0.76  & 0.10  & 0.10  & 0.93  & 0.33  & 0.33  & -0.07 & 0.00  & 0.00  & -0.07 & 0.00  & 0.00 \\
      & \textbf{\tool(C)} & 564.45 & 995.45 & 1,357.77 & 764.92 & 1,487.92 & 2,015.70 & 785.60 & 1,471.26 & 1,967.05 & 462.24 & 851.80 & 1,116.80 & 406.39 & 755.18 & 972.92 \\
      & \textbf{\tool(T)} & 2,697.77 & 3,735.38 & 3,917.91 & 3,153.97 & 4,481.93 & 4,681.28 & 3,052.62 & 4,544.65 & 4,759.71 & 1,953.57 & 2,729.83 & 2,854.89 & 1,532.91 & 2,137.53 & 2,221.66 \\
      & \textbf{\tool(S)} & 142.31 & 311.06 & 612.08 & 146.51 & 451.93 & 877.79 & 147.70 & 461.30 & 870.72 & 101.21 & 275.58 & 523.04 & 95.05 & 259.88 & 508.80 \\
      & \textbf{\tool-B (C)} & 907.89 & 1,483.58 & 2,032.41 & 815.15 & 1,561.45 & 2,139.55 & 1,026.17 & 1,948.84 & 2,653.57 & 684.34 & 1,330.43 & 1,823.64 & 681.23 & 1,326.60 & 1,821.57 \\
      & \textbf{\tool-B (T)} & 2,556.56 & 3,064.19 & 3,043.95 & 2,557.49 & 3,099.16 & 2,996.62 & 3,106.02 & 3,807.87 & 3,700.27 & 1,968.85 & 2,451.50 & 2,416.83 & 1,965.72 & 2,451.54 & 2,419.52 \\
      & \textbf{\tool-B (S)} & 200.15 & 450.31 & 809.79 & 181.05 & 389.76 & 766.62 & 233.63 & 488.25 & 973.90 & 172.32 & 369.99 & 683.68 & 172.01 & 369.75 & 683.28 \\
\midrule
\multirow{10}[3]{*}{\textbf{AllenAI}} & \textbf{Seq2Sick} & 1.72  & 2.22  & 2.15  & 1.48  & 2.06  & 1.35  & 1.19  & 1.76  & 1.10  & 1.57  & 1.41  & 0.38  & 1.70  & 1.57  & 0.57 \\
      & \textbf{SynError} & 0.38  & 0.38  & 0.38  & 1.89  & 1.89  & 1.89  & 1.75  & 1.75  & 1.75  & -0.85 & -0.85 & -0.85 & -0.71 & -0.71 & -0.71 \\
      & \textbf{SIT}   & 7.06  & 4.12  & 6.67  & 1.73  & -3.24 & -4.64 & 1.73  & -3.24 & -4.60 & 3.95  & 14.25 & -2.05 & 4.12  & 14.64 & -1.60 \\
      & \textbf{TransRepair} & 0.08  & 0.08  & 0.08  & -0.37 & -0.37 & -0.37 & -0.55 & -0.55 & -0.55 & -0.15 & -0.15 & -0.15 & -0.14 & -0.14 & -0.14 \\
      & \textbf{\tool(C)} & 35.16 & 74.90 & 103.36 & 26.69 & 45.77 & 85.09 & 27.48 & 48.09 & 86.00 & 21.82 & 35.43 & 91.48 & 22.12 & 43.21 & 98.46 \\
      & \textbf{\tool(T)} & 24.83 & 42.04 & 56.75 & 49.12 & 62.84 & 67.98 & 49.99 & 62.65 & 69.06 & 30.65 & 41.32 & 46.09 & 31.00 & 41.81 & 49.66 \\
      & \textbf{\tool(S)} & 66.21 & 108.67 & 128.60 & 86.05 & 139.03 & 164.57 & 84.17 & 135.71 & 160.95 & 69.57 & 112.88 & 132.68 & 68.79 & 115.23 & 137.06 \\
      & \textbf{\tool-B (C)} & 31.78 & 84.30 & 116.75 & 225.80 & 935.84 & 143.91 & 43.25 & 132.30 & 194.46 & 27.64 & 97.85 & 148.22 & 27.41 & 97.47 & 147.72 \\
      & \textbf{\tool-B (T)} & 71.37 & 131.54 & 121.88 & 56.40 & 119.95 & 146.99 & 85.25 & 163.81 & 196.13 & 60.21 & 123.44 & 152.83 & 59.93 & 123.10 & 152.54 \\
      & \textbf{\tool-B (S)} & 65.82 & 76.05 & 90.41 & 78.24 & 90.00 & 80.83 & 110.20 & 123.05 & 113.93 & 80.90 & 92.03 & 86.14 & 80.57 & 91.73 & 85.87 \\
\midrule
\multirow{10}[3]{*}{\textbf{T5}} & \textbf{Seq2Sick} & 7.09  & 6.28  & -6.03 & 7.21  & 6.04  & -5.97 & 8.55  & 6.88  & -5.16 & 9.01  & 8.00  & -3.97 & 8.85  & 16.94 & 4.50 \\
      & \textbf{SynError} & 2.18  & 2.18  & 2.18  & 3.20  & 3.20  & 3.20  & 2.11  & 2.11  & 2.11  & 1.02  & 1.02  & 1.02  & 1.13  & 1.13  & 1.13 \\
      & \textbf{SIT}   & -8.06 & 1.05  & 6.27  & -4.51 & 7.79  & 7.38  & -3.79 & 9.84  & 10.59 & -10.99 & 3.57  & 7.74  & -10.90 & 3.78  & 8.07 \\
      & \textbf{TransRepair} & 3.73  & 8.06  & 8.06  & 4.90  & 9.47  & 9.26  & 6.42  & 11.39 & 10.74 & 3.70  & 8.34  & 8.35  & 3.76  & 8.42  & 8.39 \\
      & \textbf{\tool(C)} & 168.92 & 198.36 & 205.37 & 191.05 & 225.48 & 233.01 & 194.45 & 228.02 & 234.04 & 164.61 & 194.79 & 202.28 & 165.38 & 195.77 & 203.29 \\
      & \textbf{\tool(T)} & 307.27 & 328.94 & 328.94 & 352.14 & 376.55 & 376.55 & 347.74 & 373.85 & 373.85 & 305.37 & 325.61 & 325.61 & 331.85 & 352.25 & 352.25 \\
      & \textbf{\tool(S)} & 77.67 & 80.56 & 82.52 & 85.72 & 89.11 & 91.38 & 86.90 & 90.29 & 92.56 & 75.77 & 78.68 & 80.66 & 68.79 & 73.03 & 74.56 \\
      & \textbf{\tool-B (C)} & 231.95 & 255.70 & 259.05 & 239.17 & 257.96 & 259.95 & 279.77 & 303.05 & 305.42 & 233.26 & 257.03 & 261.46 & 233.86 & 257.69 & 262.09 \\
      & \textbf{\tool-B (T)} & 318.94 & 293.67 & 257.92 & 331.77 & 304.63 & 272.73 & 384.44 & 350.53 & 311.98 & 319.07 & 294.36 & 259.99 & 319.65 & 294.84 & 260.43 \\
      & \textbf{\tool-B (S)} & 252.44 & 279.53 & 289.54 & 260.69 & 288.66 & 295.31 & 300.39 & 333.39 & 341.33 & 252.44 & 279.23 & 288.46 & 252.48 & 279.22 & 288.41 \\
\midrule
\multirow{10}[3]{*}{\textbf{U-DL}} & \textbf{Seq2Sick} & 0.59  & 1.22  & 2.71  & -0.30 & 0.97  & 2.47  & 2.70  & 4.07  & 5.80  & 0.96  & 1.72  & 3.17  & 0.98  & 1.74  & 3.20 \\
      & \textbf{SynError} & 0.02  & 0.02  & 0.02  & -0.97 & -0.97 & -0.97 & 2.01  & 2.01  & 2.01  & -0.25 & -0.25 & -0.25 & -0.24 & -0.24 & -0.24 \\
      & \textbf{SIT}   & 16.73 & 6.40  & 4.26  & 15.50 & 4.15  & 2.45  & 18.90 & 9.37  & 7.56  & 17.16 & 6.30  & 4.81  & 17.18 & 6.33  & 4.82 \\
      & \textbf{TransRepair} & 11.36 & 10.66 & 10.82 & 7.53  & 7.09  & 7.09  & 10.07 & 9.94  & 9.95  & 11.46 & 11.45 & 11.59 & 11.43 & 11.41 & 11.55 \\
      & \textbf{\tool(C)} & 258.07 & 390.60 & 469.24 & 261.02 & 405.80 & 494.30 & 288.15 & 439.72 & 532.81 & 253.46 & 383.78 & 461.51 & 253.45 & 383.82 & 461.64 \\
      & \textbf{\tool(T)} & 604.17 & 642.38 & 642.38 & 655.13 & 696.56 & 696.56 & 697.90 & 741.86 & 741.86 & 595.88 & 634.44 & 634.44 & 596.49 & 635.08 & 635.08 \\
      & \textbf{\tool(S)} & 406.92 & 592.52 & 702.89 & 438.23 & 632.64 & 753.88 & 465.04 & 673.76 & 800.25 & 404.42 & 583.65 & 694.40 & 401.74 & 583.99 & 694.84 \\
      & \textbf{\tool-B (C)} & 488.81 & 495.81 & 502.85 & 522.33 & 517.47 & 526.17 & 559.45 & 556.72 & 567.10 & 483.54 & 490.87 & 499.09 & 483.52 & 490.88 & 499.15 \\
      & \textbf{\tool-B (T)} & 502.85 & 502.85 & 494.97 & 536.77 & 527.40 & 511.22 & 579.50 & 570.67 & 557.33 & 498.65 & 499.99 & 496.64 & 498.56 & 499.90 & 496.46 \\
      & \textbf{\tool-B (S)} & 466.97 & 490.19 & 501.44 & 507.14 & 517.67 & 525.86 & 541.40 & 558.12 & 567.17 & 463.23 & 490.47 & 511.64 & 463.31 & 490.69 & 511.80 \\
\midrule
\multirow{10}[3]{*}{\textbf{FairSeq}} & \textbf{Seq2Sick} & 0.61  & 0.64  & -1.33 & 0.46  & -0.75 & -1.48 & 3.60  & 2.91  & 2.04  & 0.05  & 0.15  & -1.25 & 0.08  & 0.20  & -1.20 \\
      & \textbf{SynError} & 0.25  & 0.25  & 0.25  & -2.67 & -2.57 & -2.57 & -0.06 & 0.03  & 0.03  & 0.07  & 0.07  & 0.07  & 0.08  & 0.08  & 0.08 \\
      & \textbf{SIT}   & -1.15 & -2.15 & -1.56 & -4.46 & -7.26 & -6.73 & 0.13  & -1.90 & -1.63 & -2.34 & -1.51 & -2.37 & -2.37 & -1.52 & -2.37 \\
      & \textbf{TransRepair} & 0.26  & 0.25  & 0.25  & 0.02  & 0.01  & 0.05  & 0.67  & 0.66  & 0.69  & -0.03 & -0.01 & -0.01 & -0.03 & 0.00  & 0.00 \\
      & \textbf{\tool(C)} & 22.53 & 37.68 & 59.26 & 15.87 & 29.07 & 49.18 & 20.47 & 34.62 & 55.44 & 18.07 & 31.53 & 51.79 & 18.08 & 31.55 & 51.85 \\
      & \textbf{\tool(T)} & 33.73 & 62.13 & 76.41 & 23.97 & 55.26 & 70.28 & 29.79 & 63.19 & 79.41 & 28.62 & 59.42 & 75.47 & 28.60 & 59.42 & 75.47 \\
      & \textbf{\tool(S)} & 19.42 & 30.87 & 37.82 & 14.01 & 23.67 & 31.31 & 18.23 & 28.59 & 36.73 & 14.84 & 24.87 & 31.68 & 14.86 & 24.91 & 31.72 \\
      & \textbf{\tool-B (C)} & 33.09 & 43.60 & 72.18 & 26.76 & 34.61 & 69.78 & 31.94 & 40.82 & 79.01 & 29.29 & 39.99 & 76.06 & 29.36 & 40.19 & 76.36 \\
      & \textbf{\tool-B (T)} & 41.71 & 66.82 & 87.19 & 32.97 & 58.12 & 75.42 & 39.15 & 66.41 & 85.81 & 36.04 & 64.10 & 85.87 & 36.10 & 64.18 & 85.95 \\
      & \textbf{\tool-B (S)} & 19.04 & 31.29 & 38.51 & 12.24 & 20.73 & 26.06 & 16.94 & 26.55 & 32.40 & 15.71 & 26.25 & 32.80 & 15.72 & 26.29 & 32.85 \\
\midrule
\multirow{10}[3]{*}{\textbf{MarianMT}} & \textbf{Seq2Sick} & -0.06 & -1.78 & -6.61 & -2.69 & -4.85 & -9.87 & 2.07  & 0.05  & -4.26 & 1.12  & -1.24 & -4.58 & 1.09  & -1.26 & -4.60 \\
      & \textbf{SynError} & 3.00  & 3.66  & 3.62  & 1.09  & 1.31  & 1.15  & 3.75  & 4.12  & 3.95  & 1.78  & 2.30  & 2.13  & 1.75  & 2.27  & 2.12 \\
      & \textbf{SIT}   & 1.94  & -0.27 & -0.82 & -1.91 & -3.95 & -5.03 & 3.95  & 2.00  & 0.71  & 1.23  & -0.59 & -1.64 & 1.20  & -0.59 & -1.62 \\
      & \textbf{TransRepair} & 0.03  & 0.95  & 0.68  & -0.89 & -0.65 & -1.06 & 1.50  & 1.77  & 1.36  & 0.06  & 0.24  & -0.02 & 0.05  & 0.23  & -0.03 \\
      & \textbf{\tool(C)} & 54.10 & 113.20 & 222.04 & 41.58 & 102.38 & 226.68 & 51.56 & 119.53 & 274.26 & 52.98 & 113.90 & 210.12 & 52.89 & 113.88 & 210.12 \\
      & \textbf{\tool(T)} & 231.65 & 550.07 & 726.61 & 234.93 & 564.34 & 770.28 & 269.58 & 660.70 & 893.87 & 223.56 & 544.49 & 728.90 & 223.48 & 544.28 & 728.68 \\
      & \textbf{\tool(S)} & 42.77 & 72.19 & 89.17 & 33.89 & 72.33 & 90.33 & 40.27 & 84.89 & 106.84 & 41.21 & 77.28 & 95.68 & 41.19 & 77.21 & 95.61 \\
      & \textbf{\tool-B (C)} & 65.70 & 185.56 & 264.37 & 55.72 & 173.09 & 298.05 & 67.01 & 200.74 & 338.54 & 55.62 & 164.33 & 282.05 & 55.46 & 163.84 & 282.20 \\
      & \textbf{\tool-B (T)} & 223.05 & 517.87 & 722.71 & 229.07 & 580.01 & 752.90 & 260.24 & 660.71 & 860.89 & 199.48 & 508.85 & 674.54 & 199.44 & 509.79 & 675.87 \\
      & \textbf{\tool-B (S)} & 42.43 & 68.25 & 78.04 & 35.36 & 64.01 & 65.18 & 44.70 & 77.00 & 78.26 & 36.10 & 64.71 & 69.38 & 36.06 & 64.67 & 69.41 \\
\midrule
\multirow{10}[3]{*}{\textbf{Flan-T5}} & \textbf{Seq2Sick} & 6.64  & 10.31 & 13.09 & 5.97  & 9.68  & 11.76 & 14.05 & 18.79 & 22.16 & 5.76  & 9.35  & 11.96 & 5.74  & 9.34  & 11.92 \\
      & \textbf{SynError} & 0.05  & 0.05  & 0.05  & -1.73 & -1.73 & -1.73 & 6.02  & 6.02  & 6.02  & 5.62  & 5.62  & 5.62  & -7.51 & -7.51 & -7.51 \\
      & \textbf{SIT}   & 87.52 & 43.57 & 42.95 & 86.40 & 40.87 & 40.36 & 85.17 & 59.87 & 55.88 & 82.96 & 43.18 & 44.38 & 83.07 & 43.33 & 44.51 \\
      & \textbf{TransRepair} & -1.48 & -1.77 & -1.48 & -1.32 & -1.73 & -1.32 & 1.99  & 1.44  & 1.84  & -1.47 & -1.81 & -1.51 & -1.54 & -1.87 & -1.56 \\
      & \textbf{\tool(C)} & 327.55 & 566.82 & 625.69 & 329.99 & 564.27 & 621.78 & 381.37 & 647.48 & 715.92 & 333.91 & 574.26 & 634.28 & 334.03 & 574.53 & 634.51 \\
      & \textbf{\tool(T)} & 1,209.50 & 1,306.26 & 1,349.04 & 1,229.54 & 1,327.42 & 1,372.96 & 1,409.23 & 1,524.04 & 1,578.49 & 1,227.99 & 1,325.01 & 1,368.67 & 1,229.06 & 1,326.23 & 1,369.93 \\
      & \textbf{\tool(S)} & 554.58 & 937.63 & 1,063.39 & 552.96 & 952.73 & 1,087.15 & 637.50 & 1,094.72 & 1,253.29 & 564.39 & 948.81 & 1,076.25 & 564.90 & 949.32 & 1,076.86 \\
      & \textbf{\tool-B (C)} & 421.97 & 628.14 & 850.14 & 426.32 & 629.20 & 845.69 & 495.38 & 726.82 & 980.89 & 426.82 & 637.66 & 897.14 & 426.87 & 638.74 & 868.34 \\
      & \textbf{\tool-B (T)} & 1,242.18 & 1,338.54 & 1,341.69 & 1,240.60 & 1,333.61 & 1,322.22 & 1,422.94 & 1,543.97 & 1,533.38 & 1,256.64 & 1,378.93 & 1,389.51 & 1,258.92 & 1,381.33 & 1,391.86 \\
      & \textbf{\tool-B (S)} & 572.45 & 892.17 & 1,082.76 & 564.03 & 879.97 & 1,064.21 & 652.28 & 1,015.91 & 1,235.05 & 567.71 & 884.32 & 1,076.55 & 568.44 & 885.57 & 1,077.70 \\
\midrule
\multirow{10}[3]{*}{\textbf{LaMini-GPT}} & \textbf{Seq2Sick} & 21.91 & 21.81 & 15.24 & 121.46 & 121.66 & 88.75 & 140.43 & 140.48 & 101.57 & 98.56 & 97.73 & 69.70 & 98.85 & 98.03 & 69.97 \\
      & \textbf{SynError} & 87.81 & 88.68 & 96.68 & 377.07 & 403.58 & 429.18 & 438.06 & 468.04 & 498.53 & 318.71 & 333.48 & 354.60 & 319.06 & 333.86 & 355.09 \\
      & \textbf{SIT}   & 153.80 & 152.34 & 137.53 & 867.02 & 916.92 & 747.65 & 1,014.25 & 1,069.92 & 870.79 & 659.99 & 651.73 & 552.49 & 660.31 & 652.99 & 553.72 \\
      & \textbf{TransRepair} & 29.39 & 26.51 & 26.74 & 144.03 & 140.82 & 140.73 & 168.75 & 163.40 & 164.34 & 130.07 & 119.77 & 130.57 & 130.42 & 120.23 & 131.11 \\
      & \textbf{\tool(C)} & 323.34 & 367.46 & 376.55 & 2,091.80 & 2,643.98 & 2,707.52 & 2,403.58 & 3,098.30 & 3,169.53 & 1,598.12 & 1,997.75 & 2,043.54 & 1,597.50 & 1,995.33 & 2,041.12 \\
      & \textbf{\tool(T)} & 368.67 & 379.73 & 379.73 & 2,539.10 & 2,588.35 & 2,588.35 & 3,066.39 & 3,130.40 & 3,130.40 & 2,106.89 & 2,148.49 & 2,148.49 & 2,104.61 & 2,146.16 & 2,146.16 \\
      & \textbf{\tool(S)} & 347.41 & 366.07 & 366.42 & 2,157.36 & 2,371.21 & 2,372.00 & 2,510.74 & 2,792.73 & 2,793.60 & 1,746.59 & 1,919.43 & 1,920.28 & 1,747.96 & 1,919.32 & 1,920.17 \\
      & \textbf{\tool-B (C)} & 140.67 & 208.14 & 240.05 & 764.30 & 1,109.99 & 1,211.88 & 943.59 & 1,385.40 & 1,525.22 & 733.99 & 1,111.11 & 1,260.97 & 734.45 & 1,112.28 & 1,263.29 \\
      & \textbf{\tool-B (T)} & 242.05 & 232.44 & 246.41 & 1,225.62 & 1,129.45 & 1,223.03 & 1,510.56 & 1,401.18 & 1,513.43 & 1,224.47 & 1,143.59 & 1,249.03 & 1,225.86 & 1,145.94 & 1,252.04 \\
      & \textbf{\tool-B (S)} & 191.80 & 208.14 & 225.11 & 998.56 & 1,069.81 & 1,127.94 & 1,222.32 & 1,328.29 & 1,407.41 & 962.31 & 1,051.01 & 1,138.46 & 962.76 & 1,052.40 & 1,140.30 \\
\midrule
\multirow{10}[3]{*}{\textbf{CodeGen}} & \textbf{Seq2Sick} & 5.10  & 5.10  & 5.10  & 13.34 & 13.34 & 13.34 & 16.68 & 16.68 & 16.68 & 17.06 & 17.06 & 17.06 & 17.10 & 17.10 & 17.10 \\
      & \textbf{SynError} & 26.48 & 26.49 & 27.54 & 83.33 & 83.26 & 85.38 & 89.39 & 89.34 & 91.54 & 87.61 & 87.54 & 89.78 & 87.65 & 87.58 & 89.81 \\
      & \textbf{SIT}   & 54.06 & 41.45 & 44.71 & 202.97 & 145.02 & 158.70 & 213.45 & 153.11 & 168.58 & 223.83 & 161.93 & 181.19 & 223.90 & 162.00 & 181.28 \\
      & \textbf{TransRepair} & 77.64 & 80.47 & 80.84 & 286.84 & 294.79 & 294.88 & 296.86 & 308.83 & 309.09 & 309.52 & 330.97 & 333.95 & 309.68 & 331.24 & 334.22 \\
      & \textbf{\tool(C)} & 109.93 & 139.88 & 168.23 & 321.08 & 434.06 & 533.72 & 336.01 & 453.64 & 558.45 & 351.20 & 478.08 & 592.39 & 351.46 & 478.36 & 592.75 \\
      & \textbf{\tool(T)} & 182.42 & 182.42 & 182.42 & 578.30 & 578.30 & 578.30 & 602.68 & 602.68 & 602.68 & 639.91 & 639.91 & 639.91 & 640.26 & 640.26 & 640.26 \\
      & \textbf{\tool(S)} & 176.30 & 187.59 & 187.59 & 575.42 & 615.61 & 615.61 & 593.11 & 635.62 & 635.62 & 607.64 & 653.51 & 653.51 & 607.97 & 653.91 & 653.91 \\
      & \textbf{\tool-B (C)} & 147.75 & 170.76 & 175.70 & 463.20 & 518.89 & 523.96 & 481.57 & 540.51 & 549.87 & 501.99 & 582.24 & 603.35 & 501.99 & 582.47 & 603.57 \\
      & \textbf{\tool-B (T)} & 118.40 & 154.85 & 146.01 & 378.35 & 496.82 & 451.94 & 394.97 & 517.96 & 475.82 & 405.33 & 551.65 & 528.71 & 405.63 & 552.12 & 529.16 \\
      & \textbf{\tool-B (S)} & 152.02 & 166.56 & 158.82 & 488.73 & 523.34 & 494.20 & 507.41 & 547.59 & 519.25 & 521.24 & 578.62 & 563.35 & 521.44 & 578.88 & 563.62 \\
    \bottomrule[1.2pt]
    \end{tabular}%
    }
  \label{tab:severity_all}%
\end{table*}%

\begin{table*}[tp]
  \centering
  \caption{The Maximum Effectiveness Results of \tool in Degrading LLMs Performance}
    \resizebox{\linewidth}{!}{
        \begin{tabular}{llccccccccccccccc}
  
    \toprule[1.2pt]
    \multirow{2}[2]{*}{\textbf{Subject}} & \multirow{2}[2]{*}{\textbf{Methods}} & \multicolumn{3}{c}{\textbf{I-Loops}} & \multicolumn{3}{c}{\textbf{I-Latency(CPU)}} & \multicolumn{3}{c}{\textbf{I-Energy(CPU)}} & \multicolumn{3}{c}{\textbf{I-Latency(GPU)}} & \multicolumn{3}{c}{\textbf{I-Energy(GPU)}} \\
    \cmidrule{3-17}
          &       & \multicolumn{1}{c}
          {$\bm{\epsilon = 1 }$} & \multicolumn{1}{c}
          {$\bm{\epsilon = 2 }$} & \multicolumn{1}{c}
          {$\bm{\epsilon = 3 }$} & \multicolumn{1}{c}
          {$\bm{\epsilon = 1 }$} & \multicolumn{1}{c}
          {$\bm{\epsilon = 2 }$} & \multicolumn{1}{c}
          {$\bm{\epsilon = 3 }$} & \multicolumn{1}{c}
          {$\bm{\epsilon = 1 }$} & \multicolumn{1}{c}
          {$\bm{\epsilon = 2 }$} & \multicolumn{1}{c}
          {$\bm{\epsilon = 3 }$} & \multicolumn{1}{c}
          {$\bm{\epsilon = 1 }$} & \multicolumn{1}{c}
          {$\bm{\epsilon = 2 }$} & \multicolumn{1}{c}
          {$\bm{\epsilon = 3 }$} & \multicolumn{1}{c}
          {$\bm{\epsilon = 1 }$} & \multicolumn{1}{c}
          {$\bm{\epsilon = 2 }$} & \multicolumn{1}{c}
          {$\bm{\epsilon = 3 }$} \\
\midrule
\multirow{10}[4]{*}{\textbf{H-NLP}} & \textbf{\textbf{Seq2Sick}} & 1,922.22 & 1,922.22 & 2,000.00 & 2,557.01 & 2,557.01 & 2,557.01 & 3,081.88 & 3,081.88 & 3,081.88 & 2,031.63 & 2,031.63 & 2,031.63 & 2,035.46 & 2,035.46 & 2,035.46 \\
      & \textbf{SynError} & 975.00   & 975.00   & 975.00   & 398.24 & 398.24 & 398.24 & 492.18 & 492.18 & 492.18 & 397.80 & 397.80 & 397.80 & 396.97 & 396.97 & 396.97 \\
      & \textbf{SIT} & 1,416.67 & 1,666.67 & 540.00   & 2,100.68 & 1,383.50 & 361.29 & 2,541.75 & 1,675.15 & 474.76 & 1,876.29 & 1,251.05 & 388.86 & 1,872.66 & 1,249.04 & 387.36 \\
      & \textbf{TransRepair} & 66.67 & 81.56 & 81.56 & 36.92 & 32.87 & 86.91 & 53.98 & 57.88 & 120.45 & 45.82 & 52.88 & 69.05 & 45.78 & 52.57 & 68.90 \\
      & \textbf{\tool(C)} & 8,433.33 & 8,433.33 & 8,433.33 & 7,969.98 & 7,969.98 & 7,969.98 & 10,554.40 & 10,554.40 & 10,554.40 & 7,691.45 & 7,691.45 & 7,691.45 & 7,673.37 & 7,673.37 & 7,673.37 \\
      & \textbf{\tool(T)} & 12,700.00 & 12,700.00 & 12,700.00 & 8,258.55 & 8,258.55 & 8,258.55 & 10,486.75 & 10,486.75 & 10,486.75 & 5,990.07 & 5,990.07 & 5,990.07 & 5,984.31 & 5,984.31 & 5,984.31 \\
      & \textbf{\tool(S)} & 3,100.00 & 6,300.00 & 8,433.33 & 4,162.90 & 6,704.67 & 7,357.19 & 5,193.61 & 8,286.11 & 8,810.90 & 2,945.15 & 5,305.42 & 5,305.42 & 2,944.76 & 5,292.81 & 5,292.81 \\
      & \textbf{\tool-B (C)} & 10,140.00 & 12,700.00 & 10,140.00 & 8,456.70 & 8,220.03 & 8,240.21 & 11,418.85 & 11,059.20 & 11,091.90 & 6,244.01 & 6,231.62 & 7,711.93 & 6,180.36 & 6,214.44 & 7,697.90 \\
      & \textbf{\tool-B (T)} & 10,140.00 & 10,140.00 & 10,140.00 & 9,579.03 & 9,388.97 & 9,286.91 & 12,430.16 & 12,355.92 & 12,241.07 & 6,483.82 & 6,480.49 & 6,717.57 & 6,483.59 & 6,487.76 & 6,700.88 \\
      & \textbf{\tool-B (S)} & 3,100.00 & 6,300.00 & 12,700.00 & 3,599.50 & 5,417.91 & 9,132.01 & 4,452.36 & 6,740.22 & 12,109.99 & 3,513.36 & 4,508.14 & 6,781.85 & 3,507.85 & 4,512.49 & 6,767.54 \\
\midrule
\multirow{10}[4]{*}{\textbf{AllenAI}} & \textbf{Seq2Sick} & 33.33 & 33.33 & 45.16 & 17.98 & 20.79 & 28.50 & 46.67 & 50.97 & 60.87 & 13.57 & 15.48 & 41.13 & 13.28 & 15.44 & 41.08 \\
      & \textbf{SynError} & 12.81 & 12.81 & 12.81 & 9.89  & 9.89  & 9.89  & 31.33 & 31.33 & 31.33 & 12.27 & 12.27 & 12.27 & 12.21 & 12.21 & 12.21 \\
      & \textbf{SIT}   & 23.81 & 28.57 & 28.57 & 34.08 & 26.92 & 16.56 & 63.45 & 46.50  & 43.62 & 25.41 & 15.28 & 13.28 & 25.3  & 15.22 & 13.24 \\
      & \textbf{TransRepair} & 14.29 & 14.29 & 14.29 & 4.34  & 4.34  & 4.34  & 27.52 & 27.52 & 27.52 & 8.95  & 8.95  & 8.95  & 8.91  & 8.91  & 8.91 \\
      & \textbf{\tool(C)} & 93.33 & 247.37 & 440.54 & 128.38 & 140.21 & 332.09 & 165.37 & 194.69 & 442.04 & 172.72 & 202.17 & 368.06 & 172.1 & 201.49 & 368.99 \\
      & \textbf{\tool(T)} & 173.33 & 285.71 & 683.33 & 83.94 & 266.41 & 611.60 & 119.43 & 336.99 & 789.79 & 90.91 & 283.61 & 602.78 & 90.85 & 282.94 & 603.10 \\
      & \textbf{\tool(S)} & 175.00   & 414.29 & 589.66 & 142.19 & 292.91 & 573.46 & 182.65 & 386.68 & 703.88 & 160.93 & 377.36 & 538.89 & 161.42 & 377.22 & 539.05 \\
      & \textbf{\tool-B (C)} & 157.58 & 852.38 & 852.38 & 184.80 & 1,021.41 & 1,021.18 & 269.93 & 1,357.36 & 1,349.91 & 212.59 & 1,052.79 & 1,023.75 & 211.45 & 1,047.56 & 1,018.38 \\
      & \textbf{\tool-B (T)} & 1,233.33 & 1,233.33 & 900.00   & 1,032.59 & 996.48 & 846.78 & 1,394.01 & 1,356.31 & 1,130.11 & 1,109.14 & 1,074.40 & 798.5 & 1,100.56 & 1,066.07 & 796.15 \\
      & \textbf{\tool-B (S)} & 761.54 & 1,233.33 & 1,186.67 & 920.27 & 756.81 & 729.93 & 1,193.81 & 958.69 & 954.42 & 891.05 & 851.72 & 852.57 & 889.57 & 848.38 & 849.46 \\
\midrule
\multirow{10}[4]{*}{\textbf{T5}} & \textbf{Seq2Sick} & 90.91  & 121.15  & 247.62  & 87.43  & 130.42  & 252.47  & 127.68  & 156.29  & 294.40  & 95.72  & 118.58  & 248.67  & 95.87  & 118.50  & 249.29  \\
      & \textbf{SynError} & 61.54  & 61.54  & 61.54  & 63.81  & 63.81  & 63.81  & 70.27  & 70.27  & 70.27  & 61.74  & 61.74  & 61.74  & 61.75  & 61.75  & 61.75  \\
      & \textbf{SIT}    & 945.45  & 576.47  & 576.47  & 1,063.35  & 565.74  & 608.96  & 1,323.72  & 666.54  & 714.72  & 961.39  & 570.17  & 567.80  & 963.22  & 571.87  & 568.55  \\
      & \textbf{TransRepair} & 422.73  & 422.73  & 422.73  & 439.91  & 439.91  & 439.91  & 493.85  & 493.85  & 493.85  & 425.92  & 425.92  & 425.92  & 425.95  & 425.95  & 425.95  \\
      & \textbf{\tool(C)} & 945.45  & 945.45  & 945.45  & 1,470.96  & 1,470.96  & 1,470.96  & 1,751.91  & 1,751.91  & 1,751.91  & 885.92  & 885.92  & 885.92  & 879.21  & 879.21  & 879.21  \\
      & \textbf{\tool(T)} & 1,816.67  & 1,816.67  & 1,816.67  & 1,829.31  & 1,829.31  & 1,829.31  & 2,443.06  & 2,443.06  & 2,443.06  & 1,796.62  & 1,796.62  & 1,796.62  & 1,798.34  & 1,798.34  & 1,798.34  \\
      & \textbf{\tool(S)} & 945.45  & 945.45  & 945.45  & 1,010.14  & 1,010.14  & 1,010.14  & 1,271.41  & 1,271.41  & 1,271.41  & 963.27  & 963.27  & 963.27  & 964.11  & 964.11  & 964.11  \\
      & \textbf{\tool-B (C)}  & 1,816.67  & 1,816.67  & 1,816.67  & 1,898.90  & 1,837.56  & 1,920.93  & 2,500.77  & 2,524.75  & 2,598.89  & 1,835.70  & 1,831.20  & 1,824.22  & 1,840.98  & 1,836.47  & 1,828.48  \\
      & \textbf{\tool-B (T)} & 1,816.67  & 945.45  & 784.62  & 1,892.24  & 1,000.72  & 866.48  & 2,569.88  & 1,272.44  & 1,060.74  & 1,783.49  & 962.94  & 785.66  & 1,786.73  & 964.82  & 787.57  \\
      & \textbf{\tool-B (S)} & 945.45  & 945.45  & 945.45  & 992.67  & 1,001.37  & 992.03  & 1,259.61  & 1,276.53  & 1,264.36  & 958.68  & 960.53  & 959.96  & 960.48  & 961.78  & 961.22  \\
\midrule
\multirow{10}[4]{*}{\textbf{U-DL}} & \textbf{Seq2Sick} & 27.03  & 36.84  & 75.76  & 27.75  & 52.99  & 90.40  & 31.39  & 56.60  & 91.85  & 28.21  & 39.38  & 74.23  & 28.30  & 39.58  & 74.31  \\
  & \textbf{SynError} & 15.79  & 15.79  & 15.79  & 12.12  & 12.12  & 12.12  & 17.72  & 17.72  & 17.72  & 16.95  & 16.95  & 16.95  & 16.85  & 16.85  & 16.85  \\
  & \textbf{SIT}   & 545.16  & 327.27  & 327.27  & 579.53  & 361.57  & 379.32  & 416.79  & 377.14  & 393.91  & 553.80  & 323.51  & 323.51  & 553.80  & 322.16  & 322.16  \\
  & \textbf{TransRepair} & 850.00  & 850.00  & 850.00  & 523.63  & 566.02  & 566.02  & 612.79  & 686.76  & 686.76  & 855.37  & 937.09  & 937.09  & 852.58  & 934.06  & 934.06  \\
   & \textbf{\tool(C)} & 4,900.00  & 4,900.00  & 4,900.00  & 4,065.42  & 4,065.42  & 4,065.42  & 5041.52  & 5041.52  & 5,041.52  & 4,692.21  & 4,692.21  & 4,692.21  & 4,663.56  & 4,663.56  & 4,663.56  \\
  & \textbf{\tool(T)} & 4,900.00  & 4,900.00  & 4,900.00  & 3,978.94  & 3,978.94  & 3,978.94  & 4,831.58  & 4,831.58  & 4,831.58  & 4,448.49  & 4,448.49  & 4,448.49  & 4,444.52  & 4,444.52  & 4,444.52  \\
  & \textbf{\tool(S)} & 1,718.18  & 4,900.00  & 4,900.00  & 1,984.05  & 3,844.74  & 3,844.74  & 2,176.62  & 4,694.89  & 4,694.89  & 1,720.49  & 4,566.83  & 4,566.83  & 1,722.57  & 4557.84  & 4,557.84  \\
   & \textbf{\tool-B (C)} & 4,900.00  & 4,900.00  & 4,900.00  & 4,976.52  & 4,979.55  & 4,902.19  & 6,293.19  & 6,183.34  & 6,153.30  & 4,889.23  & 4,884.94  & 4,823.49  & 4,826.34  & 4,893.77  & 4,916.49  \\
  & \textbf{\tool-B (T)} & 4,900.00  & 4,900.00  & 4,900.00  & 4,961.40  & 4,942.95  & 4,899.74  & 6,161.10  & 5,931.57  & 6,154.24  & 4,858.21  & 4,884.94  & 4,887.12  & 4,820.22  & 4,847.27  & 4,848.92  \\
   & \textbf{\tool-B (S)} & 4,900.00  & 4,900.00  & 4,900.00  & 5,120.57  & 4,991.13  & 4,945.80  & 6,359.07  & 6,255.39  & 6,063.02  & 4,825.21  & 4,876.52  & 4,952.21  & 4,819.50  & 4,878.73  & 4,952.97  \\
\midrule
\multirow{10}[4]{*}{\textbf{FairSeq}} & \textbf{Seq2Sick} & 39.13  & 39.13  & 39.13  & 65.76  & 65.76  & 65.76  & 68.53  & 68.53  & 68.53  & 42.23  & 42.23  & 42.23  & 42.19  & 42.19  & 42.19  \\
 & \textbf{SynError} & 10.00  & 10.00  & 10.00  & 19.33  & 19.33  & 19.33  & 15.98  & 15.98  & 15.98  & 12.53  & 12.53  & 12.53  & 12.67  & 12.67  & 12.67  \\
  & \textbf{SIT}   & 37.50  & 37.50  & 37.50  & 64.17  & 36.24  & 49.49  & 73.91  & 40.20  & 50.67  & 20.97  & 27.18  & 21.13  & 20.90  & 26.82  & 21.10  \\
 & \textbf{TransRepair} & 13.16  & 13.16  & 13.16  & 22.00  & 22.00  & 22.00  & 18.89  & 18.89  & 18.89  & 7.36  & 7.27  & 7.27  & 7.36  & 7.25  & 7.25  \\
  & \textbf{\tool(C)} & 80.00  & 150.00  & 589.66  & 75.28  & 89.34  & 371.87  & 91.66  & 101.22  & 402.55  & 81.85  & 90.86  & 403.03  & 81.70  & 90.73  & 404.14  \\
   & \textbf{\tool(T)} & 769.57  & 1,011.11  & 1,011.11  & 576.82  & 1,071.13  & 1,071.13  & 661.12  & 1,157.19  & 1,157.19  & 713.87  & 1,030.73  & 1,030.73  & 713.45  & 1,030.07  & 1,030.07  \\
 & \textbf{\tool(S)} & 62.50  & 100.00  & 125.00  & 51.89  & 78.08  & 86.72  & 59.25  & 87.21  & 107.37  & 58.85  & 76.68  & 100.38  & 58.56  & 76.63  & 100.20  \\
  & \textbf{\tool-B (C)} & 769.57  & 386.96  & 818.75  & 739.45  & 469.90  & 1094.49  & 796.85  & 513.22  & 1179.19  & 777.23  & 542.11  & 1007.78  & 777.18  & 542.04  & 1012.06  \\
  & \textbf{\tool-B (T)} & 769.57  & 1,011.11  & 1,328.57  & 751.56  & 1,074.00  & 1,322.30  & 836.38  & 1,140.75  & 1,434.50  & 731.03  & 1,005.78  & 1,230.04  & 733.89  & 1,004.80  & 1,229.26  \\
   & \textbf{\tool-B (S)} & 62.50  & 100.00  & 125.00  & 55.40  & 74.50  & 92.28  & 62.82  & 80.25  & 103.05  & 56.03  & 75.48  & 84.32  & 56.00  & 75.43  & 84.30  \\
\midrule
\multirow{10}[4]{*}{\textbf{MarianMT}} & \textbf{Seq2Sick} & 57.69  & 56.52  & 36.36  & 56.00  & 51.82  & 39.27  & 76.41  & 65.69  & 48.38  & 74.30  & 74.30  & 87.19  & 74.06  & 74.06  & 87.02  \\
  & \textbf{SynError} & 44.44  & 66.67  & 71.05  & 40.48  & 67.97  & 70.53  & 51.09  & 79.28  & 79.28  & 43.19  & 63.05  & 63.96  & 43.18  & 63.04  & 64.56  \\
  & \textbf{SIT}   & 77.78  & 65.00  & 60.87  & 54.35  & 36.79  & 53.29  & 61.71  & 50.24  & 62.93  & 50.84  & 32.85  & 59.25  & 50.72  & 32.84  & 59.20  \\
    & \textbf{TransRepair} & 22.22  & 35.71  & 34.78  & 34.95  & 32.78  & 28.83  & 40.93  & 40.95  & 40.95  & 43.33  & 33.71  & 32.67  & 43.29  & 33.71  & 32.48  \\
  & \textbf{\tool(C)} & 200.00  & 1,869.23  & 6,300.00  & 228.97  & 2,323.96  & 7,306.69  & 276.90  & 2,699.20  & 9,529.47  & 256.83  & 1,664.78  & 4,687.01  & 256.63  & 1,669.21  & 4,685.35  \\
  & \textbf{\tool(T)} & 3,557.14  & 6,300.00  & 6,300.00  & 3,877.44  & 5,235.32  & 5,235.32  & 4,489.69  & 6,768.05  & 6,768.05  & 3,459.49  & 5,290.32  & 5,290.32  & 3,452.27  & 5,280.51  & 5,280.51  \\
   & \textbf{\tool(S)} & 757.89  & 926.67  & 926.67  & 490.25  & 1,673.06  & 1,673.06  & 429.09  & 1,945.97  & 1,945.97  & 561.70  & 1,336.00  & 1,336.00  & 562.66  & 1,336.00  & 1,336.00  \\
   & \textbf{\tool-B (C)} & 225.00  & 4,166.67  & 2,594.74  & 247.93  & 3,995.95  & 2,619.07  & 290.05  & 4,728.15  & 3,003.46  & 232.88  & 3,211.10  & 2,563.35  & 230.91  & 3,196.36  & 2,559.47  \\
  & \textbf{\tool-B (T)} & 3,313.33  & 4,554.55  & 4,554.55  & 3,292.26  & 4,630.19  & 4,953.01  & 3,767.91  & 5,451.64  & 6,056.81  & 2,905.84  & 4,062.96  & 4,062.26  & 2,905.01  & 4,072.67  & 4,073.25  \\
   & \textbf{\tool-B (S)} & 757.89  & 926.67  & 752.63  & 728.78  & 1,593.17  & 710.01  & 812.40  & 1,798.47  & 786.78  & 599.04  & 1,379.27  & 618.82  & 597.73  & 1,376.98  & 621.54  \\
\midrule
\multirow{10}[4]{*}{\textbf{Flan-T5}} & \textbf{Seq2Sick} & 283.33  & 283.33  & 283.33  & 250.63  & 250.63  & 250.63  & 299.03  & 299.03  & 299.03  & 274.07  & 274.07  & 274.07  & 275.03  & 275.03  & 275.03  \\
  & \textbf{SynError} & 18.18  & 18.18  & 18.18  & 19.95  & 19.95  & 19.95  & 30.01  & 30.01  & 30.01  & 31.38  & 31.38  & 31.38  & 31.65  & 31.65  & 31.65  \\
 & \textbf{SIT}   & 1328.57  & 1,233.33  & 1,233.33  & 1,288.39  & 1,197.20  & 1,271.50  & 1,440.45  & 1,344.44  & 1,443.66  & 1,433.72  & 1,255.22  & 1,254.68  & 1,437.51  & 1,257.34  & 1,256.80  \\
  & \textbf{TransRepair} & 108.33  & 108.33  & 108.33  & 70.30  & 89.02  & 89.02  & 81.55  & 94.70  & 94.70  & 106.82  & 106.82  & 106.82  & 105.81  & 105.81  & 105.81  \\
   & \textbf{\tool(C)} & 2,400.00  & 2,400.00  & 2,400.00  & 2,454.15  & 2,454.15  & 2,454.15  & 2,989.28  & 2,989.28  & 2,989.28  & 2,508.12  & 2,508.12  & 2,508.12  & 2,507.51  & 2,507.51  & 2,507.51  \\
   & \textbf{\tool(T)} & 2,400.00  & 2,400.00  & 2,757.14  & 2,481.39  & 2,481.39  & 2,904.22  & 3,086.71  & 3,086.71  & 3,691.62  & 2,477.39  & 2,477.39  & 2,826.32  & 2,482.17  & 2,482.17  & 2,830.94  \\
  & \textbf{\tool(S)} & 2,400.00  & 2,400.00  & 2,400.00  & 2,396.28  & 2,917.05  & 2,917.05  & 2,998.45  & 3,444.31  & 3,444.31  & 2,496.46  & 2,496.46  & 2,496.46  & 2,498.29  & 2,498.29  & 2,498.29  \\
  & \textbf{\tool-B (C)} & 2,400.00  & 2,400.00  & 2,400.00  & 2,138.76  & 2,158.41  & 2,369.58  & 2,561.77  & 2,596.17  & 2,972.43  & 2,404.92  & 2,393.24  & 2,778.14  & 2,405.80  & 2,403.45  & 2,776.39  \\
  & \textbf{\tool-B (T)} & 2,400.00  & 2,400.00  & 2,757.14  & 2,369.71  & 2,661.01  & 2,710.52  & 2,950.62  & 3,290.37  & 3,424.51  & 2,569.68  & 3,145.83  & 3,212.41  & 2,572.49  & 3,151.03  & 3,210.04  \\
   & \textbf{\tool-B (S)} & 2,400.00  & 2,400.00  & 2,400.00  & 2,416.88  & 2,480.45  & 2,538.83  & 3,006.46  & 3,072.52  & 3,167.72  & 2,440.13  & 2,450.42  & 2,442.96  & 2,444.45  & 2,454.10  & 2,447.29  \\
\midrule
\multirow{10}[4]{*}{\textbf{LaMini-GPT}} & \textbf{Seq2Sick} & 669.23  & 669.23  & 426.32  & 3,856.26  & 3,856.26  & 3,856.26  & 4,367.19  & 4,367.19  & 4,367.19  & 3,019.68  & 3,019.68  & 3,019.68  & 3,028.41  & 3,028.41  & 3,028.41  \\
  & \textbf{SynError} & 769.57  & 769.57  & 769.57  & 3,478.30  & 3,718.72  & 3,718.72  & 4,012.68  & 4,211.54  & 4,211.54  & 2,824.70  & 2,824.70  & 2,824.70  & 2,822.29  & 2,822.29  & 2,822.29  \\
   & \textbf{SIT}   & 952.63  & 952.63  & 952.63  & 6,166.94  & 7,827.18  & 7,934.35  & 8,319.36  & 9,972.64  & 10,092.55  & 5,200.85  & 5,214.04  & 5,231.60  & 5,181.36  & ,5206.77  & 5,224.30  \\
   & \textbf{TransRepair} & 525.00  & 525.00  & 525.00  & 3,933.96  & 3,933.96  & 3,933.96  & 4,394.69  & 4,394.69  & 4,394.69  & 3,053.31  & 3,053.31  & 3,053.31  & 3,072.67  & 3,072.67  & 3,072.67  \\
   & \textbf{\tool(C)} & 952.63  & 1,076.47  & 1,076.47  & 8,352.47  & 18,318.66  & 18,318.66  & 10,466.74  & 25,835.15  & 25,835.15  & 6,769.64  & 15,538.18  & 15,538.18  & 6,752.79  & 15,377.14  & 15,377.14  \\
    & \textbf{\tool(T)} & 952.63  & 1,076.47  & 1,076.47  & 18,081.36  & 18,081.36  & 18,081.36  & 25,469.76  & 25,469.76  & 25,469.76  & 15,373.20  & 15,373.20  & 15,373.20  & 15,204.52  & 15,204.52  & 15,204.52  \\
 & \textbf{\tool(S)} & 852.38  & 852.38  & 852.38  & 12,388.72  & 14,396.41  & 14,396.41  & 14,680.90  & 20,952.22  & 20,952.22  & 6,434.13  & 14,818.49  & 14,818.49  & 6,432.25  & 14,664.10  & 14,664.10  \\
  & \textbf{\tool-B (C)} & 471.43  & 525.00  & 525.00  & 3,720.21  & 4,762.53  & 4,658.27  & 4,554.85  & 6,175.32  & 6,084.00  & 3,267.67  & 5,363.85  & 5,350.09  & 3,267.87  & 5,345.96  & 5,338.68  \\
  & \textbf{\tool-B (T)} & 525.00  & 525.00  & 525.00  & 4,582.59  & 4,235.27  & 5,073.44  & 6,016.12  & 5,630.82  & 6,404.08  & 5,302.61  & 4,055.18  & 5,344.19  & 5,280.60  & 4,053.63  & 5,337.63  \\
  & \textbf{\tool-B (S)} & 669.23  & 669.23  & 669.23  & 4,243.52  & 4,533.69  & 4,061.00  & 5,612.34  & 6,003.08  & 5,566.46  & 4,338.83  & 4,335.70  & 4,354.52  & 4,330.33  & 4,327.21  & 4345.99  \\
\midrule
\multirow{10}[4]{*}{\textbf{CodeGen}} & \textbf{Seq2Sick} & 189.86  & 189.86  & 189.86  & 583.95  & 583.95  & 583.95  & 610.69  & 610.69  & 610.69  & 540.78  & 540.78  & 540.78  & 541.35  & 541.35  & 541.35  \\
    & \textbf{SynError} & 334.78  & 334.78  & 334.78  & 1,492.46  & 1,492.46  & 1,492.46  & 1,582.08  & 1,582.08  & 1,582.08  & 1,571.08  & 1,571.08  & 1,571.08  & 1,568.21  & 1,568.21  & 1,568.21  \\
    & \textbf{SIT}   & 334.78  & 334.78  & 334.78  & 1,732.38  & 1,558.46  & 1,717.67  & 1,840.58  & 1,649.79  & 1,828.54  & 1,854.63  & 1,788.77  & 2,056.57  & 1,852.92  & 1,789.52  & 2,059.23  \\
  & \textbf{TransRepair} & 334.78  & 334.78  & 334.78  & 1,575.88  & 1,577.57  & 1,577.57  & 1,659.61  & 1,666.67  & 1,666.67  & 1,860.50  & 1,860.50  & 1,860.50  & 1,860.02  & 1,860.02  & 1,860.02  \\
   & \textbf{\tool(C)} & 354.55  & 354.55  & 354.55  & 1,742.25  & 1,739.49  & 1,685.36  & 1,843.62  & 1,841.95  & 1,843.32  & 1815.81  & 1,815.81  & 1,815.81  & 1,827.49  & 1,827.49  & 1,827.49  \\
  & \textbf{\tool(T)} & 354.55  & 354.55  & 354.55  & 1,736.78  & 1,736.78  & 1,736.78  & 1,843.88  & 1,843.88  & 1,843.88  & 1,820.21  & 1,820.21  & 1,820.21  & 1,829.98  & 1,829.98  & 1,829.98  \\
   & \textbf{\tool(S)} & 354.55  & 354.55  & 354.55  & 1,745.30  & 1,745.30  & 1,745.30  & 1,843.20  & 1,843.20  & 1,843.20  & 1,843.31  & 1,843.31  & 1,843.31  & 1,845.11  & 1,845.11  & 1,845.11  \\
  & \textbf{\tool-B (C)} & 354.55  & 354.55  & 354.55  & 1,737.79  & 1,727.37  & 1,716.29  & 1,833.57  & 1,829.78  & 1,823.24  & 1,761.06  & 1,764.48  & 1,759.02  & 1,762.84  & 1,772.37  & 1,760.79  \\
    & \textbf{\tool-B (T)} & 354.55  & 325.53  & 354.55  & 1,738.85  & 1,724.79  & 1,743.39  & 1,839.45  & 1,829.15  & 1,855.67  & 1,891.48  & 1,904.63  & 1,799.27  & 1,894.36  & 1,906.88  & 1,807.42  \\
  & \textbf{\tool-B (S)} & 354.55  & 354.55  & 354.55  & 1,749.17  & 1,740.43  & 1,652.96  & 1,844.60  & 18,44.65  & 1,765.17  & 1,765.54  & 1,761.92  & 1,765.10  & 1,772.19  & 1,768.46  & 1,771.65  \\
    \bottomrule[1.2pt]
     \end{tabular}
    }
  \label{tab:severity_max}%
\end{table*}%

\begin{table}[htp]
     \caption{The examples of test samples generated by \tool}
  \resizebox{0.98\textwidth}{!}{
\begin{tabular}{lll}
\toprule
\multicolumn{1}{l}{\textbf{Subject}} & \multicolumn{1}{l}{\textbf{Methods}} & \multicolumn{1}{l}{\textbf{Test samples}} \\
\midrule
\multicolumn{1}{l}{\multirow{7}{*}{\textbf{FairSeq}}} & \textbf{Original} & women over 55 are pickier about their partners than at any other time in their lives. \\
\cmidrule{2-3}
       &  \textbf{\tool (C)} & women over 55 are \textcolor{red}{5}pickier about their partners than at any other time in their lives. \\
       &  \textbf{\tool (T)} & women \textcolor{red}{dinger} 55 are pickier about their partners than at any other time in their lives. \\
       &  \textbf{\tool (S)} & women over 55 are pickier \textcolor{red}{because} their partners than at any other time in their lives. \\
       &  \textbf{\tool-B (C)} & women \textcolor{red}{G}over 55 are pickier about their partners than at any other time in their lives. \\
       &  \textbf{\tool-B (T)} & \textcolor{red}{structures} over 55 are pickier about their partners than at any other time in their lives. \\
       &  \textbf{\tool-B (S)} & \textcolor{red}{research} over 55 are pickier about their partners than at any other time in their lives. \\
\midrule
\multicolumn{1}{l}{\multirow{7}{*}{\textbf{Flan-T5}}} & \textbf{Original} & A woman is sitting at a table in a fast food restaurant while eating. She continually speaks to nobody as she eats. she \\
\cmidrule{2-3}
       &  \textbf{\tool (C)} & A woman is sitting at a table in a fast food restaurant while \textcolor{red}{\_}eating. She continually speaks to nobody as she eats. she \\
       &  \textbf{\tool (T)} & \textcolor{red}{authorities} woman is sitting at a table in a fast food restaurant while eating. She continually speaks to nobody as she eats. she \\
       &  \textbf{\tool (S)} & A woman is sitting at a table in a fast food restaurant while eating. She continually speaks to nobody as she eats. \textcolor{red}{It} \\
       &  \textbf{\tool-B (C)} & A woman is sitting at a table in a fast food restaurant while eating. She continually speaks to nobody as s\textcolor{red}{z}he eats. she \\
       &  \textbf{\tool-B (T)} & \textcolor{red}{exhibitors} woman is sitting at a table in a fast food restaurant while eating. She continually speaks to nobody as she eats. she \\
       &  \textbf{\tool-B (S)} & A woman is sitting at a table in a fast food restaurant while eating. She continually speaks to nobody as she eats. \textcolor{red}{It} \\
\midrule
\multicolumn{1}{l}{\multirow{7}{*}
            {\textbf{CodeGen}}} & \textbf{Original} & def sum\_div(number: int) -> int:$\backslash$n """$\backslash$n$\backslash$tWrite a function to return the sum of all divisors of a number.$\backslash$n$\backslash$t"""$\backslash$n \\
\cmidrule{2-3}
            & \textbf{\tool (C)} & de\textcolor{red}{g}f sum\_div(number: int) -> int:$\backslash$n """$\backslash$n$\backslash$tWrite a function to return the sum of all divisors of a number.$\backslash$n$\backslash$t"""$\backslash$n \\

            & \textbf{\tool (T)} & def \textcolor{red}{umption}\_div(number: int) -> int:$\backslash$n """$\backslash$n$\backslash$tWrite a function to return the sum of all divisors of a number.$\backslash$n$\backslash$t"""$\backslash$n \\         
            
            & \textbf{\tool (S)} & def sum\_div(number: int) -> int:$\backslash$n """$\backslash$n$\backslash$tWrite a function to return the sum \textcolor{red}{until} all divisors of a number.$\backslash$n$\backslash$t"""$\backslash$n \\   

            & \textbf{\tool-B (C)} & def sum\_div(number: int) -> int:$\backslash$n \textcolor{red}{1}"""$\backslash$n$\backslash$tWrite a function to return the sum of all divisors of a number.$\backslash$n$\backslash$t"""$\backslash$n \\

            & \textbf{\tool-B (T)} & \textcolor{red}{Huge} sum\_div(number: int) -> int:$\backslash$n """$\backslash$n$\backslash$tWrite a function to return the sum of all divisors of a number.$\backslash$n$\backslash$t"""$\backslash$n \\

            & \textbf{\tool-B (S)} & def sum\_div(number: int) -> int:$\backslash$n """$\backslash$n$\backslash$tWrite a function to return the \textcolor{red}{death} of all divisors of a number.$\backslash$n$\backslash$t"""$\backslash$n \\

    \bottomrule
    \end{tabular}%
    }
  \label{tab:test_sample}%
\end{table}
\noindent\textbf{Results.}
\tabref{tab:severity_all} and \tabref{tab:severity_max} showcase the average and maximum efficiency degradation results under varied perturbations for LLMs, respectively. Specifically, we recorded the required I-Loops, I-Latency, and I-energy for all test inputs, providing their mean and best-performing outcomes. 
Furthermore, \tabref{tab:test_sample} showcases examples of test samples generated by \tool, with \texttt{Original} denoting seed sentences and red font highlighting perturbed segments.
To elaborate, \tool(C), \tool(T), \tool(S) denote character-level, token-level, and structure-level perturbations in white-box settings, respectively. Similarly, \tool-B (C), \tool-B (T), \tool-B (S) represent character-level, token-level, and structure-level perturbations in black-box settings.
From the results, we have the following observations: \textit{(i) } For all LLMs under test, \tool generates test samples that trigger more severe efficiency degradation by a large margin compared to the baseline methods. 
Specifically, \tool generates test inputs that on average increase LLMs for translation (\ie the first six models)' CPU latency, CPU energy consumption, GPU latency, and GPU energy consumption by 100\% to 776\%, 101\% to 768\%,   96\% to 537\% , and 82\% to 539\%, respectively, through only perturbing one character or token in any seed input sentences.
Correspondingly, the LLMs for the sentence completion task (\ie Flan-T5 and LaMini-GPT) can increase 547\% to 1,890\%, 662\% to 2,245\%, 534\% to 1,682\%, 534\% to 1,682\%, respectively. For code-generated LLMs (\ie CodeGen), the increases are 321\% to 578\%, 336\% to 603\%, 351\% to 640\%, and 351\% to 640\%. Notably, \tool-B demonstrates performance comparable to \tool, signifying \tool-B equally effectively influences the efficiency of LLMs. 
In addition, \tool-B proves more effective than \tool in character type perturbations(\ie +32.49\%). This indicates our success in finding critical tokens within the black-box scenario presented in \secref{sec:black_box}. 
However, baseline methods could not effectively impact efficiency, since they are designed to reduce LLMs' accuracy, not efficiency.
\textit{(ii)} With an increased perturbation size, the corresponding test samples generated by \tool effectively degrade LLMs' efficiency to a larger degree. 
\textit{(iii)} The maximum effectiveness of our methods is far greater than the average case for most scenarios. Additionally, the computational efficiency of LLMs can be dramatically compromised through specific perturbations (\eg employing the \tool-B (C) on the H-NLP model, where a single character perturbation can lead to a maximum increase of 11,418\% in CPU energy consumption).

\begin{center}
\begin{tcolorbox}[colback=gray!10,
                  colframe=black,
                  width=0.98\textwidth,
                  arc=1mm, auto outer arc,
                  boxrule=0.9pt,
                 ]
Answers to \textbf{RQ2.1}: Test samples generated by \tool in both white-box and black-box settings significantly degrade LLMs efficiency in number of iteration loops, latency, and energy consumption. 
\end{tcolorbox}
\end{center}

\subsection{RQ2.2: Effectiveness}
\label{sec:effectiveness}

This section evaluates the effectiveness of \tool in generating useful test samples that successfully degrade the efficiency of LLMs.

\noindent\textbf{Metrics.} We define a metric of degradation success ratio ($\eta$) to evaluate the effectiveness of \tool.
\begin{equation}
        \eta = \frac{\sum_{x \in \mathcal{X}} \mathbb I( [\text{Loop}(x') - \text{Loop}(x)] \ge \lambda \times \text{MSE}_{orig} ) }{||\mathcal{X}||} \times 100\%
    \label{eq:validity}
\end{equation}
As shown in \equref{eq:validity}, $\mathcal{X}$ is a randomly selected seed input set, Loop($x$) is the function that measures the iteration number of LLMs in handling input $x$, $\text{MSE}_{orig}$ is the Mean Squared Error of the iteration number in the seed datasets that have the same input length as $x$, and $\mathbb I(\cdot)$ is the indicator function, which returns one if the statement is true, zero otherwise.
The above equation assumes that the computational costs required by an LLM given perturbed inputs shall be within $\lambda$ times the MSE produced by the seed inputs with the same input length. Otherwise, the perturbed inputs trigger efficiency degradation. Note that this same assumption is also used in existing works~\cite{tian2018deeptest}.

\begin{figure}[!ht]
    \centering

    \includegraphics[width=0.72\textwidth]{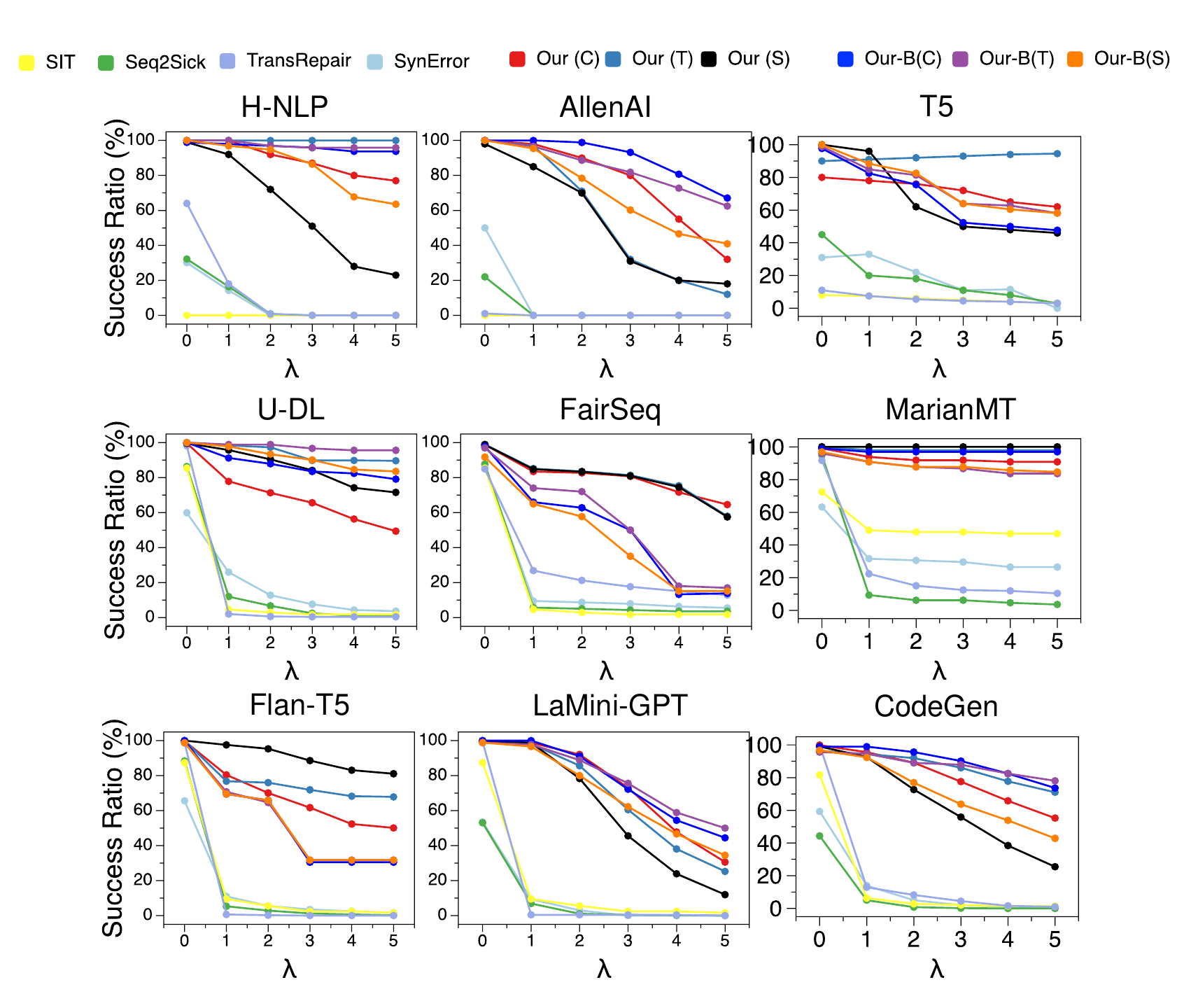}

    \caption{Degradation success ratio under different settings}
    \label{fig:validity}

\end{figure}
\noindent\textbf{Results.}
The results on the degradation successful ratio ($\eta$) under different $\lambda$ values are shown in \figref{fig:validity}. 
We observe that for all experimental settings, \tool outperforms the baseline methods by a significant margin in both white-box and black-box settings. For example, for U-DL and $\lambda=5$, \tool achieves a degradation success ratio over 50\% with all type perturbations in both white-box and
black-box scenarios; while all the comparison baseline methods' degradation success ratios are below 5\%.
The results indicate that \tool effectively generates useful test samples to trigger LLMs' efficiency degradation.
Another observation is that when $\lambda=0$, baselines may generate some test samples that require more computations than seed inputs ($\eta \ge 50$ for H-NLP). However, such extra computations are not significant enough to indicate efficiency degradation. As we studied in \secref{sec:preliminary}, the natural efficiency variance in the LLM task could be significant, and the degree of extra computations incurred under baseline methods is within the range of natural efficiency variance.
As $\lambda$ grows, $\eta$ under baseline methods drop quickly.
However, this observation does not hold for \tool, where the average degradation success ratio of \tool is still 72.32\% when $\lambda=3$.
Recall that from the statistical prospective~\cite{statistic}, $99.73\%$ of the inputs will locate in the range of 3$\text{MSE}_{orig}$.
Thus, these results clearly show that \tool successfully triggers LLMs' efficiency degradation.

\begin{center}
\begin{tcolorbox}[colback=gray!10,
                  colframe=black,
                 width=0.98\textwidth,
                  arc=1mm, auto outer arc,
                  boxrule=0.9pt,
                 ]
Answers to \textbf{RQ2.2}: \tool effectively generates test samples that trigger LLMs' efficiency degradation in both white-box and
black-box settings. 
\end{tcolorbox}
\end{center}

\label{sec:senstive}
\begin{figure}[ht]
    \centering
    \includegraphics[width=0.88\textwidth]{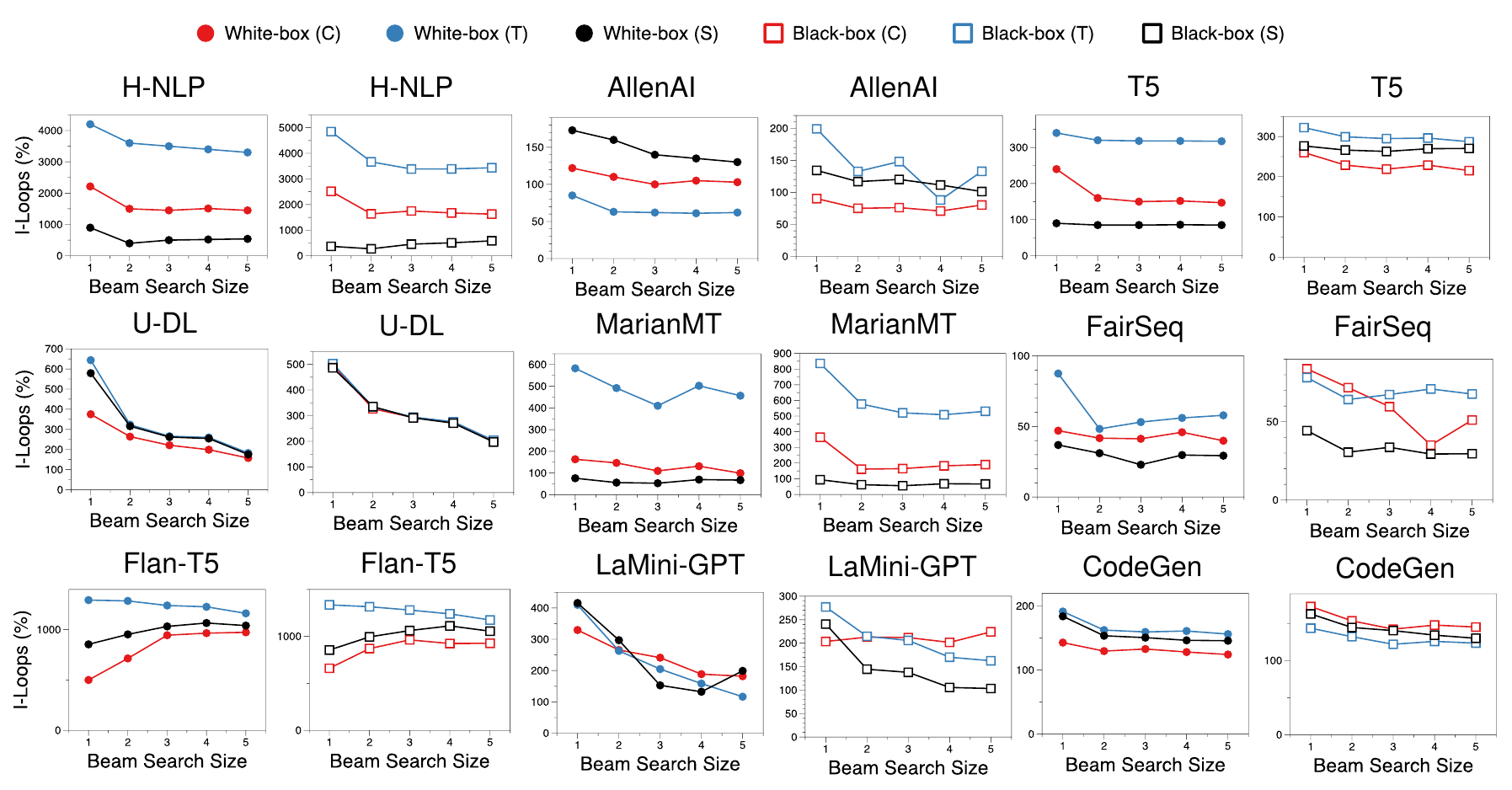}
    \caption{I-Loops under different beam search sizes}
    \label{fig:sensitive}
\end{figure}

\begin{figure}[ht]
    \centering
    \includegraphics[width=0.88\textwidth]{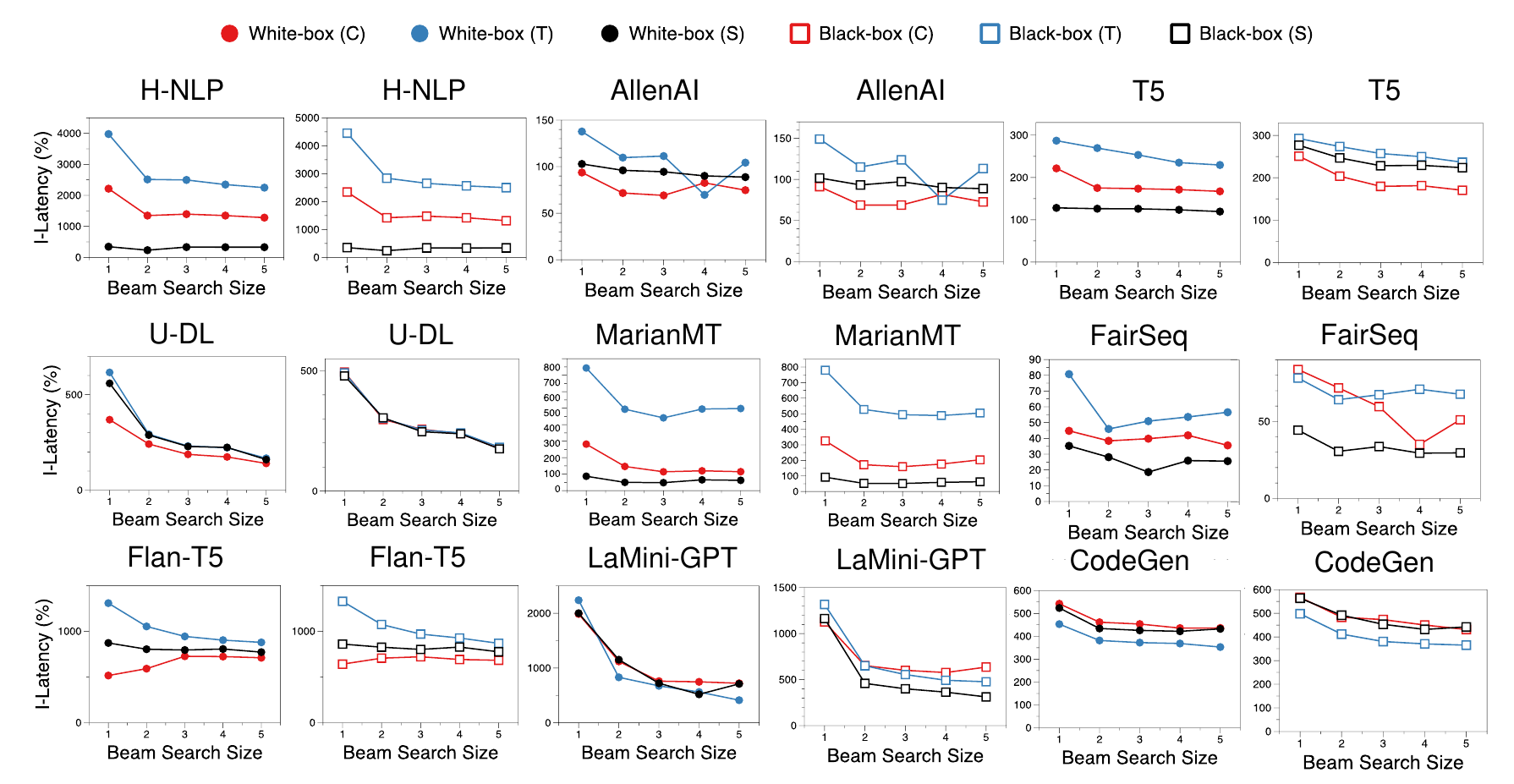}
    \caption{GPU-Latency under different beam search sizes}
    \label{fig:beam_latency}
\end{figure}

\begin{figure}[ht]
    \centering
    \includegraphics[width=0.88\textwidth]{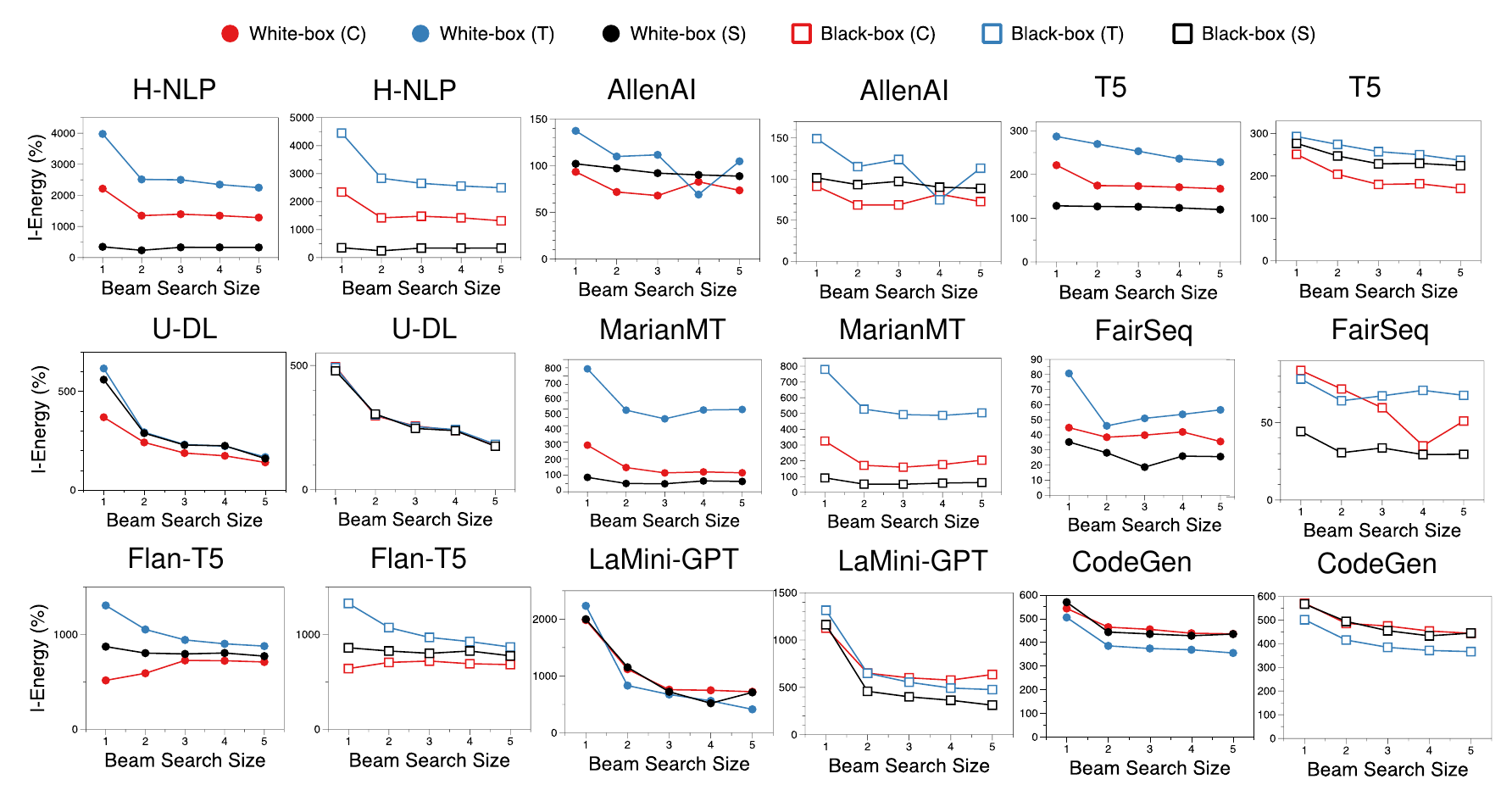}
    \caption{GPU-Energy under different beam search sizes}
    \label{fig:beam_energy}
\end{figure}

\begin{figure}[ht]
    \centering
    \includegraphics[width=0.88\textwidth]{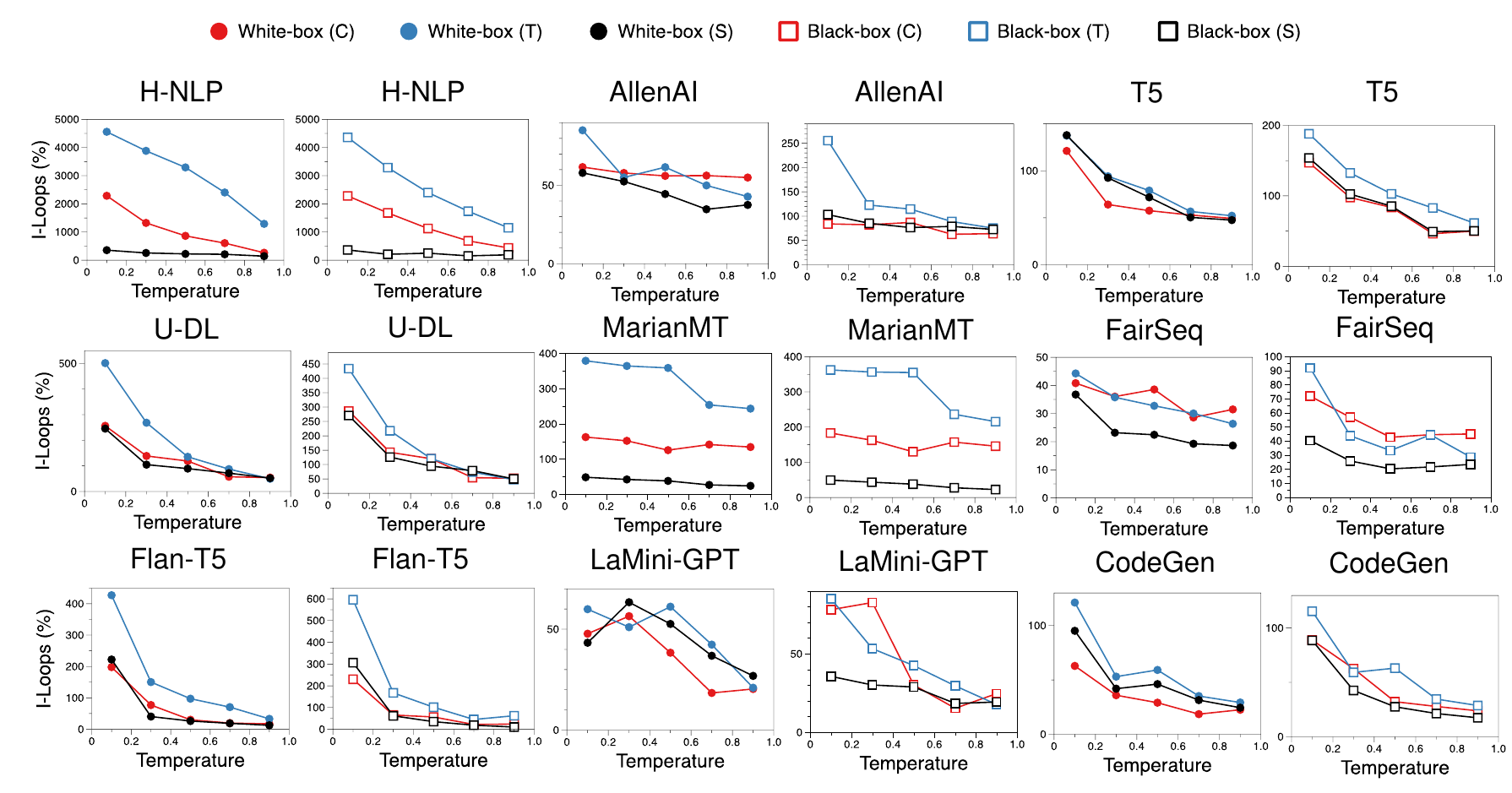}
    \caption{I-Loops under different temperatures}
    \label{fig:temp_loops}
\end{figure}

\begin{figure}[ht]
    \centering
    \includegraphics[width=0.88\textwidth]{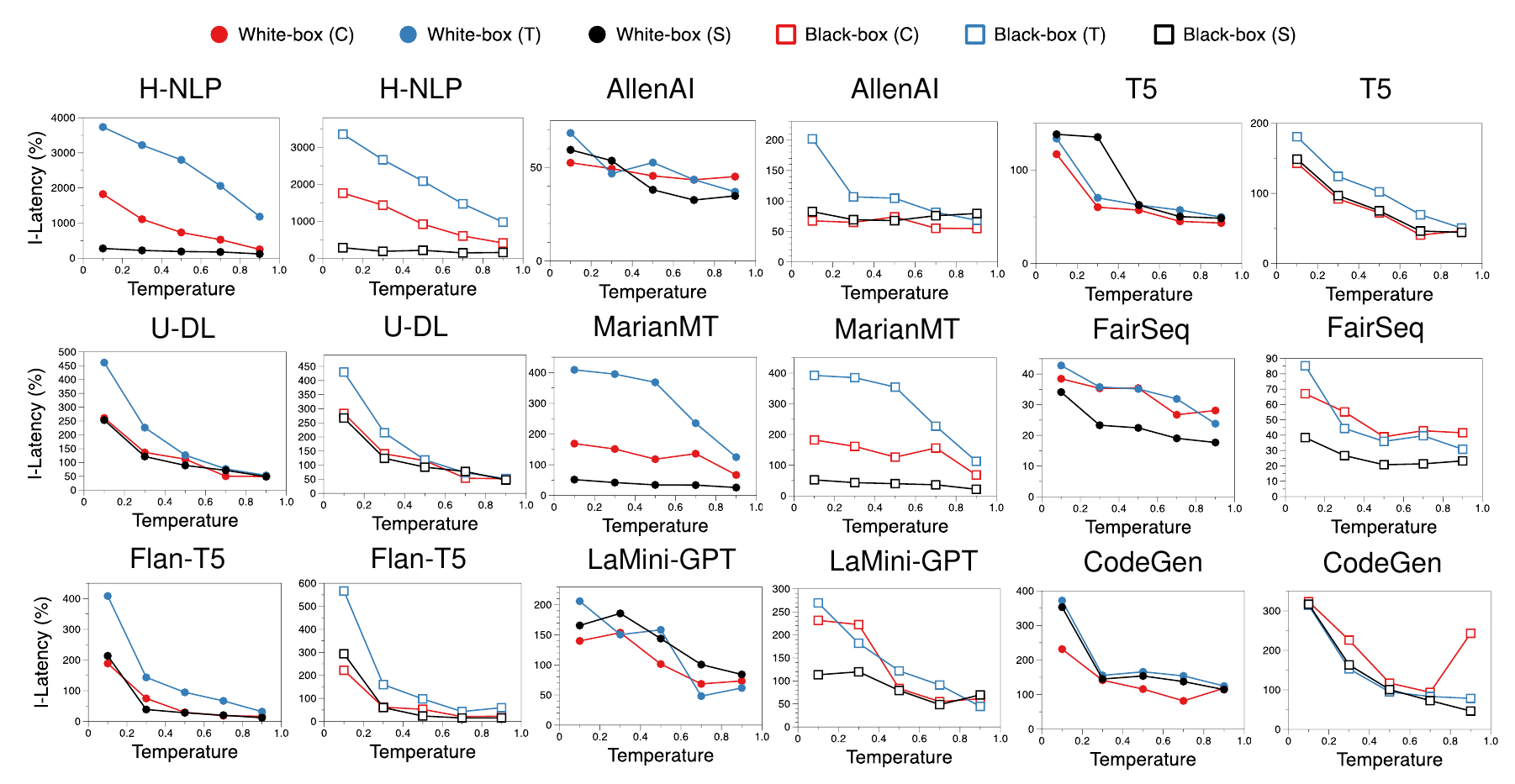}
    \caption{GPU-Latency under different temperatures}
    \label{fig:temp_latency}
\end{figure}

\begin{figure}[ht]
    \centering
    \includegraphics[width=0.88\textwidth]{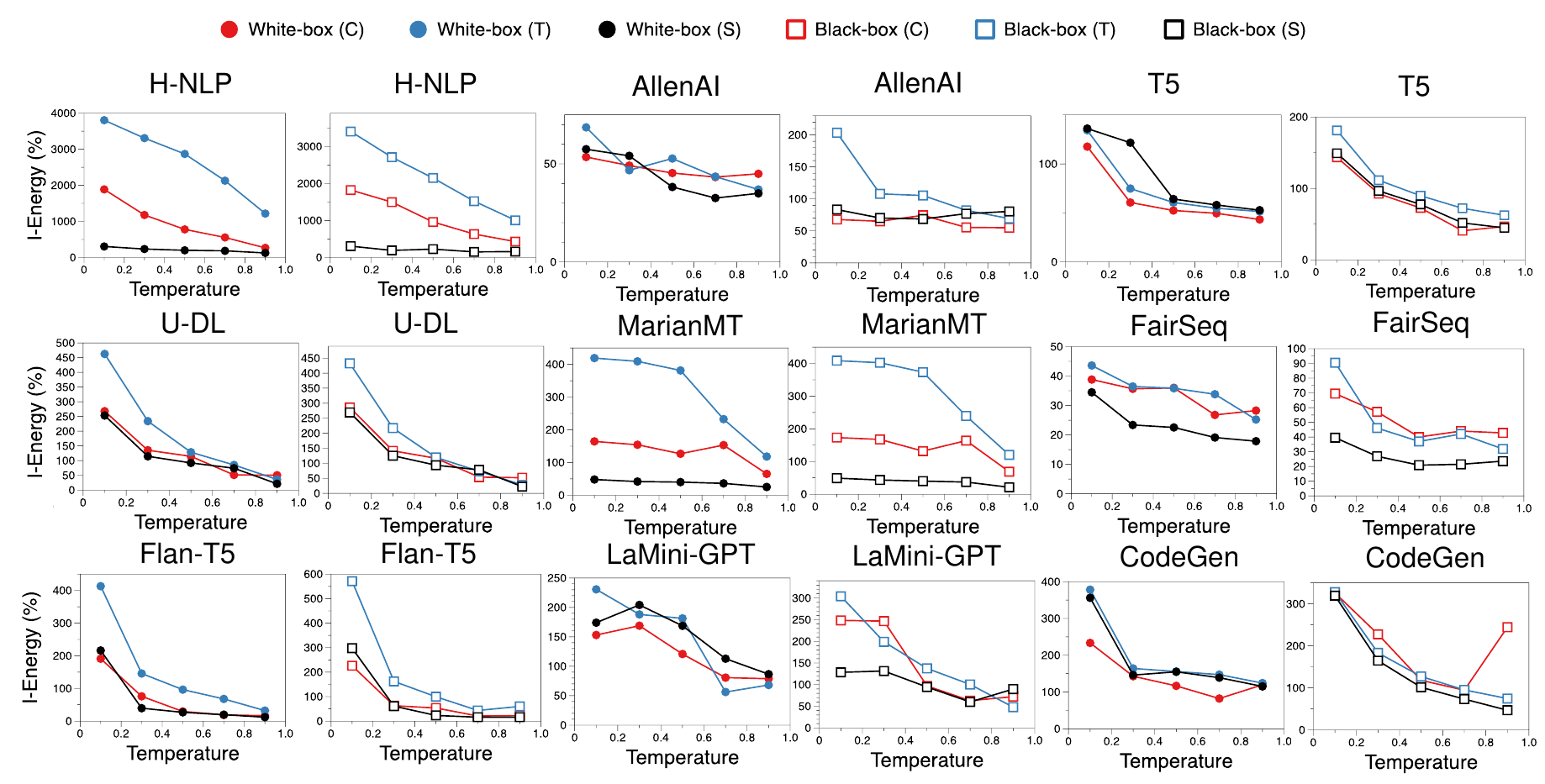}
    \caption{GPU-Energy under different temperatures}
    \label{fig:temp_energy}
\end{figure}

\subsection{RQ2.3: Sensitivity}
In this section, we implement two prevalent decoding methods from LLMs with comprehensive hyperparameter settings to thoroughly evaluate the performance of \tool: Beam Search and Temperature Sampling.

\noindent\fakeparagraph{Experimental Setup} In the first configuration, we investigate the impact of varying the beam search size on the efficiency of LLMs.  
As we introduced in \secref{sec:background}, modern LLMs apply the beam search algorithm to generate the output token. 
The beam search algorithm requires one hyper-parameter, the beam search size (\texttt{num\_beams}), to define the search space. 
In \secref{sec:effectiveness}, we evaluate the effectiveness of \tool under each LLMs' default \texttt{num\_beams}.
In this section, we evaluate whether \tool is sensitive to the configuration of \texttt{num\_beams}.
We configure each LLM under test with different \texttt{num\_beams} (ranging from 1 to 5) and measure the I-Loops, GPU latency, and GPU energy consumption of the generated test samples. 
In the second configuration, we focus on the effects of enabling sampling (do\_sample = true) and varying the temperature parameter(\ie 0.1, 0.3, 0.5, 0.7 and 0.9) to understand its impact on \tool. The temperature parameter controls the level of randomness in the sampling process, with lower temperatures leading to less variability and higher temperatures allowing for more diverse outputs.

\noindent\fakeparagraph{Experimental Results} 
The results of I-Loops, GPU-Latency, and GPU-Energy for different beam sizes under Beam Search are respectively presented in \figref{fig:sensitive}, \figref{fig:beam_latency}, and \figref{fig:beam_energy}. Similarly, the results of I-Loops, GPU-Latency, and GPU-Energy for different temperatures under Temperature Sampling are respectively presented in \figref{fig:temp_loops}, \figref{fig:temp_latency}, and \figref{fig:temp_energy}. 
From the results, we have the following observations: \textit{(i)} When the beam search size \texttt{num\_beams} is set to 1, the test samples generated by \tool can degrade LLMs efficiency slightly more than other beam search size settings in both white-box and black-box
scenarios.
This is because when \texttt{num\_beams}=1, the token generation process is a gradient-smooth process, and the token search space is limited. Thus, our gradient-guided and causal inference-based approach can trigger more severe efficiency degradation under this configuration.
%
\textit{(ii)} In temperature-controlled sampling, setting the temperature to 0.1 allows for the generation of test inputs that slightly improve the reduction of LLMs' computational efficiency. This is because at a lower temperature (\ie 0.1), the sampling process becomes more deterministic, making the model more likely to choose tokens with the highest probability. This can lead to generating sequences that are highly structured. The generated test samples are more focused and consistent to triggering inefficient computation paths within the LLMs.
\textit{(iii)} 
Across both sets of results, it is evident that \tool consistently and significantly degrades the computational efficiency of the LLMs across a diverse range of beam search size settings and temperature configurations.
(\eg T5 requires more than 100\% and 300\% computations).

\begin{center}
\begin{tcolorbox}[colback=gray!10,
                  colframe=black,
                  width=0.98\textwidth,
                  arc=1mm, auto outer arc,
                  boxrule=0.9pt,
                 ]
Answers to \textbf{RQ2.3}: \tool can generate test samples that degrade LLMs efficiency under various decoding methods with comprehensive
hyperparameter settings in both white-box and black-box settings. Moreover, the efficiency degradation is more severe when the beam search size is configured as one or temperature is set to 0.1.
\end{tcolorbox}
\end{center}

\subsection{RQ2.4: Overheads}
\label{sec:overheads}
\tabref{tab:new_overheads} and \tabref{tab:overheads_B} show the average overhead of LLMEffiChecker when generating test inputs in white-box and black-box scenarios, respectively. We report only the overhead of \tool because the comparison baselines cannot degrade LLMs' efficiency. 
The measured overhead of \tool is rather reasonable (ranging from 2.25s to 191.32s) and may increase linearly as the perturbation size increases. 
The linearly increasing overheads are due to the fact that \tool is an iterative approach (iteration number equals to $\epsilon$), and the overhead within each iteration is stable.
Additionally, the overhead of \tool-B is reduced by 16.74\% compared to \tool, as it eliminates the need for gradient calculations.
Note that such reasonable overhead is not a concern since perturbed test inputs are generated by \tool offline.
\begin{table}[htbp]
  \centering
  \caption{Average overheads of \tool (s)}
  \centering
  \resizebox{0.98\textwidth}{!}{
\begin{tabular}{c|c|ccccccccc|c}
\toprule
\multicolumn{1}{c|}{\textbf{Methods}} & \bm{$\epsilon$} & \textbf{H-NLP} & \textbf{AllenAi} & \textbf{T5} & \textbf{U-DL} & \textbf{FairSeq} & \textbf{MarianMT} & \textbf{Flan-T5} & \textbf{LaMini-GPT} & \textbf{CodeGen} & \multicolumn{1}{l}{\textbf{Average}} \\
\midrule
\multicolumn{1}{c|}{\multirow{3}[2]{*}{\textbf{\tool(C)\newline{}}}} & 1     & 11.40 & 21.14 & 18.50 & 9.00  & 12.40 & 10.05 & 5.21  & 2.57  & 20.07 & 11.13 \\
      & 2     & 31.80 & 44.66 & 45.59 & 22.09 & 28.05 & 22.83 & 12.77 & 8.42  & 46.98 & 26.52 \\
      & 3     & 59.76 & 69.56 & 74.48 & 42.26 & 47.70 & 39.91 & 20.84 & 14.63 & 75.27 & 44.74 \\
\midrule
\multicolumn{1}{c|}{\multirow{3}[2]{*}{\textbf{\tool(T)}}} & 1     & 7.50  & 18.45 & 22.62 & 31.56 & 52.80 & 38.92 & 17.85 & 5.94  & 26.70 & 22.33 \\
      & 2     & 31.41 & 39.48 & 61.86 & 66.19 & 108.75 & 84.74 & 39.16 & 13.99 & 67.80 & 51.54 \\
      & 3     & 62.50 & 62.54 & 100.01 & 101.28 & 165.80 & 131.74 & 62.09 & 22.35 & 110.76 & 82.21 \\
\midrule
\multicolumn{1}{c|}{\multirow{3}[2]{*}{\textbf{\tool(S)}}} & 1     & 10.52 & 39.19 & 6.73  & 24.74 & 25.91 & 18.83 & 12.62 & 8.01  & 30.65 & 17.82 \\
      & 2     & 23.33 & 75.21 & 17.45 & 59.05 & 53.85 & 39.83 & 29.49 & 19.17 & 69.94 & 38.93 \\
      & 3     & 38.93 & 106.35 & 27.71 & 93.07 & 82.87 & 61.92 & 49.97 & 30.19 & 111.21 & 60.52 \\
\bottomrule
\end{tabular}%
}
  \label{tab:new_overheads}%
\end{table}%
\begin{table}[htbp]
  \centering
  \caption{Average overheads of \tool-B (s)}
  \centering
  \resizebox{0.98\textwidth}{!}{
\begin{tabular}{c|c|ccccccccc|c}
\toprule
\multicolumn{1}{c|}{\textbf{Methods}} & \bm{$\epsilon$} & \textbf{H-NLP} & \textbf{AllenAi} & \textbf{T5} & \textbf{U-DL} & \textbf{FairSeq} & \textbf{MarianMT} & \textbf{Flan-T5} & \textbf{LaMini-GPT} & \textbf{CodeGen} & \multicolumn{1}{l}{\textbf{Average}} \\
\midrule
\multicolumn{1}{c|}{\multirow{3}[2]{*}{\textbf{\tool-B (C)\newline{}}}} & 1     & 9.73  & 17.49 & 6.05  & 10.85 & 24.51 & 15.27 & 2.59  & 2.25  & 10.73 & 10.05 \\
      & 2     & 20.57 & 42.29 & 10.44 & 18.02 & 55.37 & 34.38 & 5.66  & 4.57  & 16.78 & 21.01 \\
      & 3     & 31.12 & 70.62 & 14.76 & 24.93 & 92.63 & 56.17 & 9.03  & 6.88  & 22.88 & 33.20 \\
\midrule
\multicolumn{1}{c|}{\multirow{3}[2]{*}{\textbf{\tool-B (T)}}} & 1     & 6.86  & 58.78 & 6.62  & 20.97 & 63.68 & 45.49 & 10.82 & 1.89  & 8.34  & 22.45 \\
      & 2     & 11.52 & 113.03 & 10.61 & 23.80 & 130.30 & 86.04 & 12.85 & 3.34  & 12.97 & 40.65 \\
      & 3     & 15.69 & 157.59 & 11.97 & 25.93 & 191.32 & 121.11 & 14.68 & 4.82  & 16.66 & 56.28 \\
\midrule
\multicolumn{1}{c|}{\multirow{3}[2]{*}{\textbf{\tool-B (S)}}} & 1     & 3.19  & 30.84 & 16.35 & 32.84 & 29.78 & 25.58 & 15.73 & 9.66  & 24.26 & 18.92 \\
      & 2     & 7.60  & 63.07 & 31.01 & 61.34 & 62.74 & 52.28 & 31.12 & 19.04 & 45.42 & 37.56 \\
      & 3     & 13.51 & 94.78 & 46.55 & 85.30 & 98.08 & 80.89 & 45.43 & 28.42 & 65.11 & 56.11 \\
\bottomrule
\end{tabular}%
}
  \label{tab:overheads_B}%
\end{table}%
\vspace{2mm}
\begin{center}
\begin{tcolorbox}[colback=gray!10,
                  colframe=black,
                  width=0.98\textwidth,
                  arc=1mm, auto outer arc,
                  boxrule=0.9pt,
                 ]
Answers to \textbf{RQ2.4}: The overheads of \tool are reasonable and may increase linearly as the perturbation size increases. Specifically, when $\epsilon=1$, \tool costs 17.01, 16.19, and 18.81 seconds to generate character-level, token-level, and structure-level test samples. Correspondingly, \tool-B costs 10.05, 22.45 and 18.92 seconds to generate samples of the same type.
\end{tcolorbox}
\end{center}

\subsection{RQ2.5: Ablation Study}
\label{sec:(Ablation Study}

\noindent In this experiment, we carried out ablation studies to assess the efficacy of $p_i^{o_i}$ in \tool for identifying critical tokens, as illustrated in \equref{eq:obj}.
The inspiration for this component came from recent research, which showed that the sequence of tokens output by a model also affects the generation of the EOS token \cite{gao2024inducing}.
To validate this idea's effectiveness for \tool and ensure it aligns with our overarching goals, we remove $p_i^{o_i}$  from the function $f(x)$ in \equref{eq:obj} and then apply it to generate test inputs.

\begin{table*}[htbp]
  \centering
  \caption{The Ablation Results of Ours-Ablation in Degrading LLMs Performance}
    \resizebox{\linewidth}{!}{
        \begin{tabular}{llccccccccccccccc}
  
    \toprule[1.2pt]
    \multirow{2}[2]{*}{\textbf{Subject}} & \multirow{2}[2]{*}{\textbf{Methods}} & \multicolumn{3}{c}{\textbf{I-Loops}} & \multicolumn{3}{c}{\textbf{I-Latency(CPU)}} & \multicolumn{3}{c}{\textbf{I-Energy(CPU)}} & \multicolumn{3}{c}{\textbf{I-Latency(GPU)}} & \multicolumn{3}{c}{\textbf{I-Energy(GPU)}} \\
    \cmidrule{3-17}
          &       & \multicolumn{1}{c}
          {$\bm{\epsilon = 1 }$} & \multicolumn{1}{c}
          {$\bm{\epsilon = 2 }$} & \multicolumn{1}{c}
          {$\bm{\epsilon = 3 }$} & \multicolumn{1}{c}
          {$\bm{\epsilon = 1 }$} & \multicolumn{1}{c}
          {$\bm{\epsilon = 2 }$} & \multicolumn{1}{c}
          {$\bm{\epsilon = 3 }$} & \multicolumn{1}{c}
          {$\bm{\epsilon = 1 }$} & \multicolumn{1}{c}
          {$\bm{\epsilon = 2 }$} & \multicolumn{1}{c}
          {$\bm{\epsilon = 3 }$} & \multicolumn{1}{c}
          {$\bm{\epsilon = 1 }$} & \multicolumn{1}{c}
          {$\bm{\epsilon = 2 }$} & \multicolumn{1}{c}
          {$\bm{\epsilon = 3 }$} & \multicolumn{1}{c}
          {$\bm{\epsilon = 1 }$} & \multicolumn{1}{c}
          {$\bm{\epsilon = 2 }$} & \multicolumn{1}{c}
          {$\bm{\epsilon = 3 }$} \\
\midrule
\multirow{6}[2]{*}{\textbf{H-NLP}} 
& \textbf{Original (C)} & 564.45  & 995.45  & 1357.77  & 764.92  & 1487.92  & 2015.70  & 785.60  & 1471.26  & 1967.05  & 462.24  & 851.80  & 1116.80  & 406.39  & 755.18  & 972.92  \\
     & \textbf{Removed (C)} & \textbf{493.90 } & \textbf{934.79 } & \textbf{1244.70 } & \textbf{592.94 } & \textbf{1133.14 } & \textbf{1547.26 } & \textbf{613.07 } & \textbf{1117.62 } & \textbf{1506.78 } & \textbf{352.26 } & \textbf{668.56 } & \textbf{888.26 } & \textbf{309.12 } & \textbf{592.12 } & \textbf{773.13 } \\
\cmidrule{2-17}      & \textbf{Original (T)} & 2697.77  & 3735.38  & 3917.91  & 3153.97  & 4481.93  & 4681.28  & 3052.62  & 4544.65  & 4759.71  & 1953.57  & 2729.83  & 2854.89  & 1532.91  & 2137.53  & 2221.66  \\
     & \textbf{Removed (T)} & \textbf{1594.55 } & \textbf{2271.86 } & \textbf{2382.88 } & \textbf{1621.64 } & \textbf{2355.45 } & \textbf{2460.22 } & \textbf{1555.09 } & \textbf{2359.78 } & \textbf{2471.45 } & \textbf{1058.38 } & \textbf{1518.38 } & \textbf{1587.94 } & \textbf{830.35 } & \textbf{1188.80 } & \textbf{1235.59 } \\
\cmidrule{2-17}      & \textbf{Original (S)} & 142.31  & 311.06  & 612.08  & 146.51  & 451.93  & 877.79  & 147.70  & 461.30  & 870.72  & 101.21  & 275.58  & 523.04  & 95.05  & 259.88  & 508.80  \\
      & \textbf{Removed (S)} & \textbf{127.09 } & \textbf{280.58 } & \textbf{593.16 } & \textbf{136.92 } & \textbf{340.42 } & \textbf{632.91 } & \textbf{136.97 } & \textbf{346.46 } & \textbf{624.23 } & \textbf{98.81 } & \textbf{219.31 } & \textbf{408.83 } & \textbf{92.76 } & \textbf{206.57 } & \textbf{397.26 } \\
\midrule
\multirow{6}[2]{*}{\textbf{AllenAI}} 
& \textbf{Original (C)} & 35.16  & 74.90  & 103.36  & 26.69  & 45.77  & 85.09  & 27.48  & 48.09  & 86.00  & 21.82  & 35.43  & 91.48  & 22.12  & 43.21  & 98.46  \\
     & \textbf{Removed (C)} & \textbf{44.62 } & \textbf{88.40 } & \textbf{127.77 } & \textbf{34.73 } & \textbf{73.64 } & \textbf{111.38 } & \textbf{59.37 } & \textbf{108.81 } & \textbf{154.37 } & \textbf{38.93 } & \textbf{77.78 } & \textbf{115.71 } & \textbf{38.80 } & \textbf{77.59 } & \textbf{115.48 } \\
\cmidrule{2-17}      & \textbf{Original (T)} & 24.83  & 42.04  & 56.75  & 49.12  & 62.84  & 67.98  & 49.99  & 62.65  & 69.06  & 30.65  & 41.32  & 46.09  & 31.00  & 41.81  & 49.66  \\
      & \textbf{Removed (T)} & \textbf{34.73 } & \textbf{58.45 } & \textbf{89.32 } & \textbf{25.04 } & \textbf{50.13 } & \textbf{86.52 } & \textbf{45.17 } & \textbf{75.62 } & \textbf{118.70 } & \textbf{29.57 } & \textbf{55.62 } & \textbf{92.69 } & \textbf{29.45 } & \textbf{55.46 } & \textbf{92.50 } \\
\cmidrule{2-17}      & \textbf{Original (S)} & 66.21  & 108.67  & 128.60  & 86.05  & 139.03  & 164.57  & 84.17  & 135.71  & 160.95  & 69.57  & 112.88  & 132.68  & 68.79  & 115.23  & 137.06  \\
      & \textbf{Removed (S)} & \textbf{67.48 } & \textbf{99.92 } & \textbf{131.33 } & \textbf{79.34 } & \textbf{115.42 } & \textbf{144.89 } & \textbf{111.56 } & \textbf{153.76 } & \textbf{189.10 } & \textbf{84.90 } & \textbf{120.76 } & \textbf{152.29 } & \textbf{84.67 } & \textbf{120.48 } & \textbf{151.97 } \\
\midrule
\multirow{6}[2]{*}{\textbf{T5}} 
& \textbf{Original (C)} & 168.92  & 198.36  & 205.37  & 191.05  & 225.48  & 233.01  & 194.45  & 228.02  & 234.04  & 164.61  & 194.79  & 202.28  & 165.38  & 195.77  & 203.29  \\
      & \textbf{Removed (C)} & \textbf{155.44 } & \textbf{188.32 } & \textbf{190.63 } & \textbf{168.33 } & \textbf{215.73 } & \textbf{218.59 } & \textbf{173.43 } & \textbf{218.39 } & \textbf{220.48 } & \textbf{150.10 } & \textbf{183.60 } & \textbf{185.25 } & \textbf{151.30 } & \textbf{185.13 } & \textbf{186.71 } \\
\cmidrule{2-17}      & \textbf{Original (T)} & 307.27  & 328.94  & 328.94  & 352.14  & 376.55  & 376.55  & 347.74  & 373.85  & 373.85  & 305.37  & 325.61  & 325.61  & 331.85  & 352.25  & 352.25  \\
     & \textbf{Removed (T)} & \textbf{294.68 } & \textbf{315.47 } & \textbf{315.47 } & \textbf{334.63 } & \textbf{357.83 } & \textbf{357.83 } & \textbf{328.58 } & \textbf{353.25 } & \textbf{353.25 } & \textbf{294.72 } & \textbf{314.25 } & \textbf{314.25 } & \textbf{320.11 } & \textbf{339.79 } & \textbf{339.79 } \\
\cmidrule{2-17}      & \textbf{Original (S)} & 77.67  & 80.56  & 82.52  & 85.72  & 89.11  & 91.38  & 86.90  & 90.29  & 92.56  & 75.77  & 78.68  & 80.66  & 68.79  & 73.03  & 74.56  \\     & \textbf{Removed (S)} & \textbf{78.47 } & \textbf{81.24 } & \textbf{83.37 } & \textbf{85.44 } & \textbf{88.73 } & \textbf{91.24 } & \textbf{87.53 } & \textbf{90.81 } & \textbf{93.25 } & \textbf{76.75 } & \textbf{79.71 } & \textbf{82.14 } & \textbf{69.65 } & \textbf{73.96 } & \textbf{75.90 } \\
\midrule
\multirow{6}[2]{*}{\textbf{U-DL}} 
& \textbf{Original (C)} & 258.07  & 390.60  & 469.24  & 261.02  & 405.80  & 494.30  & 288.15  & 439.72  & 532.81  & 253.46  & 383.78  & 461.51  & 253.45  & 383.82  & 461.64  \\
      & \textbf{Removed (C)} & \textbf{154.67 } & \textbf{274.00 } & \textbf{350.74 } & \textbf{157.17 } & \textbf{282.11 } & \textbf{366.04 } & \textbf{170.59 } & \textbf{308.67 } & \textbf{399.07 } & \textbf{156.19 } & \textbf{274.11 } & \textbf{352.54 } & \textbf{156.32 } & \textbf{274.10 } & \textbf{352.59 } \\
\cmidrule{2-17}      & \textbf{Original (T)} & 604.17  & 642.38  & 642.38  & 655.13  & 696.56  & 696.56  & 697.90  & 741.86  & 741.86  & 595.88  & 634.44  & 634.44  & 596.49  & 635.08  & 635.08  \\
      & \textbf{Removed (T)} & \textbf{595.70 } & \textbf{595.70 } & \textbf{595.70 } & \textbf{635.50 } & \textbf{635.50 } & \textbf{635.50 } & \textbf{678.60 } & \textbf{678.60 } & \textbf{678.60 } & \textbf{590.44 } & \textbf{590.44 } & \textbf{590.44 } & \textbf{590.44 } & \textbf{590.44 } & \textbf{590.44 } \\
\cmidrule{2-17}      & \textbf{Original (S)} & 406.92  & 592.52  & 702.89  & 438.23  & 632.64  & 753.88  & 465.04  & 673.76  & 800.25  & 404.42  & 583.65  & 694.40  & 401.74  & 583.99  & 694.84  \\
      & \textbf{Removed (S)} & \textbf{329.64 } & \textbf{467.00 } & \textbf{501.56 } & \textbf{350.83 } & \textbf{494.42 } & \textbf{532.38 } & \textbf{374.35 } & \textbf{533.90 } & \textbf{573.56 } & \textbf{332.55 } & \textbf{469.61 } & \textbf{504.01 } & \textbf{332.57 } & \textbf{469.57 } & \textbf{504.01 } \\
\midrule
\multirow{6}[2]{*}{\textbf{FairSeq}} 
& \textbf{Original (C)} & 22.53  & 37.68  & 59.26  & 15.87  & 29.07  & 49.18  & 20.47  & 34.62  & 55.44  & 18.07  & 31.53  & 51.79  & 18.08  & 31.55  & 51.85  \\
      & \textbf{Removed (C)} & \textbf{19.79 } & \textbf{34.08 } & \textbf{51.59 } & \textbf{14.24 } & \textbf{25.78 } & \textbf{41.86 } & \textbf{18.16 } & \textbf{32.69 } & \textbf{48.79 } & \textbf{16.51 } & \textbf{30.98 } & \textbf{46.45 } & \textbf{16.51 } & \textbf{28.00 } & \textbf{46.49 } \\
\cmidrule{2-17}      & \textbf{Original (T)} & 33.73  & 62.13  & 76.41  & 23.97  & 55.26  & 70.28  & 29.79  & 63.19  & 79.41  & 28.62  & 59.42  & 75.47  & 28.60  & 59.42  & 75.47  \\
      & \textbf{Removed (T)} & \textbf{23.65 } & \textbf{37.82 } & \textbf{58.70 } & \textbf{16.25 } & \textbf{28.96 } & \textbf{47.25 } & \textbf{21.01 } & \textbf{34.80 } & \textbf{54.37 } & \textbf{18.99 } & \textbf{32.82 } & \textbf{52.96 } & \textbf{18.96 } & \textbf{32.81 } & \textbf{52.94 } \\
\cmidrule{2-17}      & \textbf{Original (S)} & 19.42  & 30.87  & 37.82  & 14.01  & 23.67  & 31.31  & 18.23  & 28.59  & 36.73  & 14.84  & 24.87  & 31.68  & 14.86  & 24.91  & 31.72  \\
      & \textbf{Removed (S)} & \textbf{19.17 } & \textbf{30.64 } & \textbf{36.33 } & \textbf{13.11 } & \textbf{22.57 } & \textbf{28.64 } & \textbf{16.96 } & \textbf{27.46 } & \textbf{34.15 } & \textbf{14.26 } & \textbf{23.90 } & \textbf{30.21 } & \textbf{14.75 } & \textbf{23.91 } & \textbf{30.22 } \\
\midrule
\multirow{6}[2]{*}{\textbf{MarianMT}} 
& \textbf{Original (C)} & 54.10  & 113.20  & 222.04  & 41.58  & 102.38  & 226.68  & 51.56  & 119.53  & 274.26  & 52.98  & 113.90  & 210.12  & 52.89  & 113.88  & 210.12  \\
     & \textbf{Removed (C)} & \textbf{60.13 } & \textbf{78.00 } & \textbf{105.23 } & \textbf{59.41 } & \textbf{78.72 } & \textbf{103.52 } & \textbf{70.11 } & \textbf{91.22 } & \textbf{118.31 } & \textbf{56.65 } & \textbf{74.42 } & \textbf{96.50 } & \textbf{56.78 } & \textbf{74.55 } & \textbf{96.66 } \\
\cmidrule{2-17}      & \textbf{Original (T)} & 231.65  & 550.07  & 726.61  & 234.93  & 564.34  & 770.28  & 269.58  & 660.70  & 893.87  & 223.56  & 544.49  & 728.90  & 223.48  & 544.28  & 728.68  \\
      & \textbf{Removed (T)} & \textbf{207.33 } & \textbf{284.10 } & \textbf{327.79 } & \textbf{235.99 } & \textbf{315.13 } & \textbf{398.78 } & \textbf{264.59 } & \textbf{353.72 } & \textbf{449.18 } & \textbf{201.15 } & \textbf{274.95 } & \textbf{346.98 } & \textbf{201.14 } & \textbf{274.88 } & \textbf{346.94 } \\
\cmidrule{2-17}      & \textbf{Original (S)} & 42.77  & 72.19  & 89.17  & 33.89  & 72.33  & 90.33  & 40.27  & 84.89  & 106.84  & 41.21  & 77.28  & 95.68  & 41.19  & 77.21  & 95.61  \\
\cmidrule{2-17}      & \textbf{Removed (S)} & \textbf{34.13 } & \textbf{46.25 } & \textbf{55.67 } & \textbf{29.43 } & \textbf{38.89 } & \textbf{48.23 } & \textbf{36.48 } & \textbf{46.97 } & \textbf{57.39 } & \textbf{30.14 } & \textbf{39.65 } & \textbf{48.94 } & \textbf{30.12 } & \textbf{39.62 } & \textbf{48.91 } \\
\midrule
\multirow{6}[2]{*}{\textbf{Flan-T5}} 
& \textbf{Original (C)} & 327.55  & 566.82  & 625.69  & 329.99  & 564.27  & 621.78  & 381.37  & 647.48  & 715.92  & 333.91  & 574.26  & 634.28  & 334.03  & 574.53  & 634.51  \\
     & \textbf{Removed (C)} & \textbf{440.92 } & \textbf{597.31 } & \textbf{696.80 } & \textbf{440.07 } & \textbf{591.31 } & \textbf{694.97 } & \textbf{507.54 } & \textbf{680.23 } & \textbf{800.62 } & \textbf{445.48 } & \textbf{599.07 } & \textbf{703.96 } & \textbf{446.51 } & \textbf{600.50 } & \textbf{705.66 } \\
\cmidrule{2-17}      & \textbf{Original (T)} & 1209.50  & 1306.26  & 1349.04  & 1229.54  & 1327.42  & 1372.96  & 1409.23  & 1524.04  & 1578.49  & 1227.99  & 1325.01  & 1368.67  & 1229.06  & 1326.23  & 1369.93  \\
      & \textbf{Removed (T)} & \textbf{1195.08 } & \textbf{1336.93 } & \textbf{1392.43 } & \textbf{1217.34 } & \textbf{1363.25 } & \textbf{1422.13 } & \textbf{1401.04 } & \textbf{1571.49 } & \textbf{1642.44 } & \textbf{1203.52 } & \textbf{1348.72 } & \textbf{1410.90 } & \textbf{1206.12 } & \textbf{1351.65 } & \textbf{1413.88 } \\
\cmidrule{2-17}      & \textbf{Original (S)} & 554.58  & 937.63  & 1063.39  & 552.96  & 952.73  & 1087.15  & 637.50  & 1094.72  & 1253.29  & 564.39  & 948.81  & 1076.25  & 564.90  & 949.32  & 1076.86  \\
      & \textbf{Removed (S)} & \textbf{572.68 } & \textbf{965.17 } & \textbf{1101.25 } & \textbf{579.51 } & \textbf{967.04 } & \textbf{1109.02 } & \textbf{672.54 } & \textbf{1115.38 } & \textbf{1278.07 } & \textbf{575.43 } & \textbf{972.25 } & \textbf{1108.37 } & \textbf{576.77 } & \textbf{974.47 } & \textbf{1110.90 } \\
\midrule
\multirow{6}[2]{*}{\textbf{LaMini-GPT}} 
& \textbf{Original (C)} & 323.34  & 367.46  & 376.55  & 2091.80  & 2643.98  & 2707.52  & 2403.58  & 3098.30  & 3169.53  & 1598.12  & 1997.75  & 2043.54  & 1597.50  & 1995.33  & 2041.12  \\
     & \textbf{Removed (C)} & \textbf{132.15 } & \textbf{151.14 } & \textbf{151.33 } & \textbf{582.31 } & \textbf{727.35 } & \textbf{728.12 } & \textbf{734.07 } & \textbf{924.54 } & \textbf{925.36 } & \textbf{624.50 } & \textbf{760.50 } & \textbf{761.69 } & \textbf{623.63 } & \textbf{758.88 } & \textbf{760.06 } \\
\cmidrule{2-17}      & \textbf{Original (T)} & 368.67  & 379.73  & 379.73  & 2539.10  & 2588.35  & 2588.35  & 3066.39  & 3130.40  & 3130.40  & 2106.89  & 2148.49  & 2148.49  & 2104.61  & 2146.16  & 2146.16  \\
     & \textbf{Removed (T)} & \textbf{109.46 } & \textbf{124.69 } & \textbf{128.23 } & \textbf{538.88 } & \textbf{620.64 } & \textbf{638.58 } & \textbf{681.90 } & \textbf{783.73 } & \textbf{805.72 } & \textbf{546.36 } & \textbf{629.52 } & \textbf{650.19 } & \textbf{545.67 } & \textbf{628.73 } & \textbf{649.42 } \\
\cmidrule{2-17}      & \textbf{Original (S)} & 347.41  & 366.07  & 366.42  & 2157.36  & 2371.21  & 2372.00  & 2510.74  & 2792.73  & 2793.60  & 1746.59  & 1919.43  & 1920.28  & 1747.96  & 1919.32  & 1920.17  \\
     & \textbf{Removed (S)} & \textbf{104.26 } & \textbf{137.73 } & \textbf{144.21 } & \textbf{527.32 } & \textbf{697.10 } & \textbf{722.20 } & \textbf{668.76 } & \textbf{884.99 } & \textbf{916.18 } & \textbf{545.94 } & \textbf{727.06 } & \textbf{750.39 } & \textbf{545.60 } & \textbf{726.90 } & \textbf{750.66 } \\
\midrule
\multirow{6}[2]{*}{\textbf{CodeGen}} 
& \textbf{Original (C)} & 109.93  & 139.88  & 168.23  & 321.08  & 434.06  & 533.72  & 336.01  & 453.64  & 558.45  & 351.20  & 478.08  & 592.39  & 351.46  & 478.36  & 592.75  \\
      & \textbf{Removed (C)} & \textbf{97.58 } & \textbf{117.32 } & \textbf{132.52 } & \textbf{306.48 } & \textbf{367.30 } & \textbf{404.72 } & \textbf{319.78 } & \textbf{383.35 } & \textbf{422.02 } & \textbf{327.14 } & \textbf{396.39 } & \textbf{435.68 } & \textbf{327.33 } & \textbf{396.65 } & \textbf{435.95 } \\
\cmidrule{2-17}      & \textbf{Original (T)} & 182.42  & 182.42  & 182.42  & 578.30  & 578.30  & 578.30  & 602.68  & 602.68  & 602.68  & 639.91  & 639.91  & 639.91  & 640.26  & 640.26  & 640.26  \\
      & \textbf{Removed (T)} & \textbf{131.58 } & \textbf{160.20 } & \textbf{167.04 } & \textbf{407.21 } & \textbf{512.63 } & \textbf{541.95 } & \textbf{424.16 } & \textbf{535.21 } & \textbf{566.01 } & \textbf{443.26 } & \textbf{562.78 } & \textbf{595.73 } & \textbf{443.52 } & \textbf{563.03 } & \textbf{596.04 } \\
\cmidrule{2-17}      & \textbf{Original (S)} & 176.30  & 187.59  & 187.59  & 575.42  & 615.61  & 615.61  & 593.11  & 635.62  & 635.62  & 607.64  & 653.51  & 653.51  & 607.97  & 653.91  & 653.91  \\
      & \textbf{Removed (S)} & \textbf{150.11 } & \textbf{169.18 } & \textbf{178.25 } & \textbf{501.52 } & \textbf{560.08 } & \textbf{580.77 } & \textbf{521.86 } & \textbf{582.85 } & \textbf{604.39 } & \textbf{536.43 } & \textbf{604.63 } & \textbf{627.90 } & \textbf{536.68 } & \textbf{604.95 } & \textbf{628.28 } \\
    \bottomrule[1.2pt]
    \end{tabular}%
    }
  \label{tab:ablation}%
\end{table*}%

\noindent\fakeparagraph{Experimental Setup}
In our evaluation of various LLMs, we randomly choose 1,000 seed inputs and apply \tool (with $p_i^{o_i}$ removed from $f(x)$) to generate 1,000 abnormal inputs for each type of perturbation. 
We denote the approach with the removed $p_i^{o_i}$  as \texttt{Removed}, and our original approach as \texttt{Original}.
The evaluation metrics employed adhere to those detailed in \secref{sec:severity}. Correspondingly, we average the experimental outcomes over three runs.

\noindent\fakeparagraph{Experimental Results} The results are shown in \tabref{tab:ablation}. From the results, we make two observations:
\textit{(i)}
The test samples generated in the ablation study exhibit a weaker degradation in computational efficiency for LLMs. Specifically, out of 27 control experiments conducted, 20 confirm this finding. On average, the required loops, CPU latency, CPU energy consumption, GPU latency, and GPU energy consumption decreased by 18.21\%, 20.75\%, 20.44\%, 20.04\%, and 20.11\%, respectively.
\textit{(ii)} The decoder-only models are more sensitive to such components. Notably, during the ablation study, the GPU latency of LaMini-GPT saw a significant decrease of 74.07\% compared to control experiments. In contrast, models based on an encoder-decoder architecture exhibited a maximum reduction of only 45.83\%. This heightened sensitivity in decoder-only models can be attributed to their autoregressive nature, which makes them more susceptible to the influence of output context. Therefore, the results demonstrate the effectiveness of the 
$p_i^{o_i}$ component in \tool for identifying critical tokens.

\vspace{2mm}
\begin{center}
\begin{tcolorbox}[colback=gray!10,
                  colframe=black,
                  width=0.98\textwidth,
                  arc=1mm, auto outer arc,
                  boxrule=0.9pt,
                 ]
Answers to \textbf{RQ2.5}: Each component within \tool aligns with the overall design goal and effectively contributes to its performance enhancement.
\end{tcolorbox}
\end{center}

\section{Real world study and possible mitigation strategy}

In this section, we further present a real-world case study to discuss how LLMs' efficiency degradation will impact real-world devices' battery power and the computational latency of commercial models.
After that, we show how developers could apply \tool to improve LLMs' efficiency robustness and mitigate computational resource waste.
Finally, we discuss potential threats that might threaten the applicability of \tool and how we alleviate them.

\subsection{Real-World Case Study on Mobile Devices}

\begin{table}[htp]
     \caption{Input sentences for experiments on mobile devices}
  \resizebox{0.65\textwidth}{!}{
      \begin{tabular}{c|l}
    \toprule
    \multirow{3}[2]{*}{\textbf{Seed Input}} & Death comes often to the soldiers and marines who are  \\
          & fighting in anbar province, which is roughly the size of \\
          & louisiana and is the most intractable region in iraq. \\
    \midrule
    \multirow{3}[2]{*}{\textbf{Test Input}} & Death comes often to the soldiers and marines who are  \\
          & fighting in anbar province, which is roughly the \textbf{\textcolor{red}{(}}size of \\
          & of louisiana and is the most intractable region in iraq. \\
    \bottomrule
    \end{tabular}%
    }
  \label{tab:case}%
\end{table}
\fakeparagraph{Experimental Setup}  We select Google T5 as our evaluation LLM in this case study. 
We first deploy the model on the Samsung Galaxy S9+, which has 6GB RAM and a battery capacity of 3500 mAh.
After that, we select one sentence from the dataset \texttt{ZH19} as our seed input; we then apply \tool to perturb the seed input with character-level perturbation and obtain the corresponding test sample.
The seed sentence and the corresponding test sample are shown in \tabref{tab:case}, where the perturbation is colored in red. Notice the test sample inserts only one character in the seed sentence. This one-character perturbation is very common in the real world due to a user's typo.
Finally, we feed the seed input and test sample to the deployed LLMs and measure the mobile device's battery consumption rate.

\begin{figure}[htbp!]
	\centering
    \includegraphics[width=0.50\textwidth]{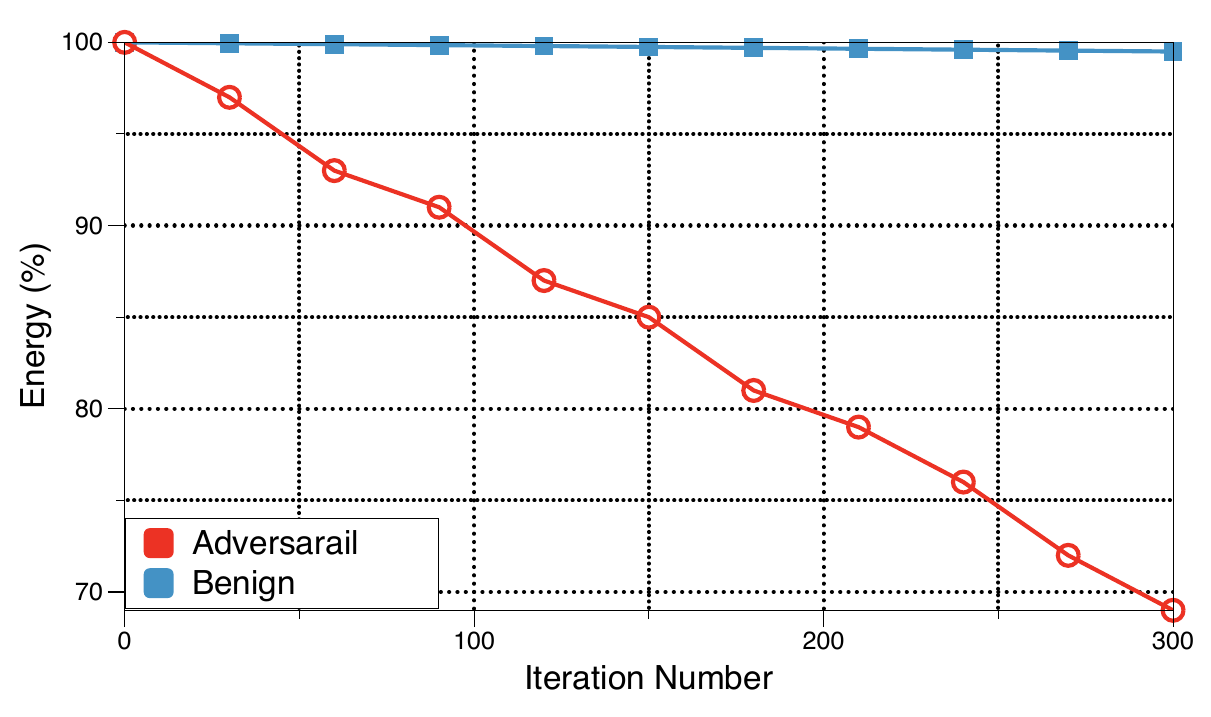}
    \caption{Remaining battery power of the mobile device after T5 
    original seed and perturbed sentences}
	\label{fig:mobile}
\end{figure}

\fakeparagraph{Experimental Results} The mobile phone's battery consumption status is shown in \figref{fig:mobile}. The red line is for the perturbed input, and the blue one is for the original seed input. The results show that the perturbed input consumes the mobile's battery power significantly more quickly than the seed input. Specifically, after 300 iterations, the perturbed input consumes 30\% of the battery power, while the seed input consumes less than 1\%.
The results demonstrate the vulnerability of the efficiency degradation for mobile devices.
Recall that the perturbed example used in our experiment only inserts one character in the seed sentence, which would mimic many practical scenarios (e.g., typo).
Thus, the results suggest the criticality and the necessity of improving LLMs' efficiency robustness.

\begin{table*}[hbtp]
    \centering
    \caption{The accuracy and extra overheads of the \tool detector}
    \vspace{-3mm}
    \label{tab:LLMdetector}
    \resizebox{\linewidth}{!}{
        \begin{NiceTabular}{lcccccccccccc}
        \CodeBefore
        \Body
    \toprule
    \textbf{Methods} & \multicolumn{4}{c}{\textbf{H-NLP}} & \multicolumn{4}{c}{\textbf{AllenAI}} & \multicolumn{4}{c}{\textbf{T5}} \\
\cmidrule{2-13}          & \textbf{Acc} & \textbf{AUC} & \textbf{Overheads} & \textbf{Energy} & \textbf{Acc} & \textbf{AUC} & \textbf{Overheads} & \textbf{Energy} & \textbf{Acc} & \textbf{AUC} & \textbf{Overheads} & \textbf{Energy} \\
    \midrule
    \textbf{\tool(C)} & 99.98 & 100.00 & 0.17  & 0.09  & 100.00 & 100.00 & 0.17  & 0.11  & 99.97 & 100.00 & 0.08  & 0.05 \\
    \textbf{\tool(T)} & 99.99 & 100.00 & 0.32  & 0.17  & 100.00 & 100.00 & 0.08  & 0.05  & 100.00 & 100.00 & 0.06  & 0.04 \\
    \textbf{\tool(S)} & 99.98 & 100.00 & 0.18  & 0.12  & 87.00 & 98.32 & 0.49  & 0.30  & 99.99 & 100.00 & 0.03  & 0.02 \\
    \textbf{Mixed} & 99.98 & 100.00 & 0.74  & 0.48  & 98.00 & 100.00 & 0.86  & 0.79  & 100.00 & 100.00 & 0.18  & 0.11 \\
    \midrule
    \textbf{Methods} & \multicolumn{4}{c}{\textbf{U-DL}} & \multicolumn{4}{c}{\textbf{FairSeq}} & \multicolumn{4}{c}{\textbf{MarianMT}} \\
\cmidrule{2-13}          & \textbf{Acc} & \textbf{AUC} & \textbf{Overheads} & \textbf{Energy} & \textbf{Acc} & \textbf{AUC} & \textbf{Overheads} & \textbf{Energy} & \textbf{Acc} & \textbf{AUC} & \textbf{Overheads} & \textbf{Energy} \\
    \midrule
    \textbf{\tool(C)} & 100.00 & 100.00 & 0.00  & 0.19  & 100.00 & 100.00 & 0.02  & 0.11  & 100.00 & 100.00 & 0.01  & 0.03 \\
    \textbf{\tool(T)} & 100.00 & 100.00 & 0.00  & 0.54  & 100.00 & 100.00 & 0.01  & 0.27  & 100.00 & 100.00 & 0.00  & 0.06 \\
    \textbf{\tool(S)} & 100.00 & 100.00 & 0.01  & 0.31  & 100.00 & 100.00 & 0.01  & 0.16  & 100.00 & 100.00 & 0.03  & 0.02 \\
    \textbf{Mixed} & 100.00 & 100.00 & 0.03  & 0.83  & 100.00 & 100.00 & 0.10  & 0.52  & 98.50 & 100.00 & 0.01  & 0.15 \\
    \midrule
    \textbf{Methods} & \multicolumn{4}{c}{\textbf{Flan-T5}} & \multicolumn{4}{c}{\textbf{LaMini-GPT}} & \multicolumn{4}{c}{\textbf{CodeGen}} \\
\cmidrule{2-13}          & \textbf{Acc} & \textbf{AUC} & \textbf{Overheads} & \textbf{Energy} & \textbf{Acc} & \textbf{AUC} & \textbf{Overheads} & \textbf{Energy} & \textbf{Acc} & \textbf{AUC} & \textbf{Overheads} & \textbf{Energy} \\
    \midrule
    \textbf{\tool(C)} & 83.65 & 89.42 & 0.01  & 0.02  & 100.00 & 100.00 & 0.00  & 0.06  & 100.00 & 100.00 & 0.13  & 0.09 \\
    \textbf{\tool(T)} & 91.00 & 91.38 & 0.04  & 0.09  & 100.00 & 100.00 & 0.01  & 0.25  & 100.00 & 100.00 & 0.37  & 0.43 \\
    \textbf{\tool(S)} & 90.50 & 93.98 & 0.04  & 0.06  & 92.50 & 100.00 & 0.01  & 0.14  & 98.47 & 100.00 & 0.27  & 0.14 \\  
    \textbf{Mixed} & 92.66 & 97.38 & 0.24  & 0.26  & 99.00 & 100.00 & 0.05  & 0.47  & 100.00 & 100.00 & 0.71  & 0.79 \\
    \bottomrule
    \end{NiceTabular}
    }
\end{table*}
\begin{table*}[hbtp]
    \centering
    \caption{The accuracy and extra overheads of the \tool-B detector}
    \vspace{-3mm}
    \label{tab:LLMdetector_B}
    \resizebox{\linewidth}{!}{
        \begin{NiceTabular}{lcccccccccccc}
        \CodeBefore
        \Body
    \toprule
    \textbf{Methods} & \multicolumn{4}{c}{\textbf{H-NLP}} & \multicolumn{4}{c}{\textbf{AllenAI}} & \multicolumn{4}{c}{\textbf{T5}} \\
\cmidrule{2-13}          & \textbf{Acc} & \textbf{AUC} & \textbf{Overheads} & \textbf{Energy} & \textbf{Acc} & \textbf{AUC} & \textbf{Overheads} & \textbf{Energy} & \textbf{Acc} & \textbf{AUC} & \textbf{Overheads} & \textbf{Energy} \\
    \midrule
    \textbf{\tool-B (C)} & 100.00 & 100.00 & 0.14  & 0.07  & 100.00 & 100.00 & 0.14  & 0.13  & 100.00 & 100.00 & 0.06  & 0.04 \\
    \textbf{\tool-B (T)} & 100.00 & 100.00 & 0.30  & 0.16  & 100.00 & 100.00 & 0.12  & 0.08  & 100.00 & 100.00 & 0.08  & 0.04 \\
    \textbf{\tool-B (s)} & 100.00 & 100.00 & 0.19  & 0.09  & 95.00 & 100.00 & 0.52  & 0.37  & 97.50 & 100.00 & 0.03  & 0.03 \\
    \textbf{Mixed} & 100.00 & 100.00 & 0.69  & 0.39  & 97.50 & 100.00 & 0.81  & 0.67  & 97.50 & 100.00 & 0.17  & 0.12 \\
    \midrule
    \textbf{Methods} & \multicolumn{4}{c}{\textbf{U-DL}} & \multicolumn{4}{c}{\textbf{FairSeq}} & \multicolumn{4}{c}{\textbf{MarianMT}} \\
\cmidrule{2-13}          & \textbf{Acc} & \textbf{AUC} & \textbf{Overheads} & \textbf{Energy} & \textbf{Acc} & \textbf{AUC} & \textbf{Overheads} & \textbf{Energy} & \textbf{Acc} & \textbf{AUC} & \textbf{Overheads} & \textbf{Energy} \\
    \midrule
    \textbf{\tool-B (C)} & 100.00 & 100.00 & 0.00  & 0.21  & 100.00 & 100.00 & 0.03  & 0.14  & 100.00 & 100.00 & 0.02  & 0.09 \\
    \textbf{\tool-B (T)} & 100.00 & 100.00 & 0.02  & 0.58  & 100.00 & 100.00 & 0.02  & 0.23  & 100.00 & 100.00 & 0.04  & 0.05 \\
    \textbf{\tool-B (S)} & 100.00 & 100.00 & 0.00  & 0.42  & 100.00 & 100.00 & 0.00  & 0.15  & 100.00 & 100.00 & 0.02  & 0.04 \\
    \textbf{Mixed} & 97.50 & 100.00 & 0.03  & 0.96  & 100.00 & 100.00 & 0.09  & 0.56  & 100.00 & 100.00 & 0.07  & 0.18 \\
    \midrule
    \textbf{Methods} & \multicolumn{4}{c}{\textbf{Flan-T5}} & \multicolumn{4}{c}{\textbf{LaMini-GPT}} & \multicolumn{4}{c}{\textbf{CodeGen}} \\
\cmidrule{2-13}          & \textbf{Acc} & \textbf{AUC} & \textbf{Overheads} & \textbf{Energy} & \textbf{Acc} & \textbf{AUC} & \textbf{Overheads} & \textbf{Energy} & \textbf{Acc} & \textbf{AUC} & \textbf{Overheads} & \textbf{Energy} \\
    \midrule
    \textbf{\tool-B (C)} & 100.00 & 100.00 & 0.07  & 0.07  & 100.00 & 100.00 & 0.03  & 0.05  & 100.00 & 100.00 & 0.15  & 0.12 \\
    \textbf{\tool-B (T)} & 100.00 & 100.00 & 0.12  & 0.09  & 100.00 & 100.00 & 0.06  & 0.32  & 100.00 & 100.00 & 0.42  & 0.45 \\
    \textbf{\tool-B (S)} & 100.00 & 100.00 & 0.05  & 0.04  & 94.47 & 98.46 & 0.05  & 0.18  & 99.87 & 100.00 & 0.29  & 0.20 \\ 
    \textbf{Mixed} & 100.00 & 100.00 & 0.17  & 0.29  & 98.34 & 99.73 & 0.08  & 0.53  & 99.98 & 100.00 & 0.84  & 0.86 \\
    \bottomrule
    \end{NiceTabular}
    }
\end{table*}

\subsection{Real-World Case Study on Commercial Model}

\fakeparagraph{Experimental Setup}
In this case study, we select OpenAI's \textsf{GPT-3.5} as the evaluation model. We randomly choose 500 entries from the test set of the HellaSwag \cite{zellers2019hellaswag} dataset as seed inputs. Given its status as a commercial model not available in open source, we opt for three types of black-box test methods from \tool (\ie \tool-B (C), \tool-B (T) and \tool-B (S)), with the perturbation level set to 1. For the baseline, we employ all the black-box methods (\ie SynError, SIT, and TransRepair) in our research. The evaluation metrics include I-Loops and I-Latency as discussed in \secref{sec:severity}. Specifically, I-Loops is calculated using the \textit{completion\_tokens} from GPT-3.5's returned JSON, which reflects the number of decoder invocations, correlating with the computational demands (\ie required FLOPs). Concurrently, I-Latency is calculated from another field in the returned JSON, namely \textit{response\_ms}, which represents the time required for the model to generate data upon receiving input.

\begin{table}[htbp]
\centering
\caption{The Average Effectiveness Results of \tool on GPT-3.5} 
\resizebox{0.50\textwidth}{!}{\begin{tabular}{c|c|c|c}
\toprule
\textbf{Subject} & \multicolumn{1}{c|}{\textbf{Methods}} & \multicolumn{1}{c|}{\textbf{I-Loops}} & \multicolumn{1}{c}{\textbf{I-Latency}} \\
\midrule
\multirow{6}{*}{\textbf{GPT3.5}} 
       & \textbf{SynError}                         & -1.19 & -1.46 \\
       & \textbf{SIT}                              & 6.98  & 6.64 \\
       & \textbf{TransRepair}                      & -1.3  & -11.21 \\
       & \textbf{\tool-B (C)}             & 25.64 & 19.96 \\
       & \textbf{\tool-B (T) }            & 176.92 & 156.53 \\
       & \textbf{\tool-B (S)}             & 90.93 & 66.18 \\
\bottomrule
\end{tabular}}
    \label{tab:severity_chatgpt}%
\end{table}

\noindent\fakeparagraph{Experimental Results}
\tabref{tab:severity_chatgpt} shows the average efficiency reduction results of GPT-3.5 under various perturbations. The results indicate that the perturbations generated by \tool-B lead to a notably steeper decline in computational efficiency compared to the baseline methods. Specifically, perturbations produced by \tool-B can increase GPT-3.5's I-Loops and I-Latency by an average of 25.64\% to 176.92\% and 19.96\% to 156.53\%, respectively. It is noteworthy that the perturbations set in this experiment are minimal, at the level of a single character or token. Furthermore, the test inputs conceived by \tool-B (S) not only replicate the structural essence of the original sentences without introducing any grammatical or lexical inaccuracies but also succeed in catalyzing a 66.18\% surge in computational latency. Therefore, the results  substantiate \tool’s efficacy and underscore the prevalent issue of computational efficiency vulnerabilities within real-world LLMs.

\subsection{Mitigating Efficiency Degradation with LLMEffiChecker}

This section shows how developers leverage \tool to develop runtime abnormal input detector, which mitigates possible efficiency degradation and computational waste under the adversary scenario (\eg DOS attack).
In detail, we propose an approach to filter out test inputs that require abnormal computational resources at runtime. Because the abnormal inputs are forced to quit at early stage, thus the computational resources waste are avoided.
The idea of applying input validation to improve DNNs' correctness robustness has been studied in recent works~\cite{dissector,wang2019adversarial}.
However, existing input validation techniques may not be suitable for improving LLMs' efficiency robustness due to the high overheads.
Our intuition is that although normal inputs and the computational resource heavy inputs look similar in human eyes, the latent representations of these two categories of inputs are quite different~\cite{dissector}.
Thus, we can leverage the latent representations of these two category inputs to train a light-weighted SVM classifier and apply the classifier to distinguish abnormal inputs at runtime.
Because the classifier should be light-weighted, getting each input's latent representations is preferable without additional computations.
Specifically, in LLMs, the hidden layer converts input data into a higher-level abstract representation, effectively capturing the essential features and patterns of the input sentences. We propose to use the information in the hidden layer as the latent representation to train a lighted-weighted SVM classifier.

\fakeparagraph{Experimental Setup} For each LLMs in our evaluation, we randomly choose 1,000 seed inputs and apply \tool to generate 1,000 abnormal inputs for each perturbation type.
We randomly select 80\% of the seed inputs and the abnormal inputs as the training data to train the SVM classifier, and use the rest 20\% for testing.
We run the trained SVM classifier on the testing dataset and measure the detectors' AUC score, extra computation overheads.

\fakeparagraph{Experimental Results}
The experimental results in white box and black box scenarios are shown in \tabref{tab:LLMdetector} and \tabref{tab:LLMdetector_B} respectively.
Each column in \tabref{tab:LLMdetector} and \tabref{tab:LLMdetector_B} represents the performance in detecting one specific perturbation type and ``Mixed'' represents the performance in detecting a mixed set of three perturbation types.
We observe that the proposed detector achieves almost perfect detection accuracy with a lowest accuracy of 83.65\%.
Moreover, the proposed detector's overheads and energy consumption are negligible compared to those incurred under the LLM.
All experimental subjects' extra overheads and the energy consumption are merely at most 1\% of the original LLMs' overheads in generation normal sentences.
The results show that our validation-based approach can effectively filter out the abnormal input sentences with negligible overheads.

\subsection{Threat Analysis.}

Our selection of the nine LLMs, namely, Google T5, AllenAI WMT14, H-NLP, U-DL, Facebook FairSeq, MarianMT, Flan-T5, LaMini-GPT and CodeGen, might threaten
the \textit{external validity} of our experimental conclusions. 
We alleviate this threat by the following efforts:
(1) the nine LLMs are very popular and have been widely used among developers (with more than 2,714,275 downloads in Nov 2023);
(2) their underlying DNN models are state-of-the-art models;
(3) these systems differ from each other by diverse topics (\eg model architecture, language, training corpus, training process).
Therefore, our experimental conclusions should generally hold, although specific
data could be inevitably different for other subjects.
Our \textit{internal threat} mainly comes from our definition of different perturbation types. Our introduced perturbation may not always be grammatically correct (\eg inserting one character may result in an unknown token).
However, as discussed in \secref{sec:background}, such perturbations may not be typical but exist in the real-world (\eg user typos, adversarial manner).
Thus, it is meaningful to understand LLMs' efficiency degradation with such realistic perturbations.
Moreover, all three perturbation types are well studied in related works \cite{character_nlp1, character_nlp2, character_nlp3, token_nlp1, token_nlp2, token_nlp3, nmt_se1, nmt_se2, nmt_se3, nmt_se4}.

\section{Related work}

\fakeparagraph{Adversarial Attacks \& DNN Robustness} 
Recent works \cite{carlini2017towards, wu2018adversarial, token_nlp1, token_nlp3, tian2018deeptest, pei2017deepxplore,ChenCL0LT02,song2024fmint} show that DNN-based applications are not robust under adversarial attacks, which generate adversarial examples to fool the state-of-the-art DNN-based applications. Existing adversarial attacks can be grouped as \textit{white-box}, and \textit{black-box} attacks based on their access to the DNN parameters. 
To improve DNNs' robustness and mitigate the threats of adversarial attacks, a series of defense approaches \cite{chen2021attackdist, kd, featuresqueeze, wang2019adversarial, kim2019guiding} have been proposed. 
For example, \texttt{FeatureSqueeze} \cite{featuresqueeze} introduces a series of feature squeeze approaches to mitigate the adversarial perturbations during DNN runtime. \texttt{NNMutate}\cite{wang2019adversarial} identifies that adversarial examples are the data points close to the DNN decision boundary and thus proposes applying model mutation techniques to detect adversarial samples.

\fakeparagraph{DNN's Efficiency}
Recently, the efficiency of DNNs has raised much concern due to their substantial \textit{inference-time} costs.
To improve DNN' \textit{inference-time} efficiency, many existing works have been proposed, categorized into two major techniques.
The first category~\cite{howard2017mobilenets,zhang2018shufflenet,kang2024c} of techniques prune the DNNs offline to identify important neurons and remove unimportant ones.
After pruning, the smaller size DNNs could achieve competitive accuracy compared to the original DNNs while incurring significantly less computational costs.
Another category of techniques~\cite{wang2018skipnet,graves2016adaptive,figurnov2017spatially}, called input-adaptive techniques, dynamically
skip a certain part of the DNNs to reduce the number of computations during inference time.
By skipping certain parts of the DNNs, the input-adaptive DNNs can trade-off between accuracy and computational costs.
However, recent studies~\citep{Haque_2020_CVPR, hong2020panda, chen2022nicgslowdown,chen2022deepperform,chen2023dark} show input-adaptive DNNs are not robustness against the adversary attack, which implies the input-adaptive will not save computational costs under attacks.

\section{Conclusions}

In this work, we study the efficiency robustness of LLMs.
Specifically, we present \tool, a comprehensive framework designed to function effectively in both white-box and black-box scenarios. This innovative framework introduces imperceptible perturbations to seed inputs, strategically reducing the computational efficiency of LLMs. 
Evaluation on nine public-available LLMs shows that \tool can generate effective test inputs that may significantly decrease LLMs' efficiency.

\balance 
\bibliographystyle{plain}
\bibliography{iclr2022_conference}

\end{document}